\pdfoutput=1
\documentclass[runningheads]{llncs}

\usepackage{eccv}

\usepackage{eccvabbrv}

\usepackage{graphicx}
\usepackage{booktabs}

\usepackage[accsupp]{axessibility}  
\usepackage[dvipsnames]{xcolor}

\usepackage[pagebackref,breaklinks,colorlinks,citecolor=eccvblue]{hyperref}

\usepackage{orcidlink}

\definecolor{cvprblue}{rgb}{0.21,0.49,0.74}

\usepackage[capitalize]{cleveref}
\crefname{section}{Sec.}{Secs.}
\Crefname{section}{Section}{Sections}
\Crefname{table}{Table}{Tabs.}
\crefname{table}{Table}{Tabs.}
\usepackage{multirow}
\usepackage{tikz}
\usetikzlibrary{positioning,calc,backgrounds}
\usepackage[]{tikzsymbols}
\usepackage[abs]{overpic}
\usepackage{verbatim}

\usepackage{wrapfig}

\newcommand{\ourdm}{\textsc{LUMEN}}

\usepackage{pifont}
\definecolor{ForestGreen}{rgb}{0.13, 0.55, 0.13}
\newcommand*{\newcite}[1]{~\cite{#1}}

\definecolor{maroon}{rgb}{0.80,0.,0.} 
\definecolor{mygreen}{rgb}{0,0.8,0}
\definecolor{odarkgreen}{HTML}{28B23D}        
\definecolor{odarkred}{HTML}{ED1C1C}            

\usepackage{xcolor}
\usepackage{colortbl}
\newcommand{\grayline}{\arrayrulecolor{gray}\hline\arrayrulecolor{black}}

\newcommand{\myparagraph}[1]{\vspace{1pt}\noindent{\bf #1}}

\begin{document}

\title{Label-free Neural Semantic Image Synthesis} 


\author{Jiayi Wang\inst{1}
\and
Kevin Alexander Laube\inst{1}
\and
Yumeng Li\inst{1,2}
\and
Jan Hendrik Metzen\inst{1}
\and
Shin-I Cheng\inst{1}
\and
Julio Borges\inst{1}
\and
Anna Khoreva\inst{1}
}

\authorrunning{J. Wang et al.}

\institute{Bosch Center for Artificial Intelligence  \\
\email{\{jiayi.wang2, kevinalexander.laube, yumeng.li, janhendrik.metzen, shin-i.cheng, julio.borges, anna.khoreva\}@de.bosch.com}\\
\and University of Mannheim 
}

\maketitle
\vspace{-0.6cm}

\newcommand{\TeaserImageWidth}{0.157}
\newcommand{\TeaserImageHSep}{0.04cm}
\newcommand{\TeaserImageHSepLarge}{0.18cm}
\newcommand{\TeaserImageHSepH}{0.025cm}
\newcommand{\TeaserImageVSep}{0.05cm}
\newcommand{\TeaserImageVSepLarge}{0.5cm}
\newcommand{\TeaserImageBlockHSep}{0.2cm}
\newcommand{\TeaserImageBlockVSep}{0.2cm}
\newcommand{\TeaserTitleHSep}{0.15cm}
\newcommand{\TeaserTitleVSep}{0.35cm}

\newcommand{\TeaserTitleHSepTeaser}{0.15cm}

\newcommand{\TeaserPromptImageWidth}{0.16}
\newcommand{\TeaserPromptVSep}{0.10cm}
\definecolor{teasercolorOne}{RGB}{199,215,229}
\definecolor{teasercolorTwo}{RGB}{178,223,138}

\begin{figure}[]
{
  \centering
  \begin{tikzpicture}[ 
  every node/.style={inner sep=0,outer sep=0},
  background rectangle/.style={fill=teasercolorOne!50},
    show background rectangle
  ]%


  \node (Input1) at (0,0) {\includegraphics[ width=\TeaserImageWidth\linewidth, trim={0cm 0cm 18cm 0cm},clip]{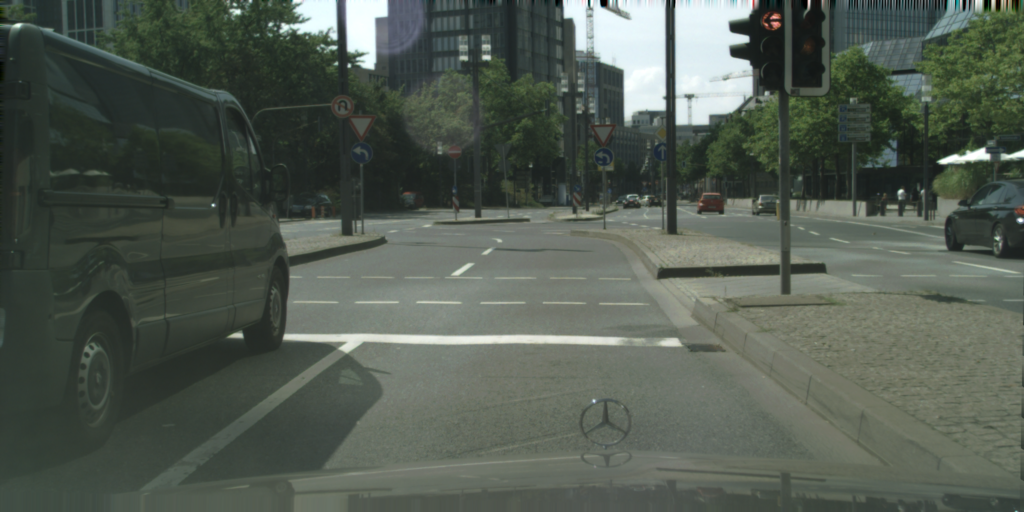}};
   
  \node[right=\TeaserImageHSep of Input1, anchor=west] (descriptor1) {\includegraphics[ width=\TeaserImageWidth\linewidth, trim={0cm 0cm 36cm 0cm},clip]{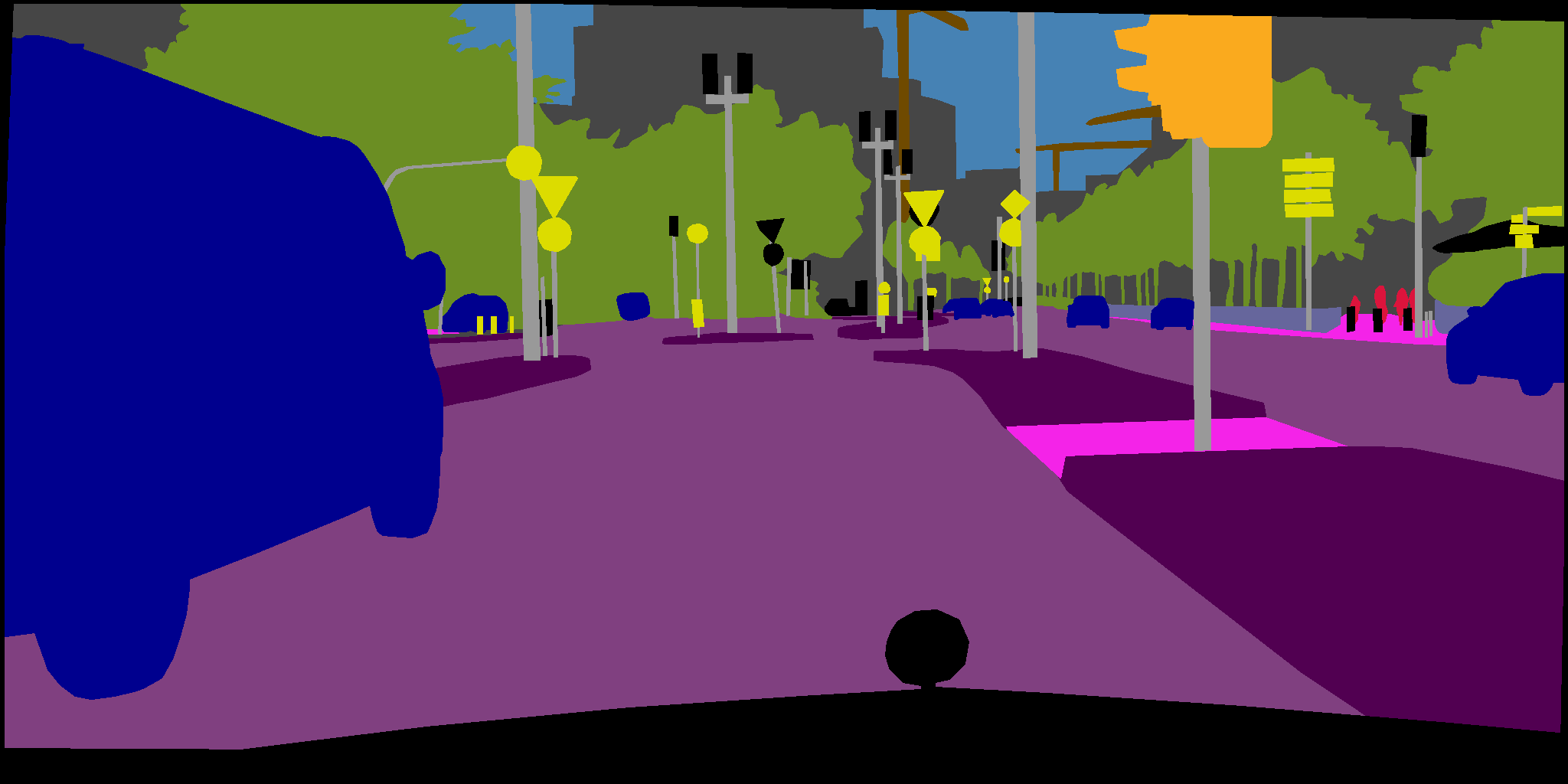}};
  \node (rect) [rectangle,draw,very thick,red,minimum width=1.2cm,minimum height=1.8cm] at (1.6,0) (){};
  
  \node[right=\TeaserImageHSep of descriptor1, anchor=west] (descriptor2) {\includegraphics[width=\TeaserImageWidth\linewidth, trim={0cm 0cm 18cm 0cm},clip]{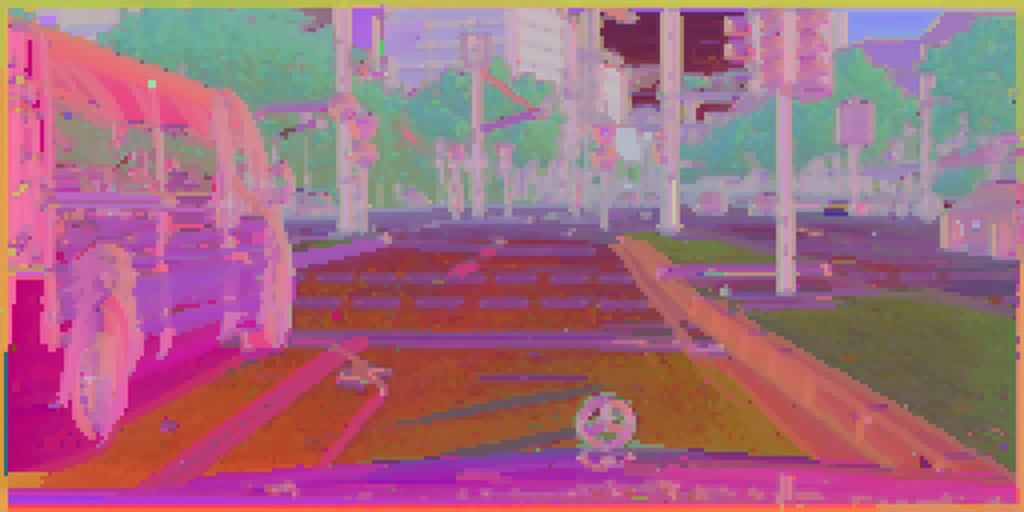}};
   
  \node[below=\TeaserImageVSep of descriptor1] (sample1) {\includegraphics[width=\TeaserImageWidth\linewidth, trim={0cm 0cm 18cm 0cm},clip]{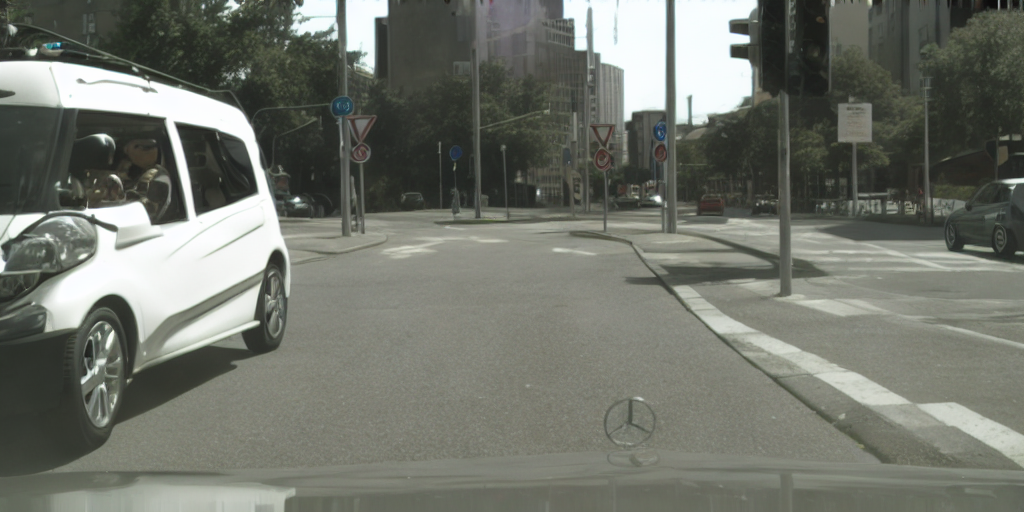}};
  \node (rect) [rectangle,draw,very thick,red,minimum width=1.2cm,minimum height=1.8cm] at (1.6,-1.95) (){};
  
  \node[below=\TeaserImageVSep of descriptor2] (sample2) {\includegraphics[width=\TeaserImageWidth\linewidth,trim={0cm 0cm 18cm 0cm},clip]{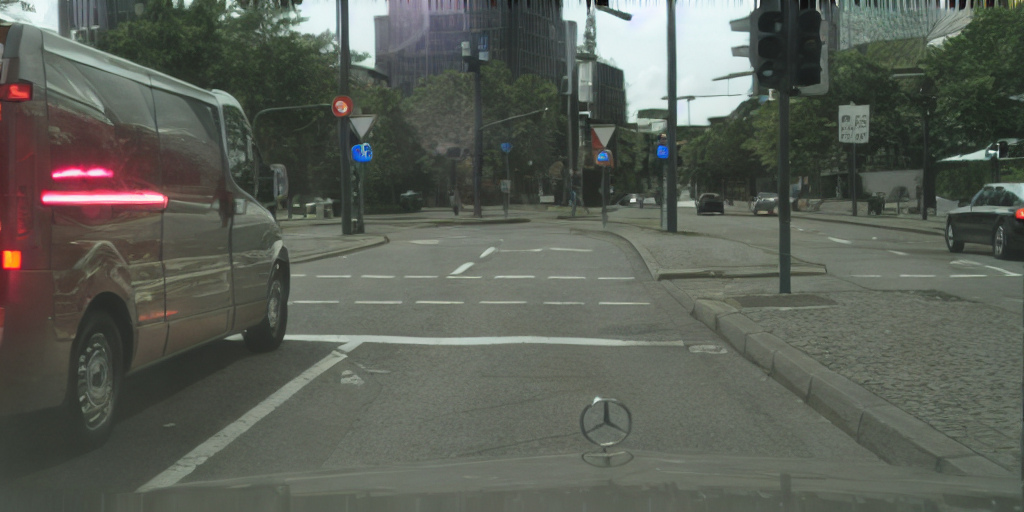}};

  \node[right=\TeaserImageHSepLarge of descriptor2, anchor=west] (descriptor3) {\includegraphics[width=\TeaserImageWidth\linewidth, trim={6cm 5cm 17cm 0cm},clip]{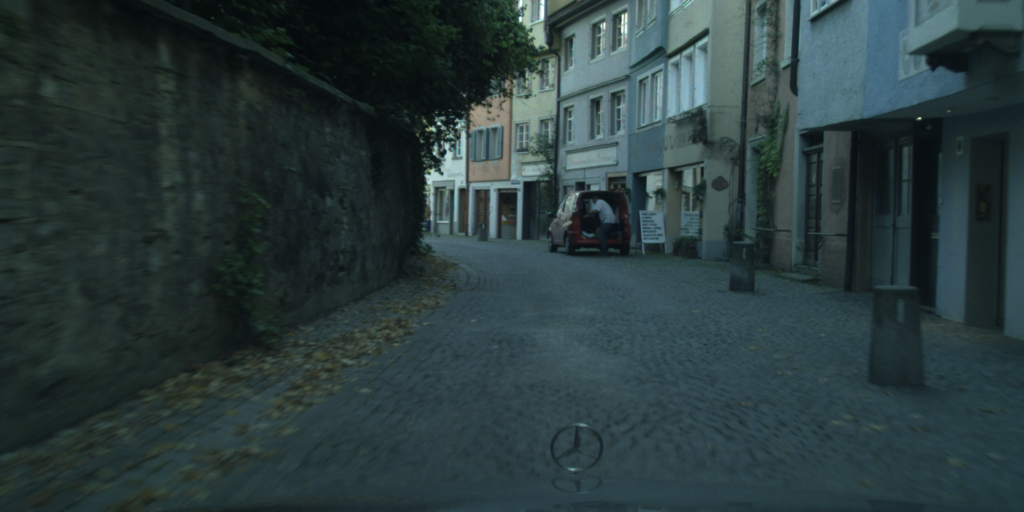}};

  \node[right=\TeaserImageHSep of descriptor3, anchor=west] (descriptor4) {\includegraphics[width=\TeaserImageWidth\linewidth, trim={6cm 5cm 17cm 0cm},clip]{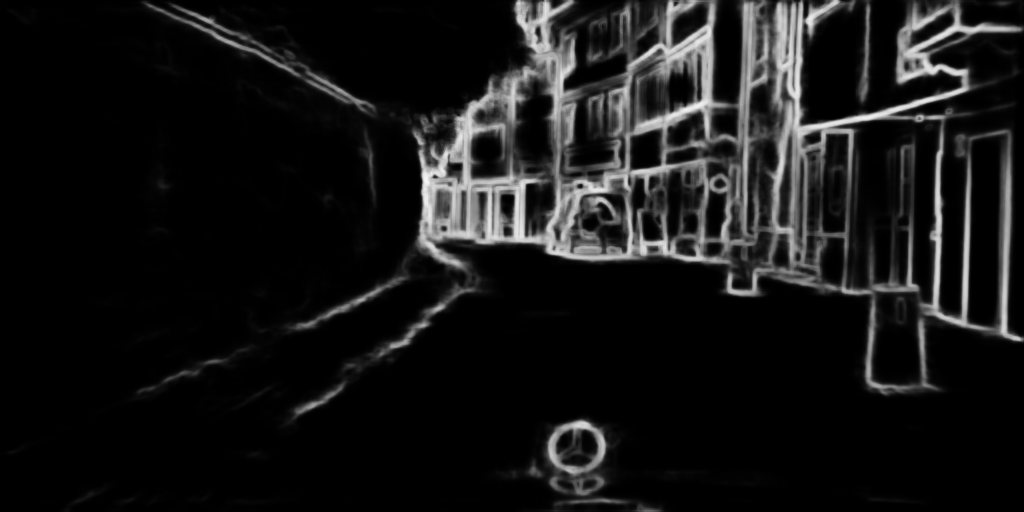}};
  \node (rect) [rectangle,draw,very thick,red,minimum width=1.9cm,minimum height=0.8cm] at (8,0.55) (){};
  \node (rect) [rectangle,draw,very thick,red,minimum width=0.5cm,minimum height=0.9cm] at (7.45,-.45) (){};

  \node[right=\TeaserImageHSep of descriptor4, anchor=west] (descriptor5) {\includegraphics[width=\TeaserImageWidth\linewidth, trim={6cm 5cm 17cm 0cm},clip]{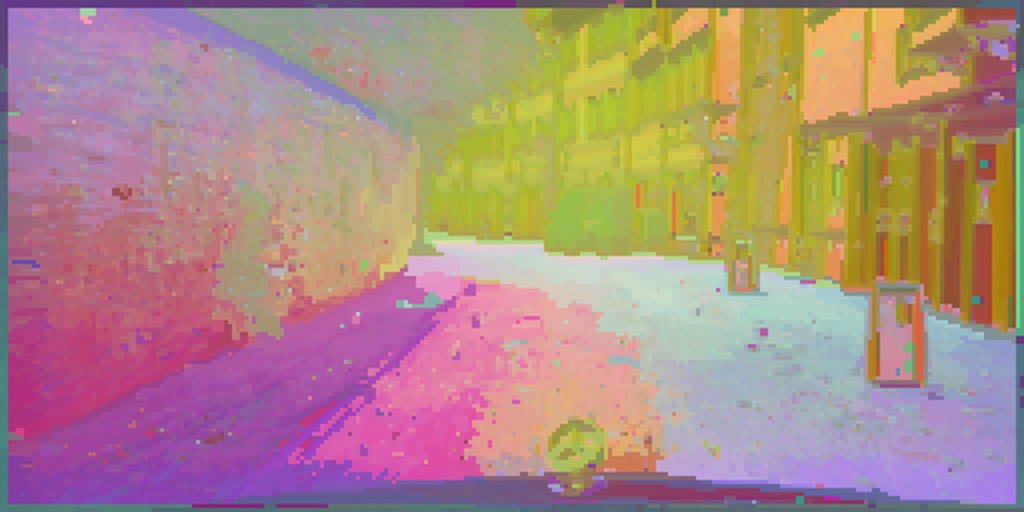}};

  \node[below=\TeaserImageVSep of descriptor4] (sample4) {\includegraphics[width=\TeaserImageWidth\linewidth, trim={6cm 5cm 17cm 0cm},clip]{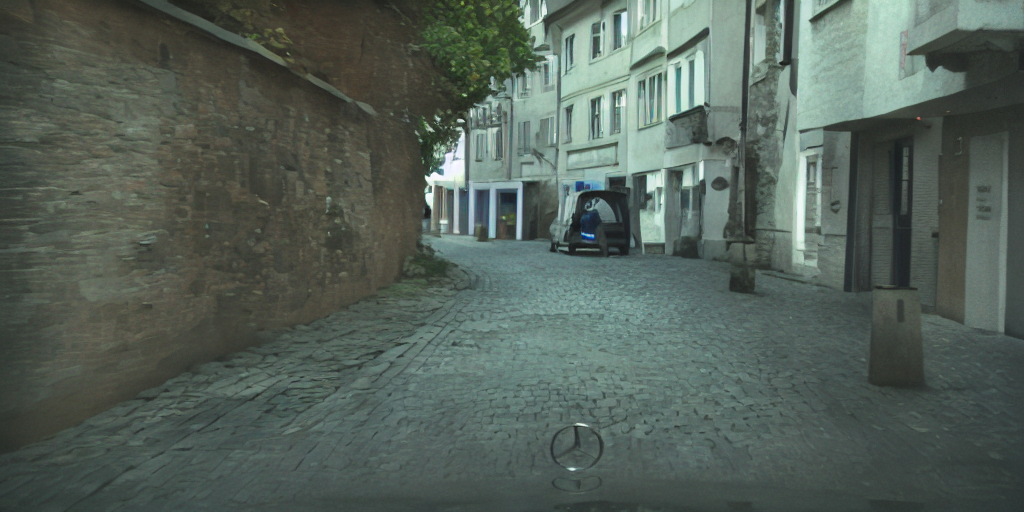}};
  \node (rect) [rectangle,draw,very thick,red,minimum width=1.9cm,minimum height=0.8cm] at (8,-1.4) (){};
  \node (rect) [rectangle,draw,very thick,red,minimum width=0.5cm,minimum height=0.9cm] at (7.45,-2.4) (){};

  \node[below=\TeaserImageVSep of descriptor5] (sample5) {\includegraphics[width=\TeaserImageWidth\linewidth, trim={6cm 5cm 17cm 0cm},clip]{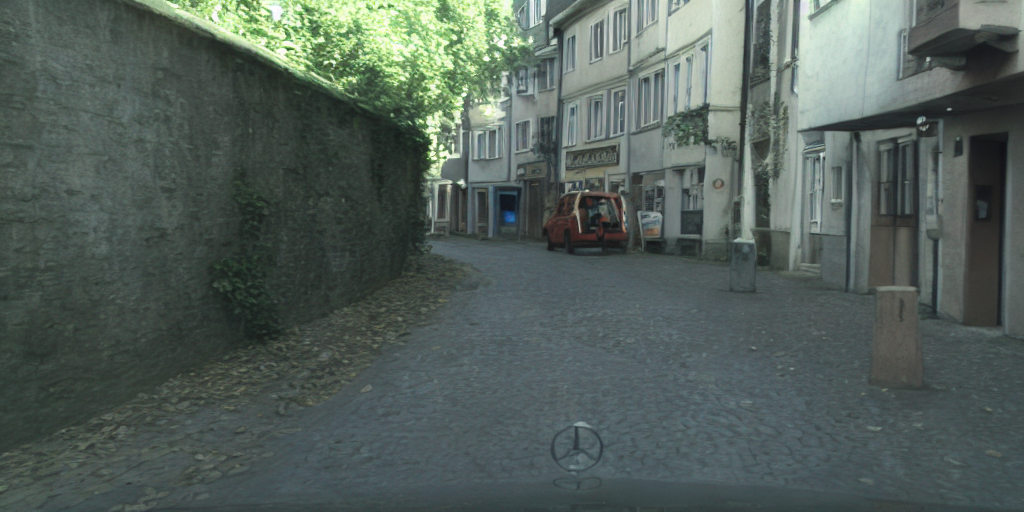}};

   \node[above=\TeaserTitleVSep of Input1, anchor=north] {{Reference}};
  \node[above=\TeaserTitleVSep of descriptor1, anchor=north] {{Sem. Seg.}};
  \node[above=\TeaserTitleVSep of descriptor2, anchor=north] 
  {{Neural Layout}};
  \node[above=\TeaserTitleVSep of descriptor3, anchor=north] {{Reference}};
  \node[above=\TeaserTitleVSep of descriptor4, anchor=north] {{HED}};
  \node[above=\TeaserTitleVSep of descriptor5, anchor=north] {{Neural Layout}};


  \node[below=\TeaserImageVSepLarge of sample1] (descriptor1_2) {\includegraphics[ width=\TeaserImageWidth\linewidth, trim={17cm 5cm 6cm 0cm},clip]{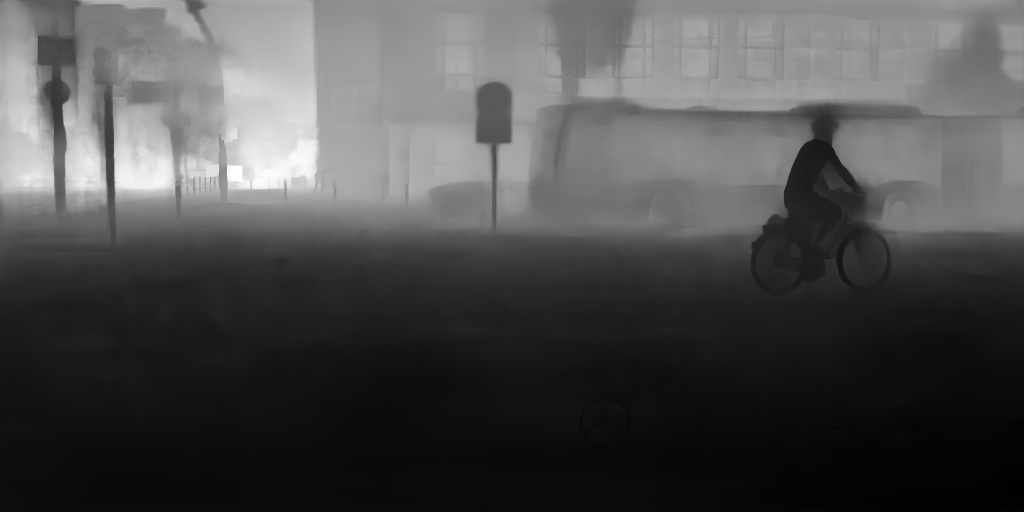}};
  \node (rect) [rectangle,draw,very thick,red,minimum width=1.8cm,minimum height=0.9cm] at (2,-4.3) (){};
  
  \node[right=\TeaserImageHSep of descriptor1_2, anchor=west] (descriptor2_2) {\includegraphics[width=\TeaserImageWidth\linewidth, trim={17cm 5cm 6cm 0cm},clip]{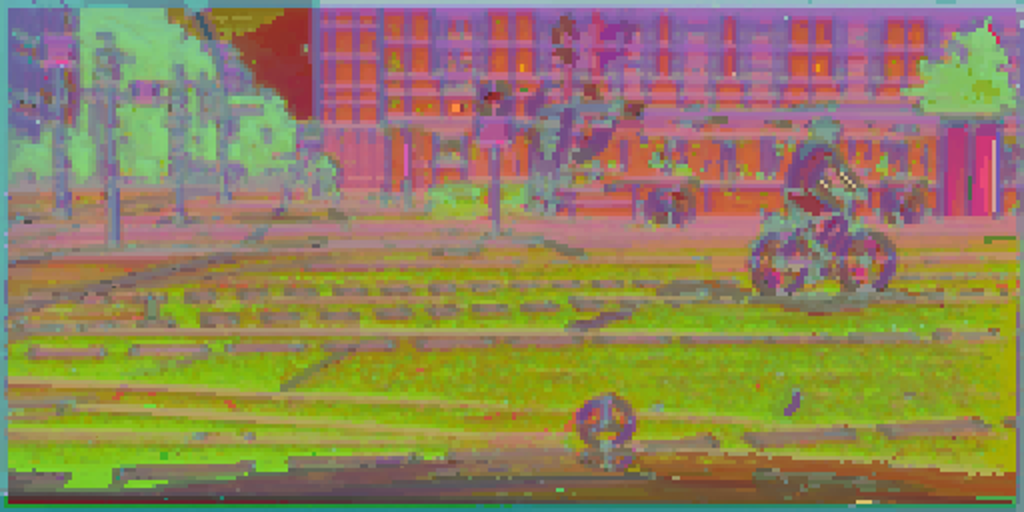}};

  \node[left=\TeaserImageHSep of descriptor1_2]  (Input2) {\includegraphics[ width=\TeaserImageWidth\linewidth, trim={17cm 5cm 6cm 0cm},clip]{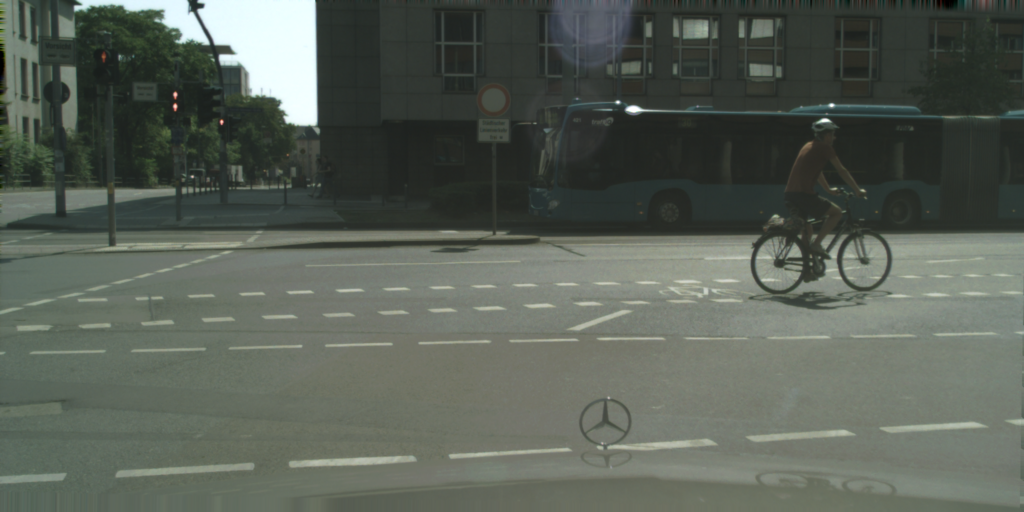}};

  \node[below=\TeaserImageVSep of descriptor1_2] (sample1_2) {\includegraphics[width=\TeaserImageWidth\linewidth, trim={17cm 5cm 6cm 0cm},clip]{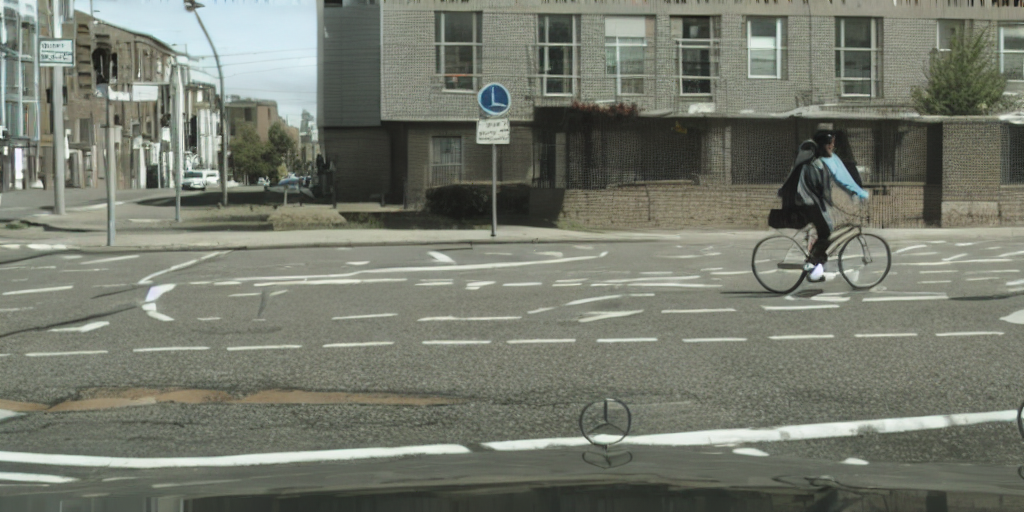}};
  \node (rect) [rectangle,draw,very thick,red,minimum width=1.8cm,minimum height=0.9cm] at (2,-6.25) (){};
  
  \node[below=\TeaserImageVSep of descriptor2_2] (sample2_2) {\includegraphics[width=\TeaserImageWidth\linewidth, trim={17cm 5cm 6cm 0cm},clip]{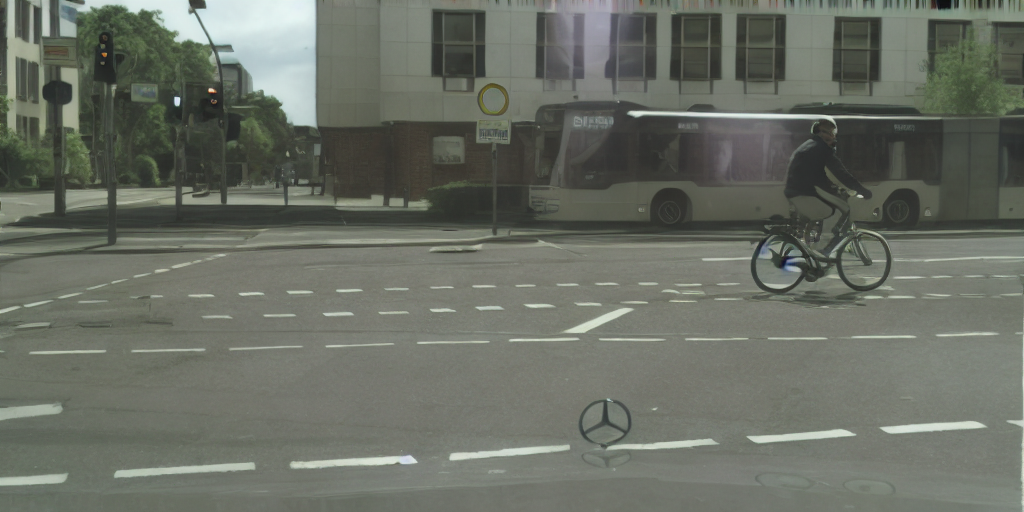}};

   \node[above=\TeaserTitleVSep of Input2, anchor=north] {{Reference}};
  \node[above=\TeaserTitleVSep of descriptor1_2, anchor=north] {{MiDaS}};
  \node[above=\TeaserTitleVSep of descriptor2_2, anchor=north] 
  {{Neural Layout}};

  \node[right=\TeaserImageHSepLarge of descriptor2_2, anchor=west] (descriptor3_2) {\includegraphics[width=\TeaserImageWidth\linewidth, trim={13cm 0cm 5cm 0cm}, clip]{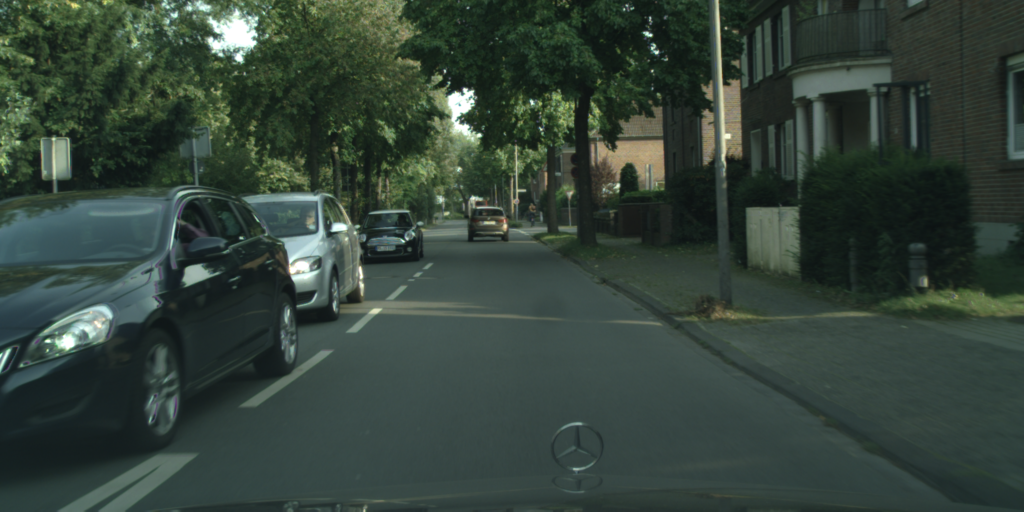}};

  \node[right=\TeaserImageHSep of descriptor3_2, anchor=west] (descriptor4_2) {\includegraphics[width=\TeaserImageWidth\linewidth, trim={13cm 0cm 5cm 0cm}, clip]{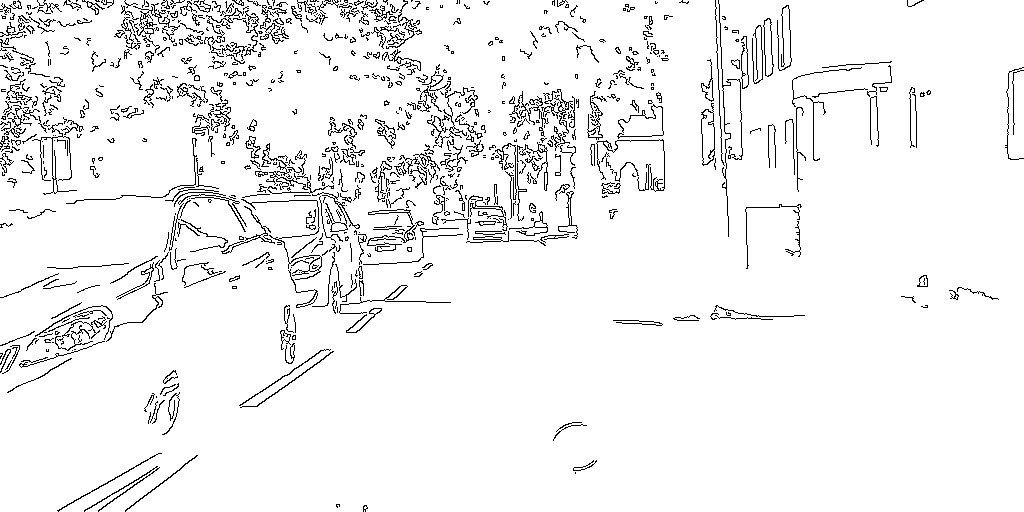}};
  \node (rect) [rectangle,draw,very thick,red,minimum width=1.1cm,minimum height=0.75cm] at (8.35,-4.9) (){};
  
  \node[right=\TeaserImageHSep of descriptor4_2, anchor=west] (descriptor5_2) {\includegraphics[width=\TeaserImageWidth\linewidth, trim={13cm 0cm 5cm 0cm}, clip]{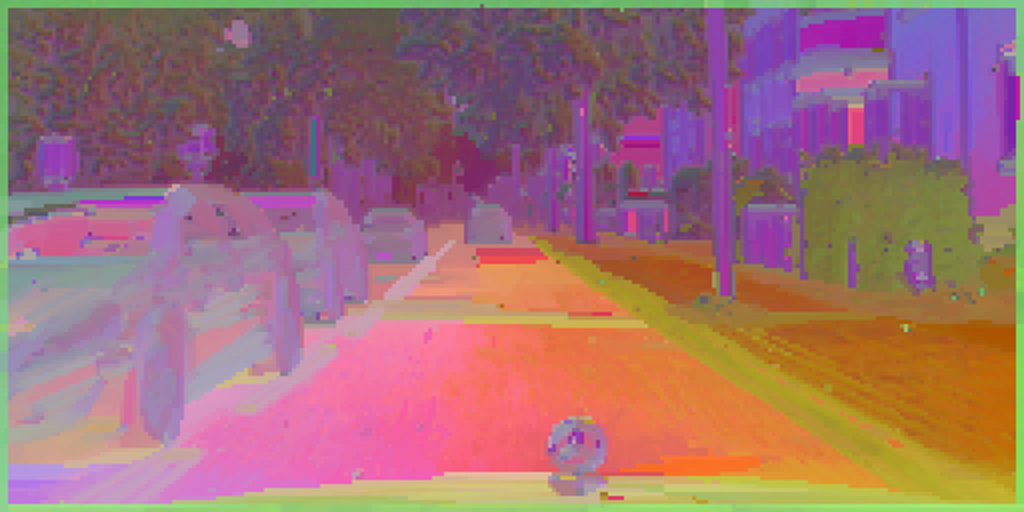}};

  \node[above=\TeaserTitleVSep of descriptor3_2, anchor=north] {{Reference}};
  \node[above=\TeaserTitleVSep of descriptor4_2, anchor=north] {{Canny}};
  \node[above=\TeaserTitleVSep of descriptor5_2, anchor=north] {{Neural Layout}};
  
  \node[below=\TeaserImageVSep of descriptor4_2] (sample4_2) {\includegraphics[width=\TeaserImageWidth\linewidth, trim={13cm 0cm 5cm 0cm}, clip]{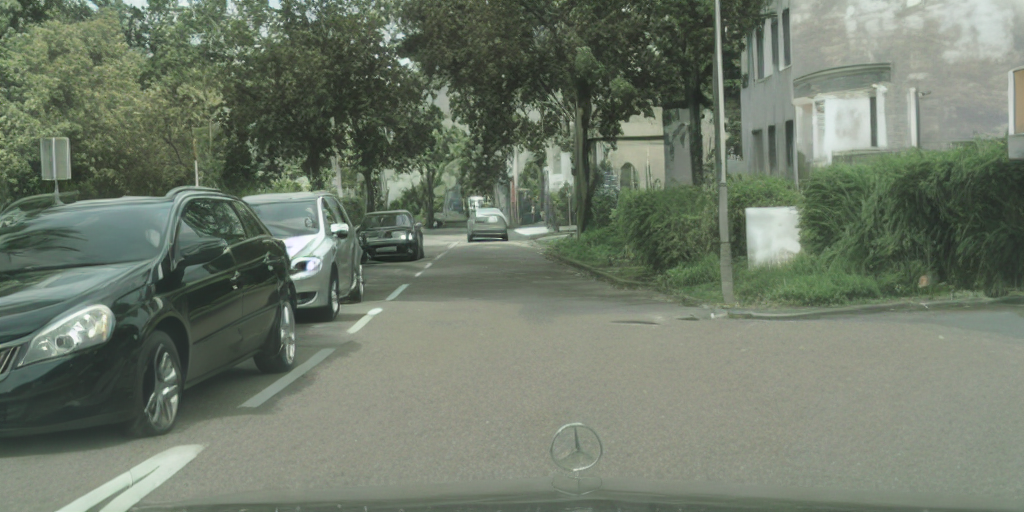}};
  \node (rect) [rectangle,draw,very thick,red,minimum width=1.1cm,minimum height=0.75cm] at (8.35,-6.85) (){};
    
  \node[below=\TeaserImageVSep of descriptor5_2] (sample5) {\includegraphics[width=\TeaserImageWidth\linewidth, trim={13cm 0cm 5cm 0cm}, clip]{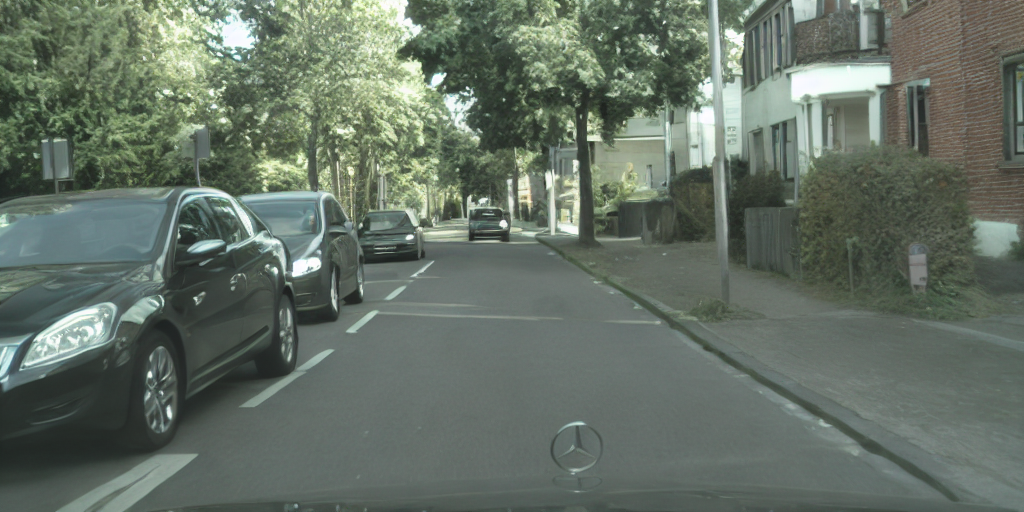}};
  \end{tikzpicture}

  

\begin{tikzpicture}[
  every node/.style={inner sep=0,outer sep=0},
  background rectangle/.style={fill=teasercolorTwo!30},
    show background rectangle
  ]%

  \begin{scope}[local bounding box=group1]

  \node (reference) at (0,0) {\includegraphics[width=\TeaserPromptImageWidth\linewidth]{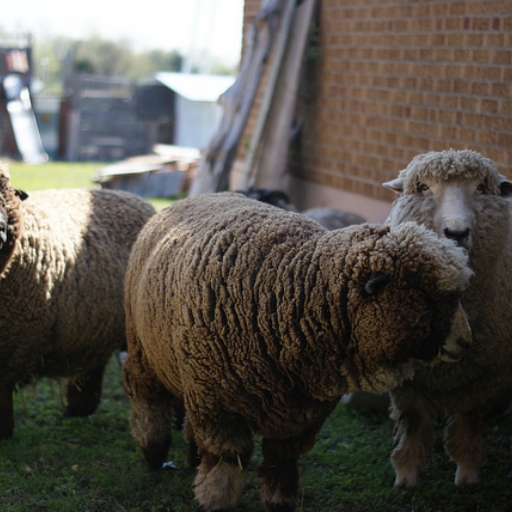}};
  \node[below=\TeaserPromptVSep of reference, anchor=north] {Reference};

\node[right=\TeaserImageHSep of reference] (OG) {\includegraphics[width=\TeaserPromptImageWidth\linewidth]{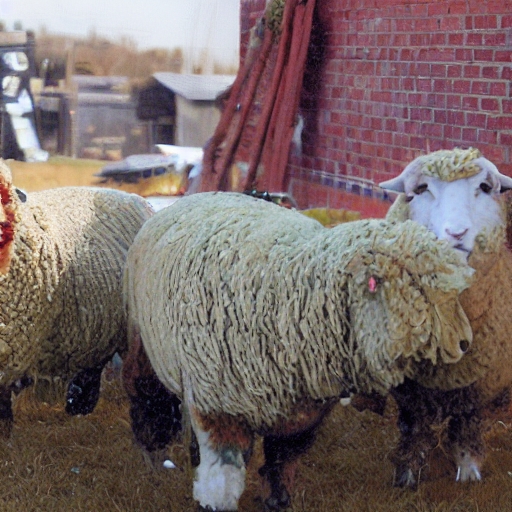}};
  \node[below=\TeaserPromptVSep of OG, anchor=north] { Sample };
  
  \node[right=\TeaserImageHSep of OG] (image1) {\includegraphics[width=\TeaserPromptImageWidth\linewidth]{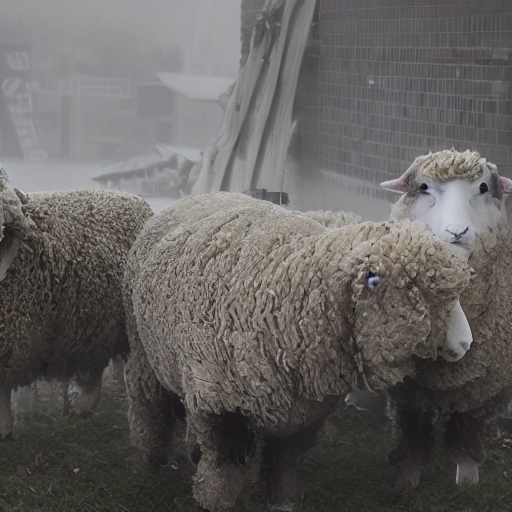}};
  \node[below=\TeaserPromptVSep of image1, anchor=north] {$+$ foggy};
  
  \node[right=\TeaserImageHSep of image1] (image2) {\includegraphics[width=\TeaserPromptImageWidth\linewidth]{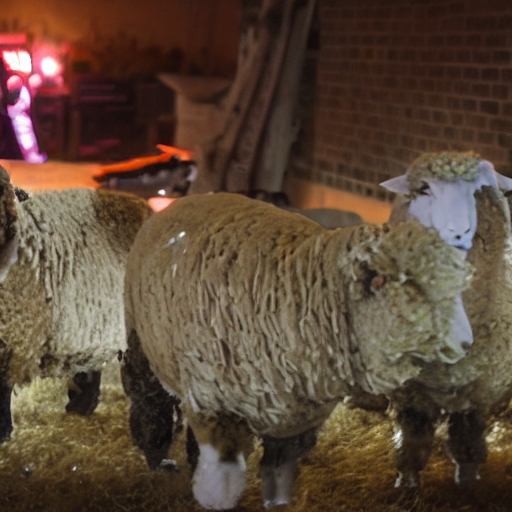}};
  \node[below=\TeaserPromptVSep of image2, anchor=north] {$+$  night};
  
  \node[right=\TeaserImageHSep of image2] (image3) {\includegraphics[width=\TeaserPromptImageWidth\linewidth]{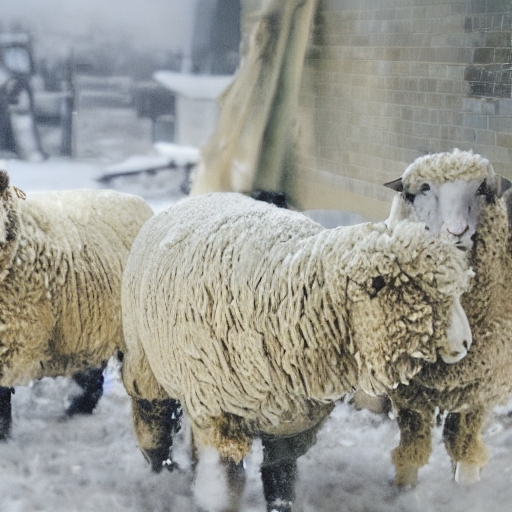}};
  \node[below=\TeaserPromptVSep of image3, anchor=north] {$+$ winter};

  \node[right=\TeaserImageHSep of image3] (image4) {\includegraphics[width=\TeaserPromptImageWidth\linewidth]{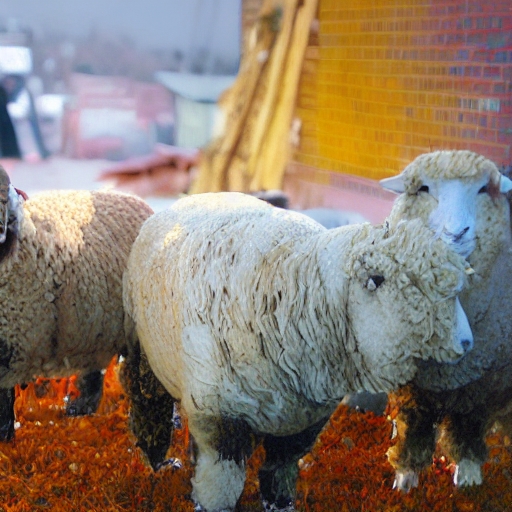}};
  \node[below=\TeaserPromptVSep of image4, anchor=north] {$+$ autumn};

  \end{scope}
  
   \node[above=\TeaserTitleVSep of group1, anchor=north] {{Original caption: A small group of sheep standing together next to a building.}};
  
\end{tikzpicture}
\captionof{figure}{
  Our proposed neural layout conditioning enables the concept of~\emph{neural semantic image synthesis}. 
  Neural layout allows the simultaneous specification of both semantic and spatial concepts, such as scene geometry, object semantics and orientation, all without requiring expensive pixel-wise label annotations for training.
  This is in contrast to existing conditioning, which as shown here in the red boxes can introduce spatial (Sem.Seg.) or semantic (MiDaS, Canny, HED) ambiguity.
  Furthermore, neural layouts are compatible with textual prompts, which further enhances diversity in the images generated via neural semantic image synthesis. 
  } 
  \label{fig:teaser} 
}\end{figure}
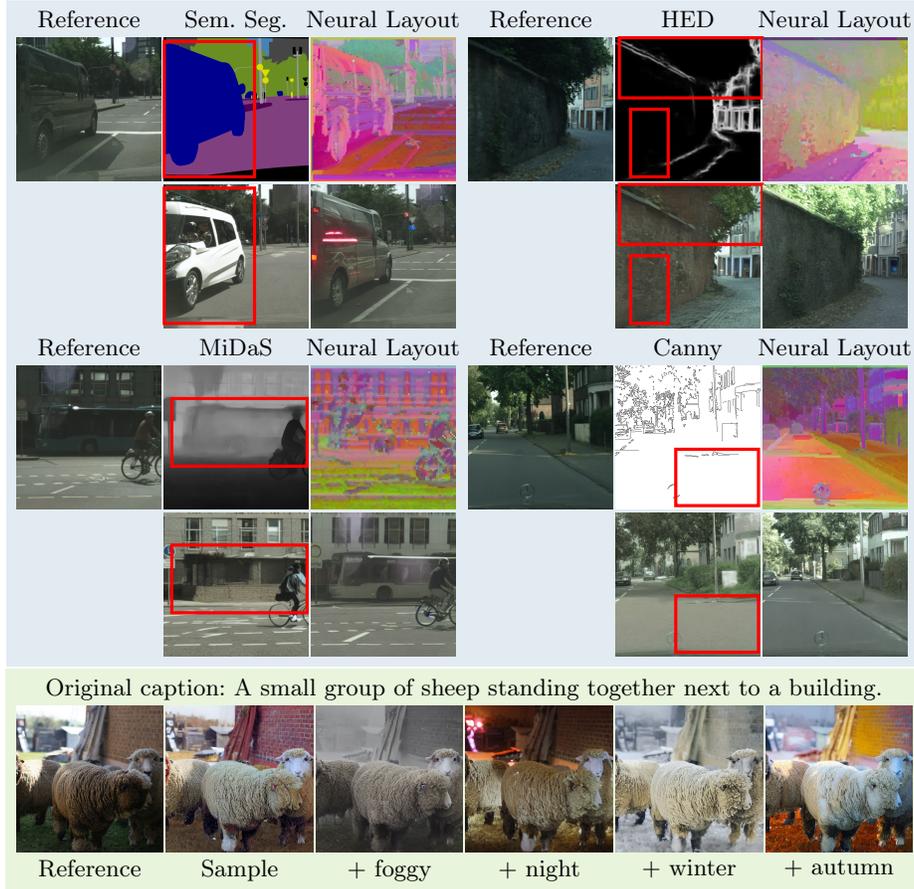
\begin{abstract}
Recent work has shown great progress in integrating spatial conditioning to control large, pre-trained text-to-image diffusion models. 
Despite these advances, existing methods describe the spatial image content using hand-crafted conditioning inputs, which are either semantically ambiguous (e.g., edges) or require expensive manual annotations (e.g., semantic segmentation).
To address these limitations, we propose a new label-free way of conditioning diffusion models to enable fine-grained spatial control. We introduce the concept of \emph{neural semantic image synthesis}, which uses neural layouts extracted from pre-trained foundation models as conditioning. Neural layouts are advantageous as they provide rich descriptions of the desired image, containing both semantics and detailed geometry of the scene. 
We experimentally show that images synthesized via neural semantic image synthesis achieve similar or superior pixel-level alignment of semantic classes compared to those created using expensive semantic label maps. At the same time, they capture better semantics, instance separation, and object orientation than other label-free conditioning options, such as edges or depth.
Moreover, we show that images generated by neural layout conditioning can effectively augment real data for training various perception tasks. 
\vspace{-0.5em}


\end{abstract}

\section{Introduction}
\label{sec:intro}

Controllable image synthesis enables users to specify the desired image content, while relying on a generative model to fill in details that align with the distribution of natural images. 
One way to express the content is through natural language, as popularized by the recent advances of large-scale text-to-image (T2I) diffusion models \newcite{midjourney,ramesh2022dalle2,balaji2022ediffi,rombach2022SD}. 
However, it can be extremely time-consuming to comprehensively specify the detailed spatial composition of the desired image using only textual prompts. Precisely describing a complex scene (such as in~\cref{fig:teaser}), including layout,  pose, shape, and orientation of objects can be a tedious task. It usually requires trial-and-error for the user to refine the prompt such that the generation matches the desired image.

This becomes prohibitive when images need to be generated on a large scale without human-in-the-loop; 
for instance, when using them for training models in downstream perception tasks.
For applications such as 2D/3D object detection, pose estimation, or semantic segmentation, it is crucial to ensure both \emph{diversity} of generated images and their \emph{faithfulness} to spatial details provided in the conditioning, e.g., correct position, size, pose and orientation of objects.

Recent work\newcite{gligen-cvpr,zhang2023controlnet,zhao2023unicontrolnet,mou2023t2i-adapter,GlueGen23ICCV} proposed to introduce additional adapters to integrate spatial conditioning control into the diffusion process for direct image content specification. 
These methods have shown that it is possible to employ segmentation, edge, depth, and normal maps as well as skeletal poses of the reference image as conditioning for description of the image's spatial composition.
Given this variety of conditioning descriptors, it is natural to ask what descriptor is best suited as spatial and semantic descriptor of scenes.
We argue that two properties are key to the general applicability of a descriptor: \emph{richness of semantic and spatial content} and the \emph{ease to obtain descriptor-image pairs} for fine-tuning pre-trained T2I diffusion models. 

\emph{Semantic segmentation maps} are a popular descriptor choice\newcite{xue2023freestyle,wang2022piti,palette,li2024adversarial}, being interpretable high-level abstractions. However, creating them for real images requires either costly and tedious pixel-wise manual annotation or a suitable pre-trained network to generate noisy pseudo-labels
Even more so, segmentation maps by definition do not contain full information about the object pose, orientation, or geometry. 
These ambiguities can lead to unintended scenes where, for example, a car is driving the wrong way down a street (see~\cref{fig:teaser}).
On the other hand, image \emph{edge} and \emph{depth} can simlarly be obtained from unlabeled images (e.g., by using a pretrained edge detector\newcite{xie15hed} or depth estimators\newcite{Ranftl2022midas}) to cheaply create descriptor-image pairs\newcite{zhang2023controlnet} that better capture object geometry. 
However, they are ambiguous in terms of the object semantics.
For example, both ``cat'' and ``blanket'' are plausible interpretations for the object boundaries seen in~\cref{fig:vis-compare}.
Similarly for depth maps, the same geometric shape seen in~\cref{fig:teaser} can be interpreted as either ``building'' or ``bus''.
In short, existing conditioning descriptors can not satisfy both desired properties at once.

\begin{wrapfigure}{R}{0.45\textwidth}
\vspace{-2em}
\centering
\includegraphics[width=0.98\linewidth]{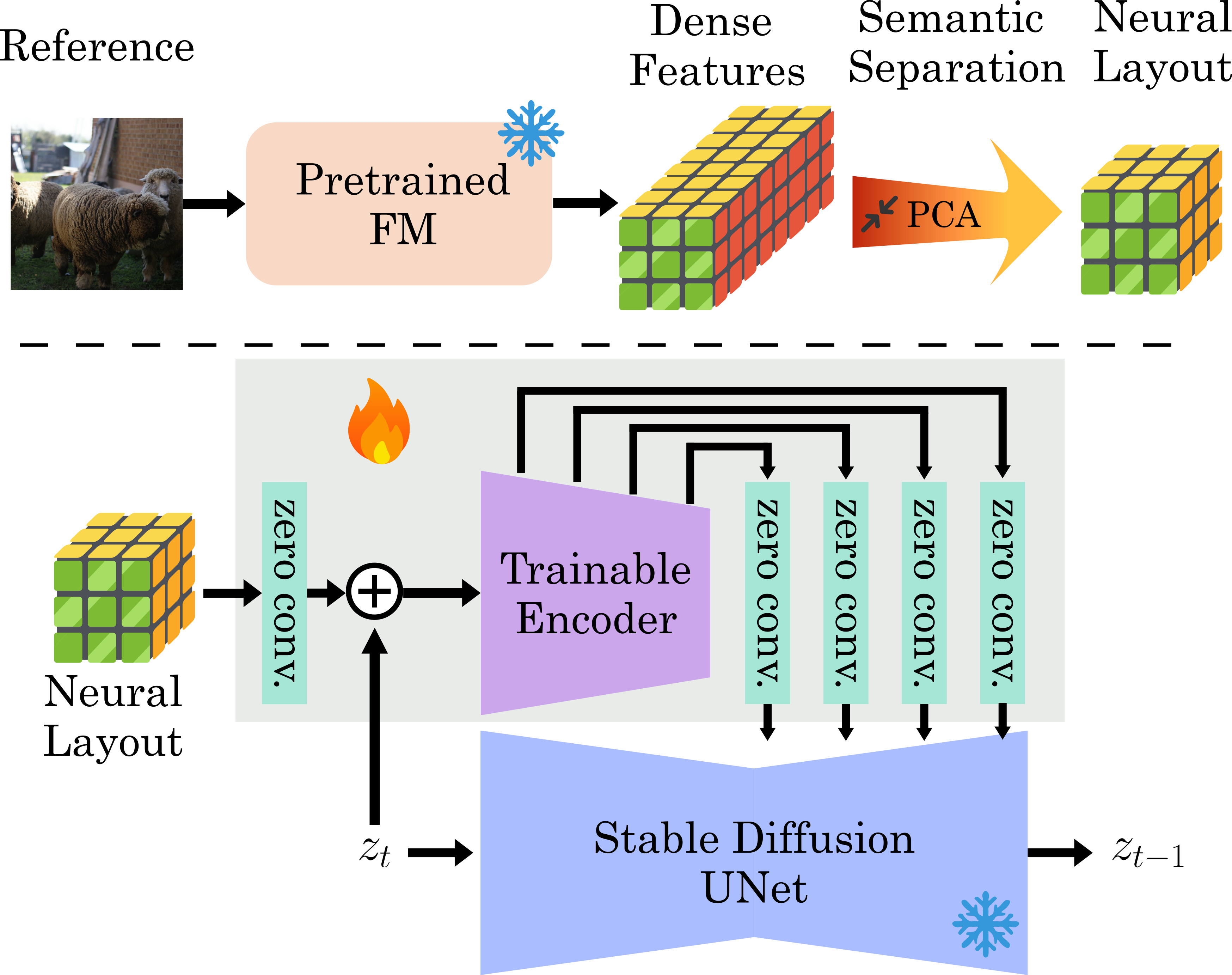}
\caption{~{\ourdm}~uses foundation models (FMs) features to extract neural layouts as conditioning input.
}
\label{fig:method-overview}
\vspace{-2.2em}
\end{wrapfigure}

In this work, we propose a new way of conditioning T2I diffusion models to enable fine-grained spatial control which does not require expensive human annotations.
We introduce the concept of \emph{neural semantic image synthesis}, which derives its conditioning from dense neural features extracted from large-scale foundation models (FMs).
Recent work~\cite{zhou2022extract,oquab2023dinov2,zhao2023vpd,li2023maskdino} has shown that these features preserve well the semantic content and geometry of the images, and thus are well-suited for being rich spatial descriptors of the desired scene. 
However, these features encode nuisance appearance variations which must be removed to ensure diverse synthesis. 
Therefore, we introduce an \emph{semantic separation} step using PCA decomposition to extract only the desired information.
We refer to these compressed features as a \emph{``neural layout''} (see \cref{fig:method-overview}). 

Using neural layouts for conditioning reduces the need to create detailed textual descriptions or costly pixel-level annotations for training images.
This makes it easier to scale up the training data needed to learn conditional control of T2I diffusion models, and thus further boost the quality of image synthesis.
At the same time, neural layout conditioning ensures both faithfulness to semantic and spatial details and diversity of appearance.
Providing both is vital for synthetic data augmentation for training different downstream perception tasks. 

To showcase the benefits of neural layout conditioning, we propose the~{\ourdm}~model which stands for \textbf{L}abel-free ne\textbf{U}ral se\textbf{M}antic imag\textbf{E} sy\textbf{N}thesis. ~{\ourdm}~builds upon ControlNet\newcite{zhang2023controlnet} and uses neural layouts extracted from an image's Stable Diffusion features~\newcite{rombach2022SD} for conditioning (see \cref{fig:method-overview}).
We show that images generated by~{\ourdm}~achieve superior alignment in semantic layout to the reference image even when compared to those created using expensive ground truth semantic label maps (see~\cref{table:comparisonOtherData}).
In comparison to other conditioning inputs obtained from pretrained networks (e.g. predicted edges and depth), images generated with neural layouts capture better the semantics and geometry of the scene (see~\cref{fig:vis-compare}).
Furthermore, we experimentally verify that this improvement in image alignment and quality results in performance gains when used to train downstream perception tasks (see~\cref{table:crossdomain-synthesis} and~\cref{table:CS-DG-compare}).

\vspace{-0.5em}
\section{Related Work}
\vspace{-0.5em}

\myparagraph{Controllable Image Synthesis.}
In the realm of generative adversarial networks (GANs), previous methods\newcite{isola2017image,zhu2017toward,wang2018pix2pixhd,park2019spade,sushko2022oasis,wang2021scgan,tan2021clade} have attempted to utilize various types of conditioning information 
to specify the spatial and semantic content of the synthesized images.
With the increasing popularity of large-scale text-to-image (T2I) diffusion models, e.g., DALL-E 2\newcite{ramesh2022dalle2}, eDiff-I\newcite{balaji2022ediffi} and Stable Diffusion (SD)\newcite{rombach2022SD}, more recent works seek to incorporate additional control mechanism into the T2I models, to better steer the generation process.
One popular direction in this effort is to leverage extra image specifications such as label maps, edges, or depth\newcite{gligen-cvpr,zhang2023controlnet,zhao2023unicontrolnet,mou2023t2i-adapter,xue2023freestyle} to provide a more detailed description of the spatial composition of the desired image, and thus gain finer control over the synthesis process. 
FreestyleNet\newcite{xue2023freestyle} finetunes the entire SD model to rectify the cross attention maps based on the semantic labels.
T2I-Adapter\newcite{mou2023t2i-adapter} and ControlNet\newcite{zhang2023controlnet} both introduce an adapter network on top of a frozen SD backbone to accommodate the conditioning information. 
Uni-ControlNet\newcite{zhao2023unicontrolnet} extends ControlNet by accepting multiple conditioning inputs via two adapters, further enhancing its controllability. 

Different from these prior works and complementary to them, we focus on improving the representation of the image descriptor used by these mechanisms.
We introduce the novel concept of ``neural semantic image synthesis'', which makes use of the rich spatial and semantic knowledge within large-scale pre-trained foundation models (FMs) as conditioning input for T2I diffusion models.
In contrast to existing representations, our proposed neural layouts can capture high-level semantic and geometry information, without human labeling efforts.

\myparagraph{Dense Neural Features.}
Benefiting from advanced pretraining techniques such as self-supervised learning\newcite{caron2021dino,oquab2023dinov2} and joint vision-language pretraining\newcite{clip,rombach2022SD}, large-scale pretrained foundation models (FMs) like CLIP\newcite{clip}, DINO\newcite{caron2021dino}, DINOv2\newcite{oquab2023dinov2}, and Stable Diffusion (SD)\newcite{rombach2022SD} itself 
can serve as robust feature extractors for diverse tasks.
For example, several works\newcite{amir2021deepvit,zhang2023tale,hedlin2023unsupervised} have discovered that these intermediate features can provide well-localized semantic correspondence.
Others have used it for various perception tasks, e.g., object detection, semantic segmentation and depth estimation\newcite{zhou2022extract,oquab2023dinov2,zhao2023vpd,li2023maskdino}.  
A few recent works attempt to leverage these features for image synthesis in the context of image editing.
PnP-Diffusion\newcite{tumanyan2023plug} represents the original image using SD features, which is more amenable to text prompting through the use of cross-attention.
Other works~\cite{ye2023ip,chen2023anydoor} use neural features to encode the exact identity of a subject so that it can be replicated in novel scenes. 

Unlike image editing task, the proposed ``neural semantic image synthesis'' aims to generate \emph{various diverse images}, all of which satisfy the high-level semantics encoded by both the text and image conditions. Hence, instead of capturing the identical appearance, the proposed method focuses on keeping the overall semantics and spatial composition of the scene intact, meanwhile maintaining the freedom to synthesize diverse results, e.g., varying appearances.

\vspace{-0.5em}
\section{Method}
\vspace{-0.1em}
In this section, we introduce the concept of \emph{neural semantic image synthesis}.
Instead of using ad-hoc conditioning to describe the desired output, neural semantic image synthesis makes use of \emph{neural layouts} derived from the dense features of pretrained foundation models.
This semantically and spatially rich representation extracted from a reference image can better specify the desired semantic content and scene geometry without requiring annotations for finetuning.

Our model~{\ourdm} (\cref{fig:method-overview}) takes advantage of the observation that information is stored in easily separable form inside the dense features of foundation models\newcite{caron2021dino,amir2021deepvit,oquab2023dinov2,tumanyan2023plug}.
It has been shown that even a simple linear projection is sufficient to extract diverse contents such as semantic segmentation, depth estimates, and part correspondences\newcite{zhou2022extract,oquab2023dinov2,hedlin2023unsupervised,zhao2023vpd}. 
Analogously, we observe that semantic and spatial scene composition useful for specifying a desired image can be separated from the exact object appearance using a simple linear projector. 
Through this, \ourdm~can align the generation with the spatial and semantic content of a reference image while providing meaningful variations in appearance through the sampling process.

We first introduce how neural layouts are extracted from a reference image in~\cref{Section:PCA}.
We then explain how this can be used for conditional image synthesis with diffusion models in~\cref{Section:Method_Background}.
Finally, we will provide the best practices for extracting semantically and spatially meaningful features from different foundation models in~\cref{Section:Method_DnD}.

\subsection{Neural Layout} \label{Section:PCA} \vspace{-0.5em}

\myparagraph{Dense Feature Extraction.} Modern foundation models make heavy use of self-attention (SA) and cross-attention (CA) modules\newcite{vaswani2017attention}.
Following the convention by Amir et al.\,\newcite{amir2021deepvit}, we introduce the query, key, value, and token features available at each attention layer that is associated with different patches of an input image $\mathbf{x_{ref}}$.
At layer $l$, a SA module takes the tokens of the previous layer $t^{l-1}$, and linearly projects them into queries $q^l_{SA}$, keys $k^l_{SA}$, values $v^l_{SA}$ using matrices $W^l_q$, $W^l_k$, $W^l_v$:
$$ q^l_{SA} = W^l_q \cdot t^{l-1}, k^l_{SA} = W^l_k \cdot t^{l-1}, v^l_{SA} = W^l_v \cdot t^{l-1}.$$
CA modules instead compute the features based on another conditioning input $y$:
$$
q^l_{CA} = W^l_q \cdot t^{l-1}, k^l_{CA} = W^l_k \cdot y, v^l_{CA} = W^l_v \cdot y.
$$
The resulting features are combined through a series of scaled dot-product attention, normalization, multi-layer perceptron, and residual connections to obtain the token feature $t^{l} = \text{TransformerBlock}(q^l, k^l, v^l, t^{l-1})$.

Once extracted, these features are reshaped into a dense features map $f$ corresponding to the original image patches. 
It was noted by several works~\newcite{amir2021deepvit, zhang2023tale} that dense features can have different properties depending on where they are extracted. 
In~\cref{Section:Method_DnD}, we detail the conventions we followed for different foundation models we investigated.

\myparagraph{Semantic Separation.}
Retaining the entire dense feature map $f$ would reveal too detailed information about the reference image $\mathbf{x}_{ref}$. 
Neural semantic image synthesis would then typically lead to samples that are highly similar to $\mathbf{x}_{ref}$, lacking diversity (see~\cref{fig:PCA-impact}).
To prevent this, it is preferable to separate semantic and spatial features from those that encode appearance details. 
Based on existing works~\cite{oquab2023dinov2}, we hypothesize that the principal directions of variation in the dense features should at least partially correspond to what humans intuitively understand as spatial and semantic image content.
Thus, we implement Principal Component Analysis (PCA) to obtain a linear projector that can remove nuisance variations. 

To obtain the neural layout $\mathbf{c}_i$, used as conditioning in~\ourdm, we retain only the information in the top $N$ PCA components, where $N$ is a hyperparameter of our method.
In practice, we compute the PCA projection on a random sample of $40,000$ feature vectors extracted from images in the training set.

\subsection{Conditional Image Synthesis} \label{Section:Method_Background}
Neural semantic image synthesis generates the images $\mathbf{x}$ based upon text prompts $\mathbf{c}_t$ and a neural layout $\mathbf{c}_i$.
Although this task is compatible with a variety of conditional synthesis frameworks, we build~\ourdm~upon ControlNet\newcite{zhang2023controlnet}, due to its popularity and the ease of comparison, as it is compatible with most existing image conditioning. 
ControlNet makes use of the frozen Stable Diffusion model\newcite{rombach2022SD} for text-conditional image synthesis and adds a trainable adapter to incorporate the conditioning.

To train ControlNet, we start with an input image $\mathbf{x}_0$ and encode it into its latent representation $\mathbf{z}_0 = \mathcal{E}(\mathbf{x}_0)$ using the encoder $\mathcal{E}$. We add noise to $\mathbf{z}_0$ and obtain $\mathbf{z}_t$, where $t$ is a time step specifying the scale of the noise added.
The corresponding neural layout $\mathbf{c}_i$ is then also extracted from $\mathbf{x}_0$ to use as conditioning input.
We train a denoiser $\epsilon_{\theta}$, to predict the noise $\epsilon$ added to $\mathbf{z}_t$, with the following loss $\mathcal{L}$:
$$		\mathcal{L}(\theta) = \mathbb{E}_{\mathbf{z}_0, \mathbf{t}, \mathbf{c}_t, \mathbf{c}_i, \epsilon \sim \mathcal{N}(0, 1) }\Big[ \Vert \epsilon - \epsilon_\theta(\mathbf{z}_{t}, \mathbf{t}, \mathbf{c}_t, \mathbf{c}_i)) \Vert_{2}^{2}\Big].$$

\subsection{Foundation Model Backbones} \label{Section:Method_DnD}
We considered 4 different foundation models for extracting neural layouts $\mathbf{c}_i$. 
This covers features from state-of-the-art methods trained in a self-supervised fashion (DINO \cite{caron2021dino} and DINOv2 \cite{oquab2023dinov2}), vision-language models trained contrastively (CLIP \cite{clip}), as well as text-to-image synthesis (Stable Diffusion \cite{rombach2022SD}). 
Since the output resolution of these features varies, all features are upscaled using exact nearest neighbor interpolation to match the image resolution.

\myparagraph{DINO.}
We follow established work~\cite{amir2021deepvit} and use the ``key'' feature from SA layers when processing $\mathbf{x_{ref}}$ as these features contain information useful for challenging semantic correspondence problems.
We use the last SA layer and exclude the key corresponding to the CLS token to compute the neural layout.

\myparagraph{DINOv2.} As a scaled-up version of DINO that is trained on a large amount of curated data, DINOv2 potentially contains more generalizable features.
Here, we use the non-CLS tokens of the last transformer layer for neural layout~\cite{zhang2023tale}.

\myparagraph{CLIP.}
We use the CLIP vision encoder to encoding $\mathbf{x_{ref}}$ and keeping all but the CLS tokens from the last feature layer before the final pooling~\cite{zhou2022extract}.
These features were shown to be well-suited for segmentation tasks.

\myparagraph{Stable Diffusion.}
We first use BLIP~\cite{li2022blip} to obtain text prompts $\mathbf{c}_t$ for the respective image $\mathbf{x_{ref}}$, and encode $\mathbf{x_{ref}}$ in the latent space of Stable Diffusion 1.5 (SD): $\mathbf{z}_0 = \mathcal{E}(\mathbf{x}_{ref})$.
Next, we apply SD's U-Net to predict the noise term $\epsilon_\theta(\mathbf{z}_{0}, \mathbf{0}, \mathbf{c}_{t})$. 
We extract the intermediate activations from layer 2, 5, and 8 of SD's U-Net and upsampled them to match the resolution of layer 8.
All activations are then concatenated across the channel dimension.
These SD features~\cite{zhang2023tale} were shown to have a strong sense of spatial layout, which makes them a promising candidate for neural layouts.

\section{Experiments}

In this section, we first explain our experimental setup and evaluation metrics.
Then we show the impact of our design choices for neural layout conditioning in \cref{sec:ablation}. 
The advantages of neural layouts compared to existing conditioning inputs are demonstrated in \cref{sec:comparison}. 
Finally in~\cref{sec:appplication}, we validate the quality of images synthesized by~\ourdm~by showing its impact on downstream applications.

\myparagraph{Evaluation Metrics.}
To measure the image synthesis quality of our method, we use FID\newcite{heusel2017gansFIDal} for perceptual quality, average LPIPS\newcite{zhang2018perceptual} between generated samples for diversity, and TIFA\newcite{hu2023tifa} for text controllability.
We additionally evaluate how well each conditioning captures the semantic composition and geometry of the scene.
For alignment with semantic layouts, we use mIoU between ground truth segmentation label and those predicted by a pretrained segmenter; here we use Mask2Former\newcite{Cheng2022mask2former} for COCO-Stuff\newcite{caesar2018coco} and ADE20k~\cite{zhou2017ade20k}, DRN~\cite{Yu2017DRN} for Cityscapes~\cite{cordts2016cityscapes}.
However, since mIoU does not contain 3D information, we introduce the scale-invariant depth error (SI depth)\newcite{eigen2014depth}  as a metric for geometric consistency.
SI depth is computed on depth estimated using MiDaS~\cite{Ranftl2022midas} from the reference image and the synthesized image. 
Further evaluation and implementation details are available in the supplementary material.

\subsection{Neural Layout Design Space}
\label{sec:ablation}

We explore the design space of neural layouts on the diverse COCO-Stuff dataset \newcite{caesar2018coco} to determine how to best extract descriptive semantic and spatial information from a given reference image.

\begin{table}[t]
\begin{subtable}[t]{0.49\textwidth}
\flushleft
\resizebox{\textwidth}{!}{
\begin{tabular}{l|ccccc}
Features &  mIoU $\uparrow$ & SI Depth $\downarrow$ & FID $\downarrow$   & LPIPS $\uparrow$ & TIFA $\uparrow$\\
\hline\hline
%
%
CLIP &  41.7 & 25.7 & 15.6 &  \textbf{0.58} & \textbf{0.84}\\
DINO &  49.5 &  22.2 & 12.8 &  0.41 & 0.77\\
DINOv2  & 51.1 & 22.0 &  12.8 &  0.42 & 0.78\\
SD  & \textbf{51.4} & \textbf{21.5} & \textbf{12.2} &  0.42 & 0.79 \\
\end{tabular}   
}
\caption{
Comparison of FMs for extracting neural layouts using 10 PCA components. SD features produces the best overall image quality.
}
\label{table:ablation-densefeat}
\end{subtable}
\begin{subtable}[t]{0.49\textwidth}
\flushright
\resizebox{\textwidth}{!}{
\begin{tabular}{l|ccccc}
Comp. &  mIoU $\uparrow$ & SI Depth $\downarrow$ & FID $\downarrow$   & LPIPS $\uparrow$ & TIFA $\uparrow$ \\
\hline\hline
$1$ & 45.0 & 25.5 & 14.5 & 0.53 & 0.87\\
$5$ & 50.0 & 22.8 & 12.9 & 0.44 & 0.81\\
$10$ & 51.4 & 21.5 & 12.2 &  0.42 & 0.79\\
$20$ & 52.9 & 21.1 & 11.8 & 0.36 & 0.66\\
$100$ & 54.6 & 18.6 &  8.8 & 0.25 & 0.47\\
\grayline
Image & 56.1 & 15.6 & 5.9 & 0.15 & 0.30
\end{tabular}   
}
\caption{
Impact of the number of PCA components on image quality for SD features.
}
\label{table:ablation-pca}
\end{subtable}

\vspace{-0.5em}
\caption{
We ablate the choice of FM feature (a) and the number of PCA components (b) on the COCO dataset. 
We observe that the number of SD components can be used to select the desired trade-off between faithful generation (mIoU, FID, SI Depth) and sample diversity (LPIPS)/text controllability (TIFA). At large number of components, the performance approaches using the reference `Image' itself as conditioning.
}
\vspace{-3em}

\end{table}

\myparagraph{Best Features for Neural Layout.}
For the task of neural semantic image synthesis, we want neural layouts to preserve semantic and geometry content but to discard appearance details.
Thus, we want a backbone that extracts features where this is readily separable. 
To evaluate this, we compare in~\cref{table:ablation-densefeat} the image quality of DINO, DINOv2, CLIP, and SD features at $N=10$ PCA components.

We observe that SD features provide the best perceptual image quality and also retain the semantic content best, while DINOv2 is a close second.
Although CLIP conditioning can generate more varied images, this diversity is due to the weak semantic and spatial constraints imposed during synthesis.
This is likely because CLIP's image-level training objective is less suitable to capture precise pixel-level information without further processing.

\myparagraph{Number of PCA Components.}
Since discarding more PCA components removes more information, useful or not, the resulting synthesis task also becomes less constrained~(See \cref{fig:PCA-impact}).
This is reflected in~\cref{table:ablation-pca} where we compare synthesis results across different numbers of PCA components. 
We observe a clear trade-off between metrics linked to semantic layout and geometry (mIoU and SI Depth), image diversity (LPIPS), and text editability (TIFA).
As additional PCA components impose further constraints, it becomes increasingly likely that these constraints conflict with the text description. 
In these cases, the generator prioritizes the more constraining image conditioning over the more ambiguous text conditions. 
Note that when the number of components is chosen large enough, conditioning with neural layout simply reproduces the reference image and the performance approaches that obtained from directly using the reference image as conditioning input (see~\cref{table:ablation-pca}).

\newcommand{\PCACompImageWidth}{0.115} 
\newcommand{\PCACompImageWidthHalf}{0.083}
\newcommand{\PCACompImageHSep}{0.05cm}
\newcommand{\PCACompImageHSepLarge}{0.5cm}
\newcommand{\PCACompImageHSepH}{0.025cm}
\newcommand{\PCACompImageVSep}{0.05cm}
\newcommand{\PCACompImageVSepLarge}{0.4cm}
\newcommand{\PCACompImageBlockHSep}{0.2cm}
\newcommand{\PCACompImageBlockVSep}{0.2cm}
\newcommand{\PCACompTitleHSep}{0.15cm}
\newcommand{\PCACompTitleVSep}{0.35cm}
\newcommand{\PCACompTitleHSepPCAComp}{0.15cm}

\begin{figure}[t]
\vspace{-1em}
    \centering
  \begin{tikzpicture}[every node/.style={inner sep=0,outer sep=0}]%

  \node (descriptor1) at (0,0) {\includegraphics[width=\PCACompImageWidth\linewidth, trim={1cm 5cm 5cm 1cm},clip]{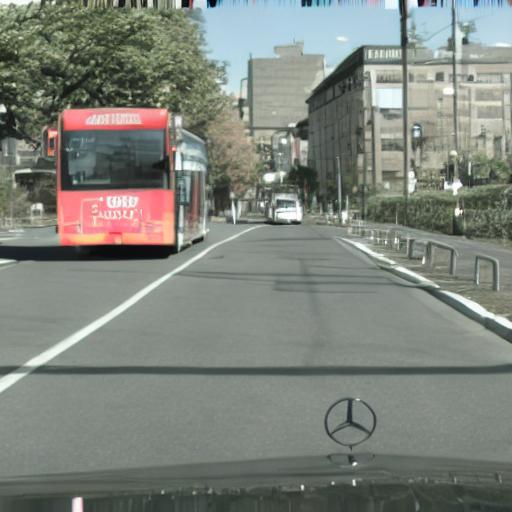}};
  \node[right=\PCACompImageHSep of descriptor1, anchor=west] (descriptor2) {\includegraphics[width=\PCACompImageWidth\linewidth, trim={1cm 5cm 5cm 1cm},clip]{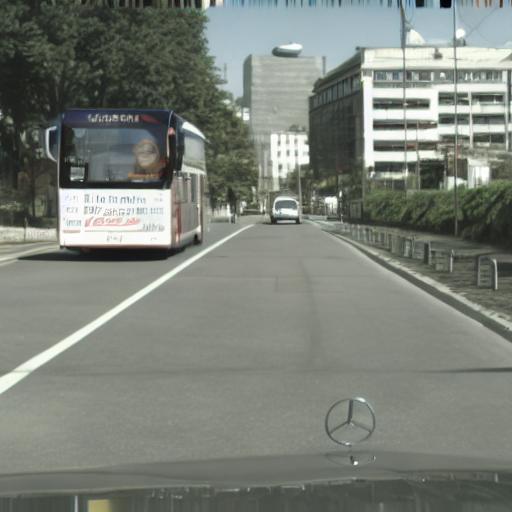}};
  \node[right=\PCACompImageHSep of descriptor2, anchor=west] (descriptor3) {\includegraphics[width=\PCACompImageWidth\linewidth, trim={1cm 5cm 5cm 1cm},clip]{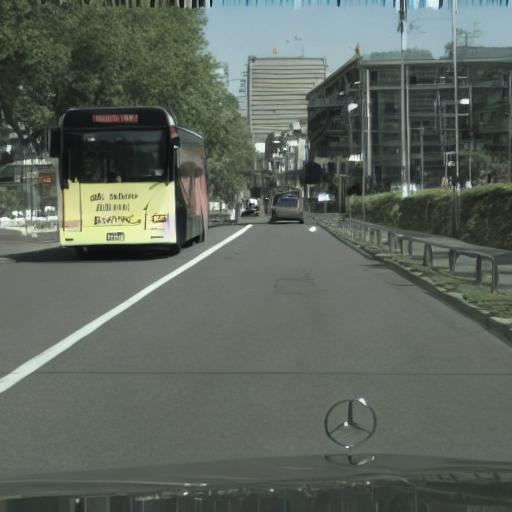}};
  \node[right=\PCACompImageHSep of descriptor3, anchor=west] (descriptor4) {\includegraphics[width=\PCACompImageWidth\linewidth, trim={1cm 5cm 5cm 1cm},clip]{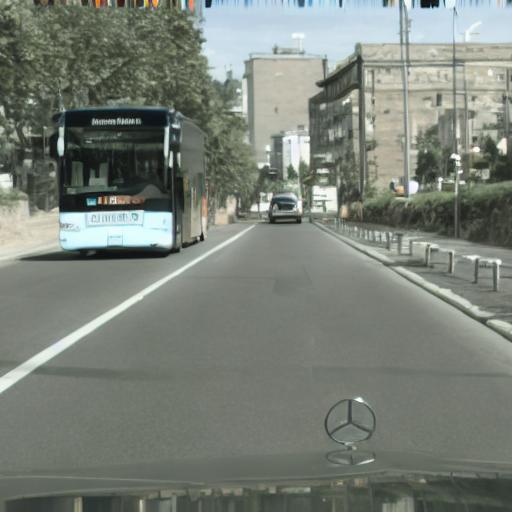}};
  \node[left=\TeaserTitleHSepTeaser of descriptor1, anchor=south, rotate=90] {\tiny{PCA: 1}};

  \node[right=\PCACompImageVSepLarge of descriptor4] (descriptor1_2) {\includegraphics[ width=\PCACompImageWidth\linewidth, trim={1cm 5cm 5cm 1cm},clip]{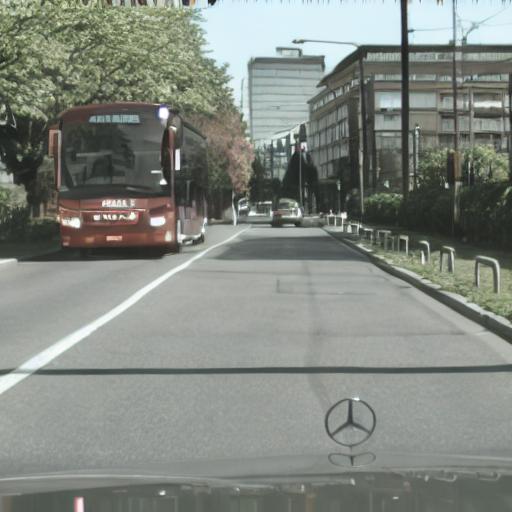}};
  \node[right=\PCACompImageHSep of descriptor1_2, anchor=west] (descriptor2_2) {\includegraphics[width=\PCACompImageWidth\linewidth, trim={1cm 5cm 5cm 1cm},clip]{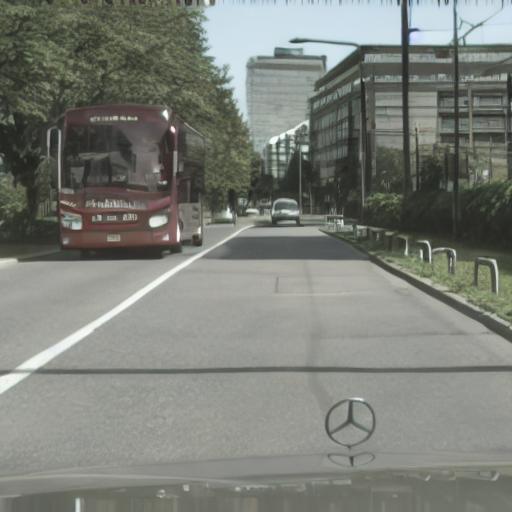}};
  \node[right=\PCACompImageHSep of descriptor2_2, anchor=west] (descriptor3_2) {\includegraphics[width=\PCACompImageWidth\linewidth, trim={1cm 5cm 5cm 1cm},clip]{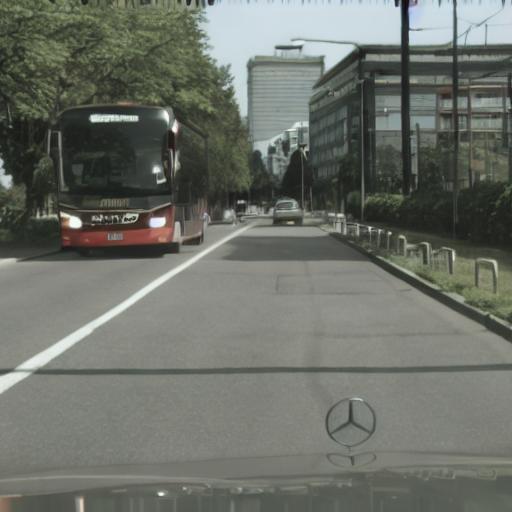}};
  \node[right=\PCACompImageHSep of descriptor3_2, anchor=west] (descriptor4_2) {\includegraphics[width=\PCACompImageWidth\linewidth, trim={1cm 5cm 5cm 1cm},clip]{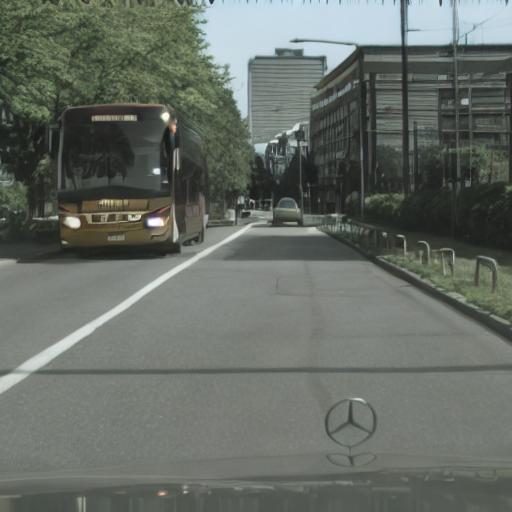}};
  \node[left=\TeaserTitleHSepTeaser of descriptor1_2, anchor=south, rotate=90] {\tiny{PCA: 20}};


  \node[below=\PCACompImageVSep of descriptor1] (sample1) {\includegraphics[width=\PCACompImageWidth\linewidth, trim={1cm 5cm 5cm 1cm},clip]{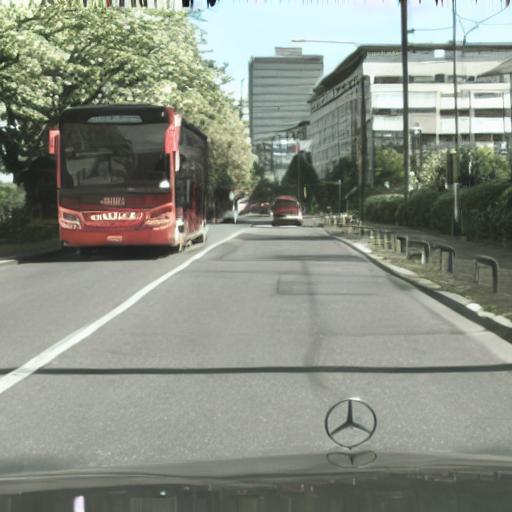}};
  \node[right=\PCACompImageHSep of sample1, anchor=west] (sample2) {\includegraphics[width=\PCACompImageWidth\linewidth, trim={1cm 5cm 5cm 1cm},clip]{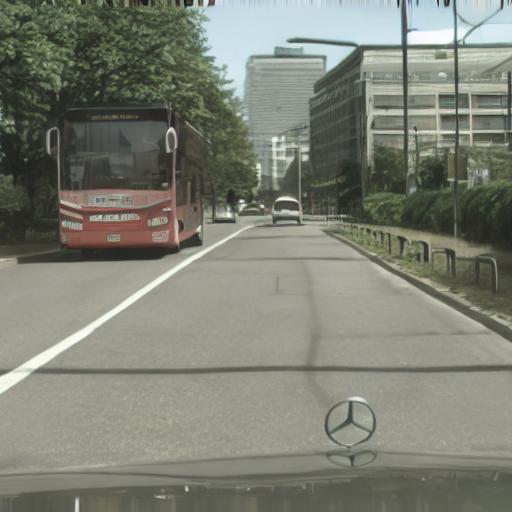}};
  \node[right=\PCACompImageHSep of sample2, anchor=west] (sample3) {\includegraphics[width=\PCACompImageWidth\linewidth, trim={1cm 5cm 5cm 1cm},clip]{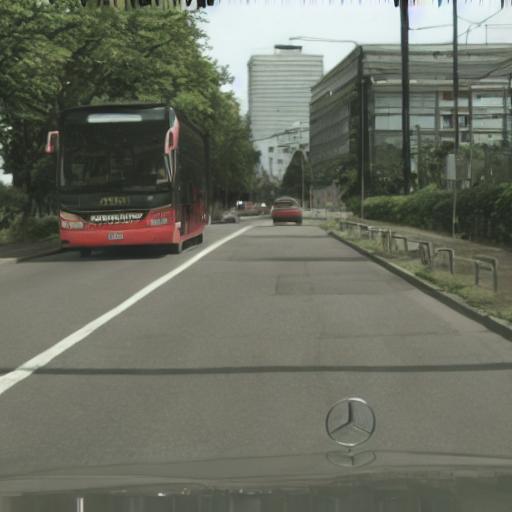}};
  \node[right=\PCACompImageHSep of sample3, anchor=west] (sample4) {\includegraphics[width=\PCACompImageWidth\linewidth, trim={1cm 5cm 5cm 1cm},clip]{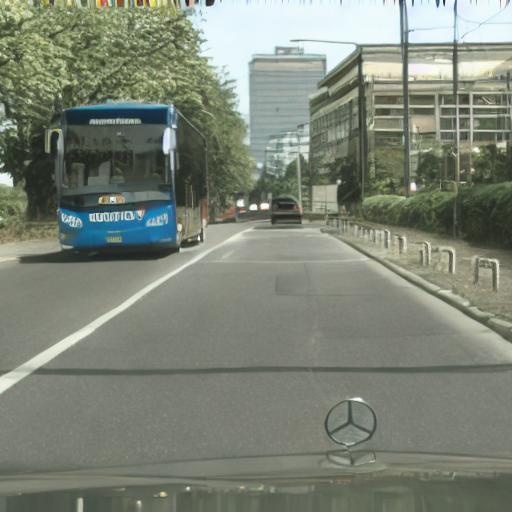}};
  \node[left=\TeaserTitleHSepTeaser of sample1, anchor=south, rotate=90] {\tiny{PCA: 10}};

  \node[below=\PCACompImageVSep of descriptor1_2] (sample1_2) {\includegraphics[width=\PCACompImageWidth\linewidth, trim={1cm 5cm 5cm 1cm},clip]{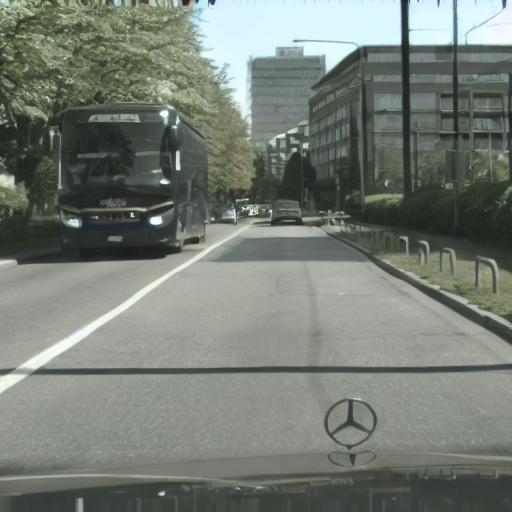}};
  \node[right=\PCACompImageHSep of sample1_2, anchor=west] (sample2_2) {\includegraphics[width=\PCACompImageWidth\linewidth, trim={1cm 5cm 5cm 1cm},clip]{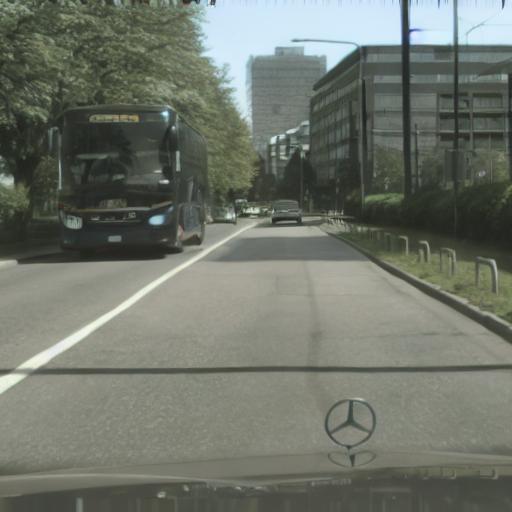}};
  \node[right=\PCACompImageHSep of sample2_2, anchor=west] (sample3_2) {\includegraphics[width=\PCACompImageWidth\linewidth, trim={1cm 5cm 5cm 1cm},clip]{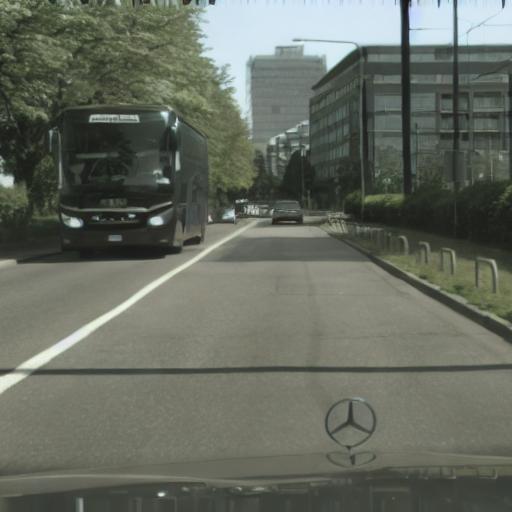}};
  \node[right=\PCACompImageHSep of sample3_2, anchor=west] (sample4_2) {\includegraphics[width=\PCACompImageWidth\linewidth, trim={1cm 5cm 5cm 1cm},clip]{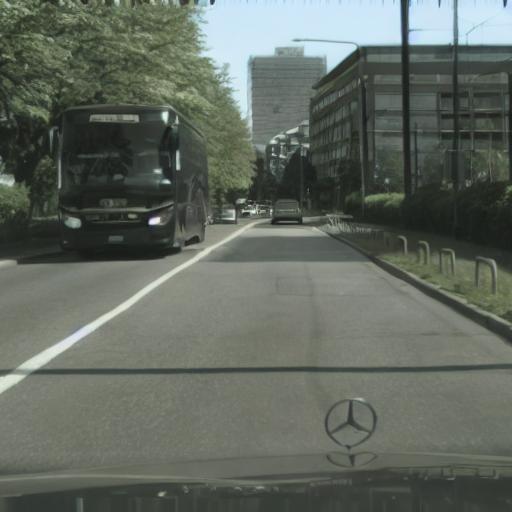}};
  \node[left=\TeaserTitleHSepTeaser of sample1_2, anchor=south, rotate=90] {\tiny{PCA: 100}};
  \end{tikzpicture}
\caption{
Visual comparison of four random samples drawn using {\ourdm} trained with a different number of PCA components. Using fewer trades fidelity for diversity. 
}
\label{fig:PCA-impact}

\end{figure}
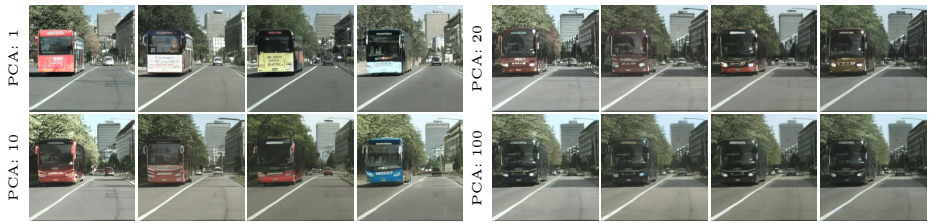

These results demonstrate that the number of components utilized in the neural layout provides an intuitive mechanism to balance the trade-off between faithfulness, diversity, and text responsiveness. 
This level of control is absent in most other layout conditions.

\myparagraph{LUMEN.}
Given these results, we design~\ourdm~to use neural layout extracted from SD features using $N=10$ and $N=20$ PCA components, as this result in a satisfactory range of diversity while remaining faithful to the reference image.
However, this can be further fine-tuned to suit specific downstream tasks.

\begin{table}[t]
\begin{center}
\resizebox{0.97\textwidth}{!}{
\begin{tabular}{l|c|ccccc|ccccc} 
 & & \multicolumn{5}{c|}{COCO-Stuff} & \multicolumn{5}{c}{ADE20K} \\
Conditioning & Pix. Annot. & mIoU $\uparrow$ & SI Depth $\downarrow$ & FID $\downarrow$  & LPIPS $\uparrow$ & TIFA $\uparrow$ & mIoU $\uparrow$ & SI Depth $\downarrow$ & FID $\downarrow$  & LPIPS $\uparrow$ & TIFA $\uparrow$ \\
\hline
\hline
Edges (Canny) & \color{PineGreen}No & 44.4 & 24.7 & 13.2 &  0.48 & 0.84 & 35.1 & 25.0 & 19.1 & 0.49 & 0.86\\
Edges (HED) & \color{orange}Indirect & 49.3 & 21.4 & 12.1 & 0.39 & 0.74 & 41.8 & \textbf{21.1} & 17.4 & 0.37 & 0.73\\
Depth (MiDaS) & \color{orange}Indirect &  45.3 & 24.0 & 14.3 &  0.53 & \textbf{0.88} & 34.0 & 22.1 & 21.2 & 0.52 & \textbf{0.87}\\
Sem. Seg. (pred) & \color{orange}Indirect & 42.9 & 27.7 & 16.0 & 0.64 & \textbf{0.88} & 33.0 & 24.6 & 23.4 & 0.62 & 0.85\\
Sem. Seg. (GT) & \color{red}Yes & 43.3 & 28.8 & 15.3 & \textbf{0.65} & \textbf{0.88} & 35.1 & 27.1 & 22.6 & \textbf{0.63 }& 0.85\\
\textbf{Neural Layout}& \color{PineGreen}No & \textbf{52.9} & \textbf{21.1} & \textbf{11.8}  & 0.36 & 0.66 & \textbf{45.7} & \textbf{21.1} & \textbf{16.1} & 0.33 & 0.67 
\vspace{-1em}
\end{tabular}   
}
\end{center} 
\caption{
Comparison of using different conditioning options for ControlNet across datasets. Neural layout outperforms all conditioning options in terms of image quality (FID), as well as semantic and spatial alignment (mIoU, SI).
} 
\label{table:comparisonOtherData}
\vspace{-2.5em}
\end{table}

\subsection{Comparison to Existing Conditioning}
\label{sec:comparison}
We compare \ourdm~ using neural layouts with $N=20$ PCA components against different established image conditioning.
For this, we consider the common ControlNet condition: Canny edges\newcite{canny1986computational} (Canny), HED edges\newcite{xie15hed} (HED), MiDaS depth\newcite{Ranftl2022midas} (MiDaS),  predicted semantic segmentation (Sem. Seg. pred), and ground truth semantic segmentation (Sem. Seg. GT). 
For Sem. Seg. (pred) conditioning, we use the same pretrained network as the one used for evaluating the mIoU metric.

It is important to first discuss the relative cost of obtaining the conditioning input for each of the types.
Sem. Seg. (GT) requires expensive pixel-level annotations making it difficult to scale the generator training.
HED, MiDaS, and Sem. Seg. (pred) can be cheaply obtained through pretrained estimators.
But the estimators themselves require training and thus indirectly require pixel-wise annotation.
In contrast, neural layout do not require any pixel-wise annotation as it is either weakly supervised with more affordable image captions or is entirely self-supervised depending on the backbone.
This distinction becomes crucial for applications with domain shifts where the feature extraction network itself need to be fine-tuned. 
The only other conditioning input that fulfil this criteria is Canny edges.
These requirements for obtaining the conditioning is summarized in~\cref{table:comparisonOtherData} (Pix. Annot.).

\newcommand{\DescriptorCompImageWidth}{0.145}
\newcommand{\DescriptorCompImageHSep}{0.05cm}
\newcommand{\DescriptorCompImageHSepLarge}{0.5cm}
\newcommand{\DescriptorCompImageHSepH}{0.025cm}
\newcommand{\DescriptorCompImageVSep}{0.05cm}
\newcommand{\DescriptorCompImageBlockHSep}{0.2cm}
\newcommand{\DescriptorCompImageBlockVSep}{0.2cm}
\newcommand{\DescriptorCompTitleHSep}{0.15cm}
\newcommand{\DescriptorCompTitleVSep}{0.35cm}
\newcommand{\DescriptorCompTitleHSepDescriptorComp}{0.15cm}

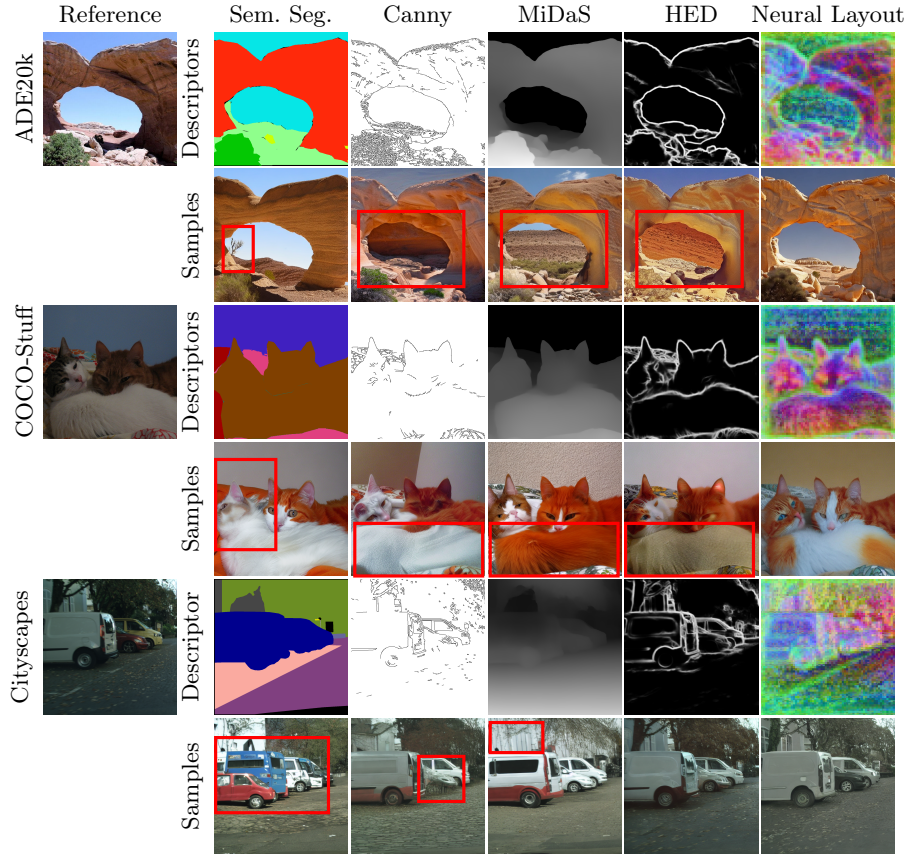
\begin{figure*}[t]
\vspace{-0.5em}
    \centering
  \begin{tikzpicture}[every node/.style={inner sep=0,outer sep=0}]%

  \node (Input1) at (0,0) {\includegraphics[ width=\DescriptorCompImageWidth\linewidth]{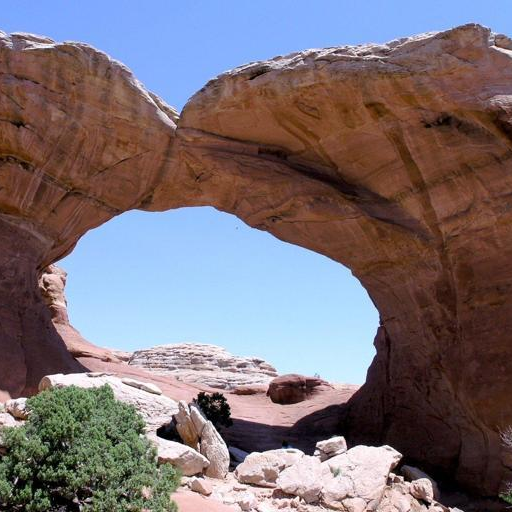}};
  \node[left=\TeaserTitleHSepTeaser of Input1, anchor=south, rotate=90] {\small{ADE20k}};

  \node[right=\DescriptorCompImageHSepLarge of Input1, anchor=west] (descriptor1) {\includegraphics[ width=\DescriptorCompImageWidth\linewidth]{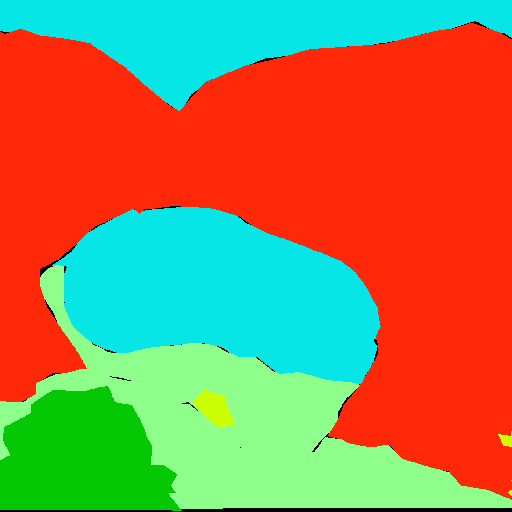}};
  \node[right=\DescriptorCompImageHSep of descriptor1, anchor=west] (descriptor2) {\includegraphics[width=\DescriptorCompImageWidth\linewidth]{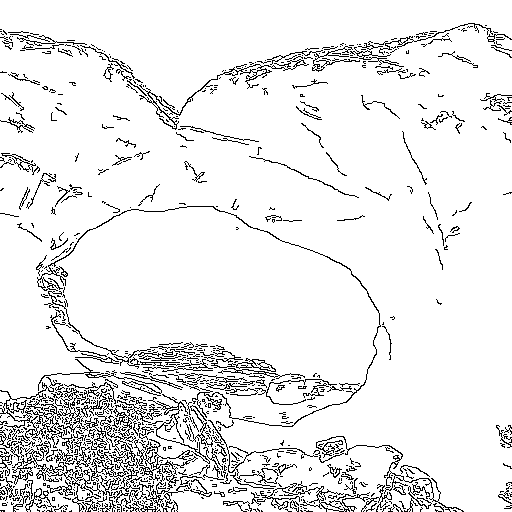}};
  \node[right=\DescriptorCompImageHSep of descriptor2, anchor=west] (descriptor3) {\includegraphics[width=\DescriptorCompImageWidth\linewidth]{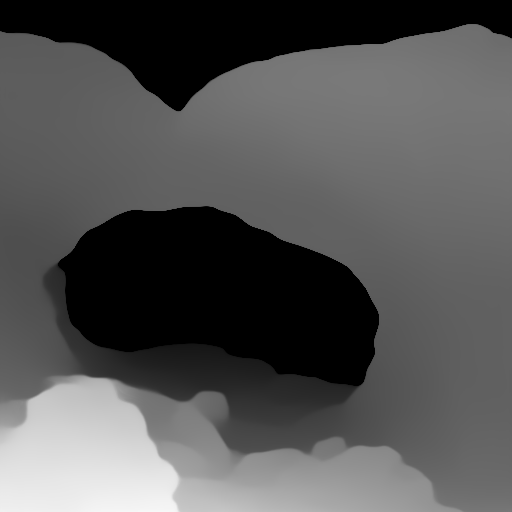}};
  \node[right=\DescriptorCompImageHSep of descriptor3, anchor=west] (descriptor4) {\includegraphics[width=\DescriptorCompImageWidth\linewidth]{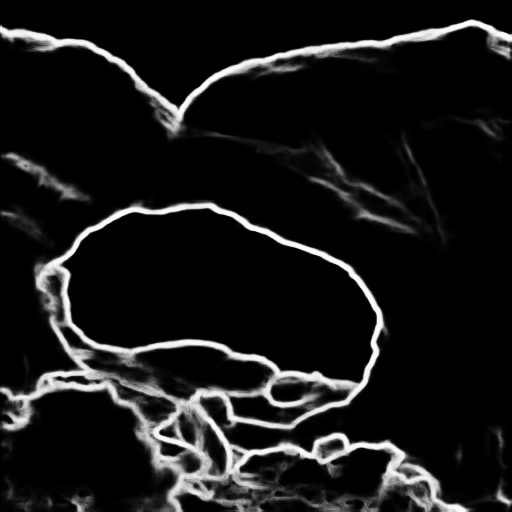}};
  \node[right=\DescriptorCompImageHSep of descriptor4, anchor=west] (descriptor5) {\includegraphics[width=\DescriptorCompImageWidth\linewidth]{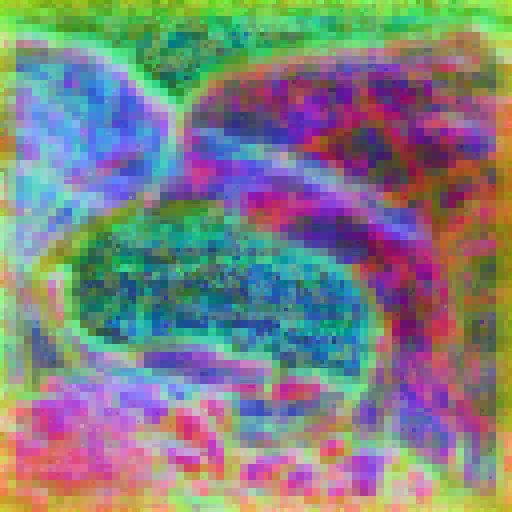}};
  \node[left=\TeaserTitleHSepTeaser of descriptor1, anchor=south, rotate=90] {\small{Descriptors}};

   \node[above=\DescriptorCompTitleVSep of Input1, anchor=north] {\small{Reference}};
  \node[above=\DescriptorCompTitleVSep of descriptor1, anchor=north] {\small{Sem. Seg.}};
  \node[above=\DescriptorCompTitleVSep of descriptor2, anchor=north] 
  {\small{Canny}};
  \node[above=\DescriptorCompTitleVSep of descriptor3, anchor=north] {\small{MiDaS}};
  \node[above=\DescriptorCompTitleVSep of descriptor4, anchor=north] {\small{HED}};
  \node[above=\DescriptorCompTitleVSep of descriptor5, anchor=north] {\small{Neural Layout}};

  \node[below=\DescriptorCompImageVSep of descriptor1] (sample1) {\includegraphics[width=\DescriptorCompImageWidth\linewidth]{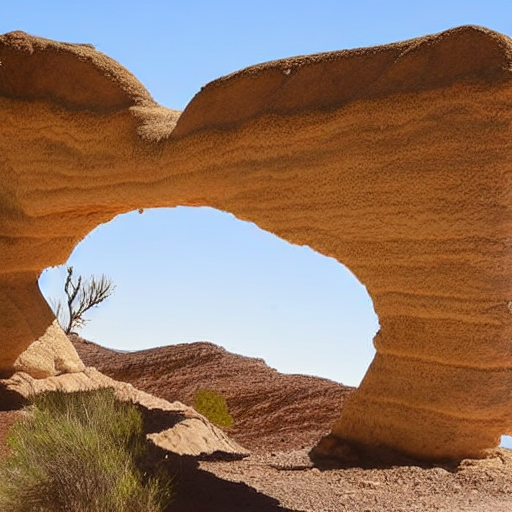}};
  \node[right=\DescriptorCompImageHSep of sample1, anchor=west] (sample2) {\includegraphics[width=\DescriptorCompImageWidth\linewidth]{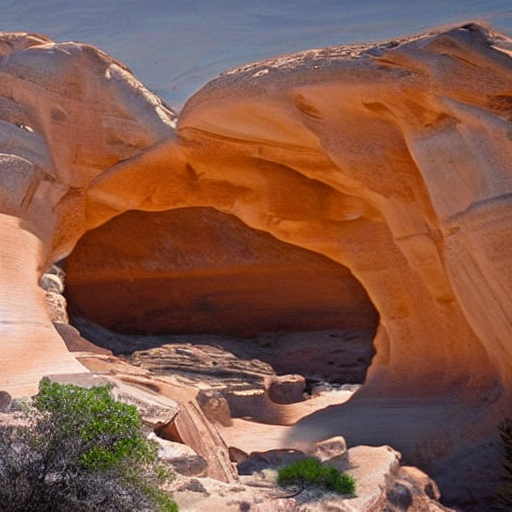}};
  \node[right=\DescriptorCompImageHSep of sample2, anchor=west] (sample3) {\includegraphics[width=\DescriptorCompImageWidth\linewidth]{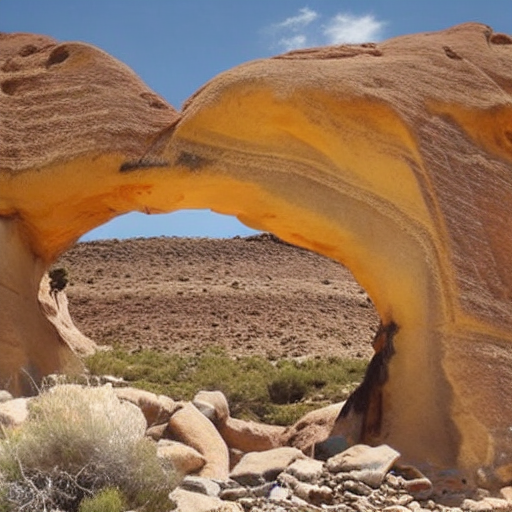}};
  \node[right=\DescriptorCompImageHSep of sample3, anchor=west] (sample4) {\includegraphics[width=\DescriptorCompImageWidth\linewidth]{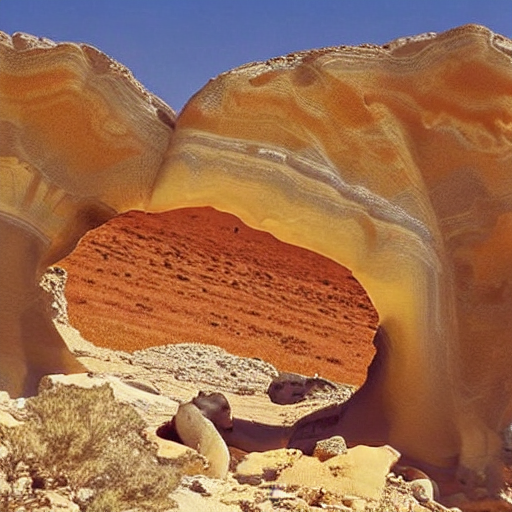}};
  \node[right=\DescriptorCompImageHSep of sample4, anchor=west] (sample5) {\includegraphics[width=\DescriptorCompImageWidth\linewidth]{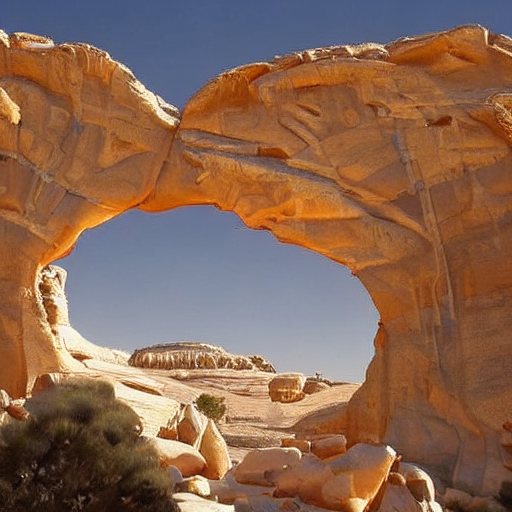}};
  \node[left=\TeaserTitleHSepTeaser of sample1, anchor=south, rotate=90] {\small{Samples}};

  \node[below=\DescriptorCompImageVSep of sample1] (descriptor1_2) {\includegraphics[ width=\DescriptorCompImageWidth\linewidth]{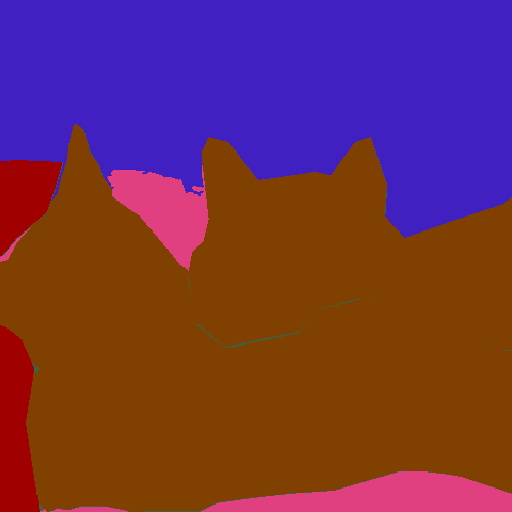}};
  \node[right=\DescriptorCompImageHSep of descriptor1_2, anchor=west] (descriptor2_2) {\includegraphics[width=\DescriptorCompImageWidth\linewidth]{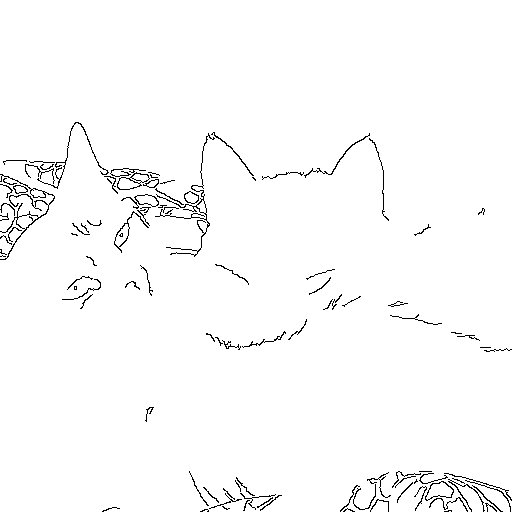}};
  \node[right=\DescriptorCompImageHSep of descriptor2_2, anchor=west] (descriptor3_2) {\includegraphics[width=\DescriptorCompImageWidth\linewidth]{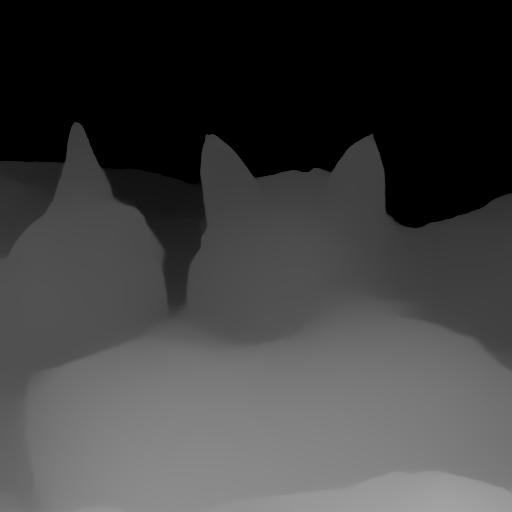}};
  \node[right=\DescriptorCompImageHSep of descriptor3_2, anchor=west] (descriptor4_2) {\includegraphics[width=\DescriptorCompImageWidth\linewidth]{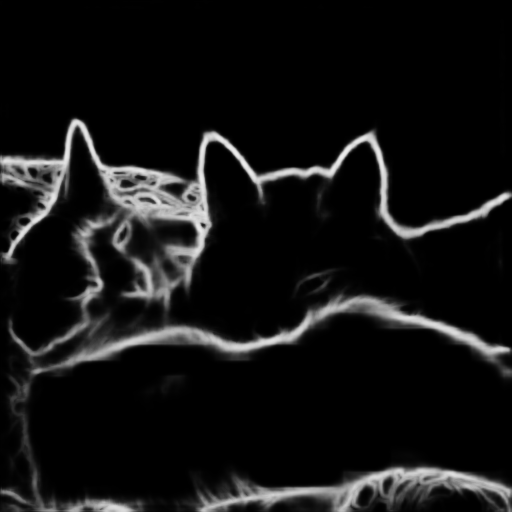}};
  \node[right=\DescriptorCompImageHSep of descriptor4_2, anchor=west] (descriptor5_2) {\includegraphics[width=\DescriptorCompImageWidth\linewidth]{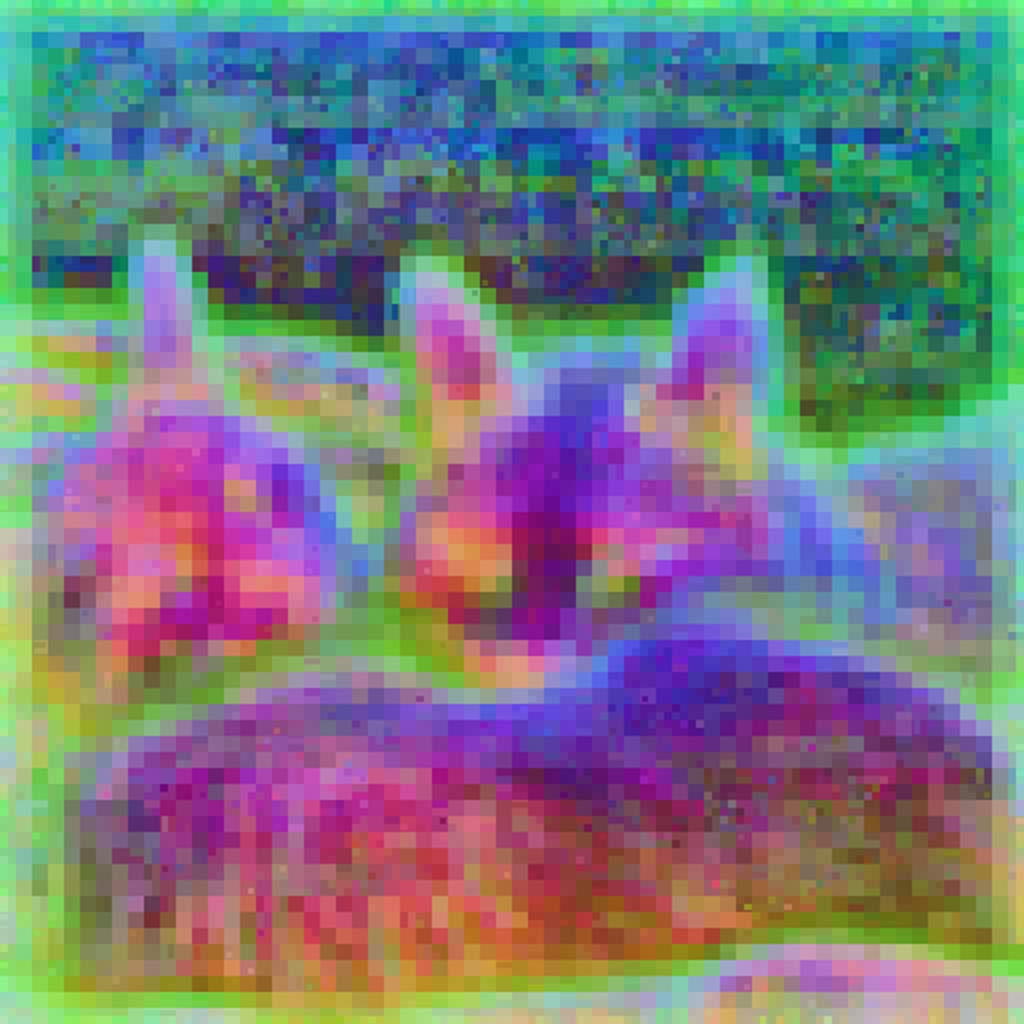}};
  \node[left=\TeaserTitleHSepTeaser of descriptor1_2, anchor=south, rotate=90] {\small{Descriptors}};

  \node[left=\DescriptorCompImageHSepLarge of descriptor1_2]  (Input2) {\includegraphics[ width=\DescriptorCompImageWidth\linewidth]{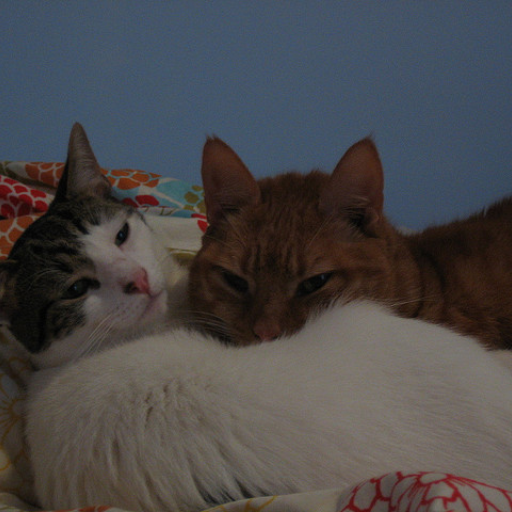}};
  \node[left=\TeaserTitleHSepTeaser of Input2, anchor=south, rotate=90] {\small{COCO-Stuff}};
  
  \node[below=\DescriptorCompImageVSep of descriptor1_2] (sample1_2) {\includegraphics[width=\DescriptorCompImageWidth\linewidth]{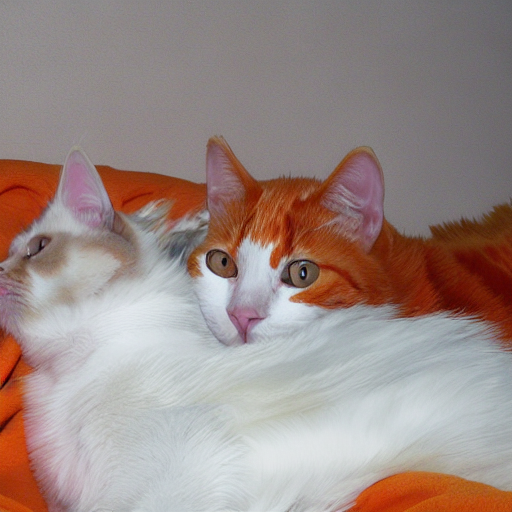}};
  \node[right=\DescriptorCompImageHSep of sample1_2, anchor=west] (sample2_2) {\includegraphics[width=\DescriptorCompImageWidth\linewidth]{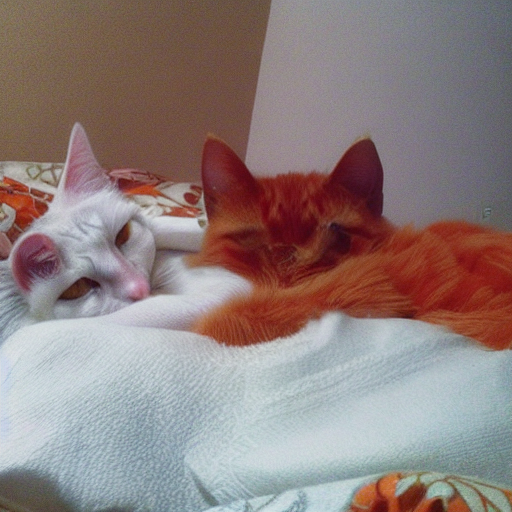}};
  \node[right=\DescriptorCompImageHSep of sample2_2, anchor=west] (sample3_2) {\includegraphics[width=\DescriptorCompImageWidth\linewidth]{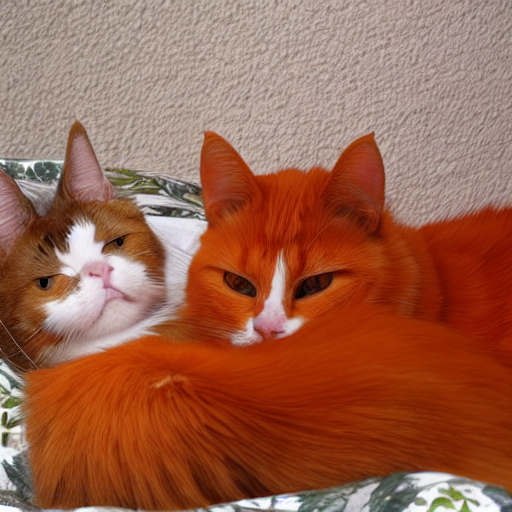}};
  \node[right=\DescriptorCompImageHSep of sample3_2, anchor=west] (sample4_2) {\includegraphics[width=\DescriptorCompImageWidth\linewidth]{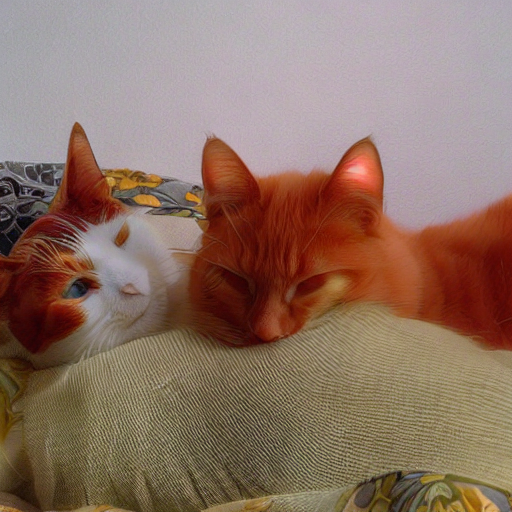}};
  \node[right=\DescriptorCompImageHSep of sample4_2, anchor=west] (sample5) {\includegraphics[width=\DescriptorCompImageWidth\linewidth]{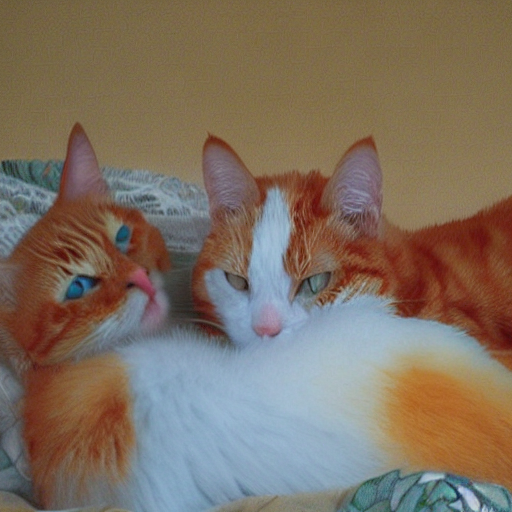}};
  \node[left=\TeaserTitleHSepTeaser of sample1_2, anchor=south, rotate=90] {\small{Samples}};

  \node[below=\DescriptorCompImageVSep of sample1_2] (Descriptor1) {\includegraphics[ width=\DescriptorCompImageWidth\linewidth, trim={0cm 0cm 36.8cm 0cm}, clip]{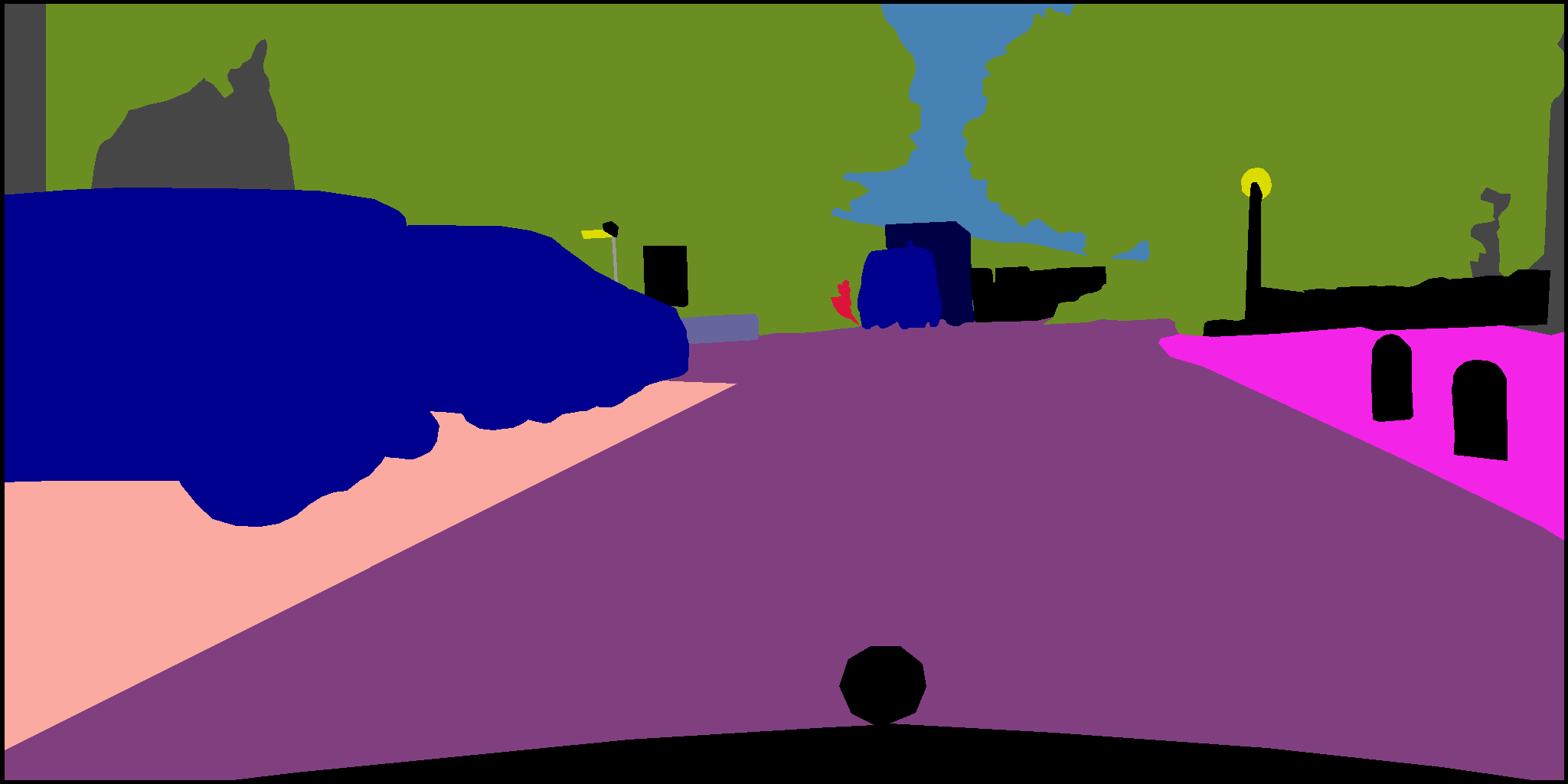}};
  \node[right=\DescriptorCompImageHSep of Descriptor1, anchor=west] (Descriptor2) {\includegraphics[width=\DescriptorCompImageWidth\linewidth, trim={0cm 0cm 18.4cm 0cm}, clip]{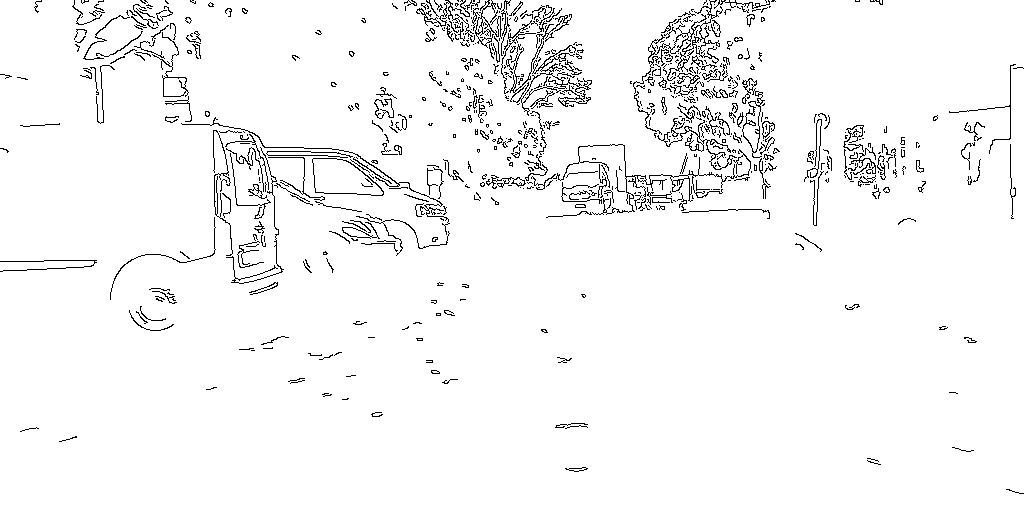}};
  \node[right=\DescriptorCompImageHSep of Descriptor2, anchor=west] (Descriptor3) {\includegraphics[width=\DescriptorCompImageWidth\linewidth, trim={0cm 0cm 18.4cm 0cm}, clip]{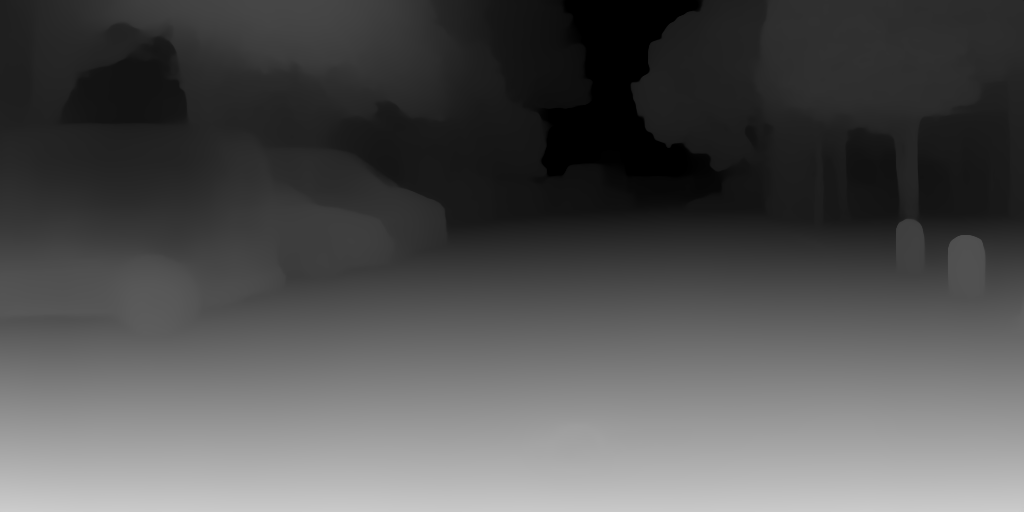}};
  \node[right=\DescriptorCompImageHSep of Descriptor3, anchor=west] (Descriptor4) {\includegraphics[width=\DescriptorCompImageWidth\linewidth, trim={0cm 0cm 18.4cm 0cm}, clip]{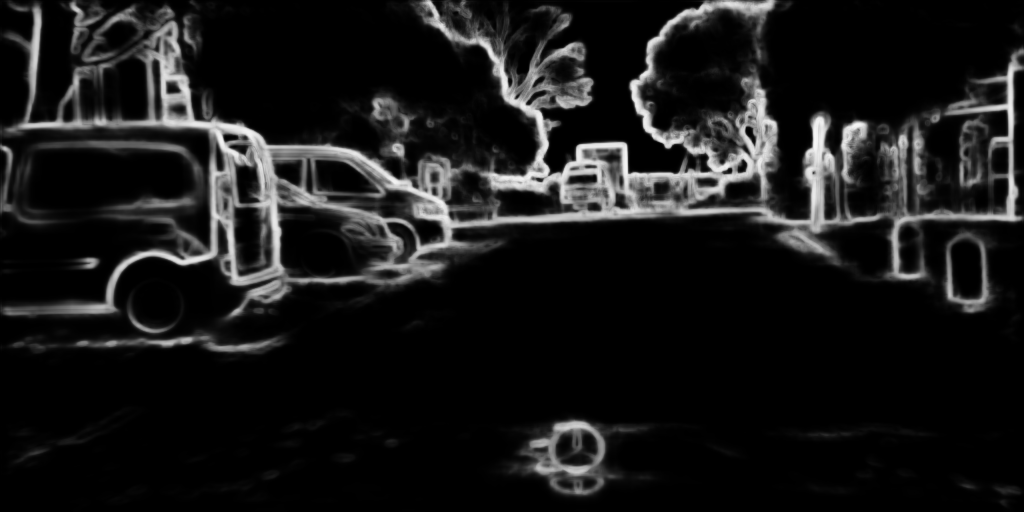}};
  \node[right=\DescriptorCompImageHSep of Descriptor4, anchor=west] (Descriptor5) {\includegraphics[width=\DescriptorCompImageWidth\linewidth, trim={0cm 0cm 36.8cm 0cm}, clip]{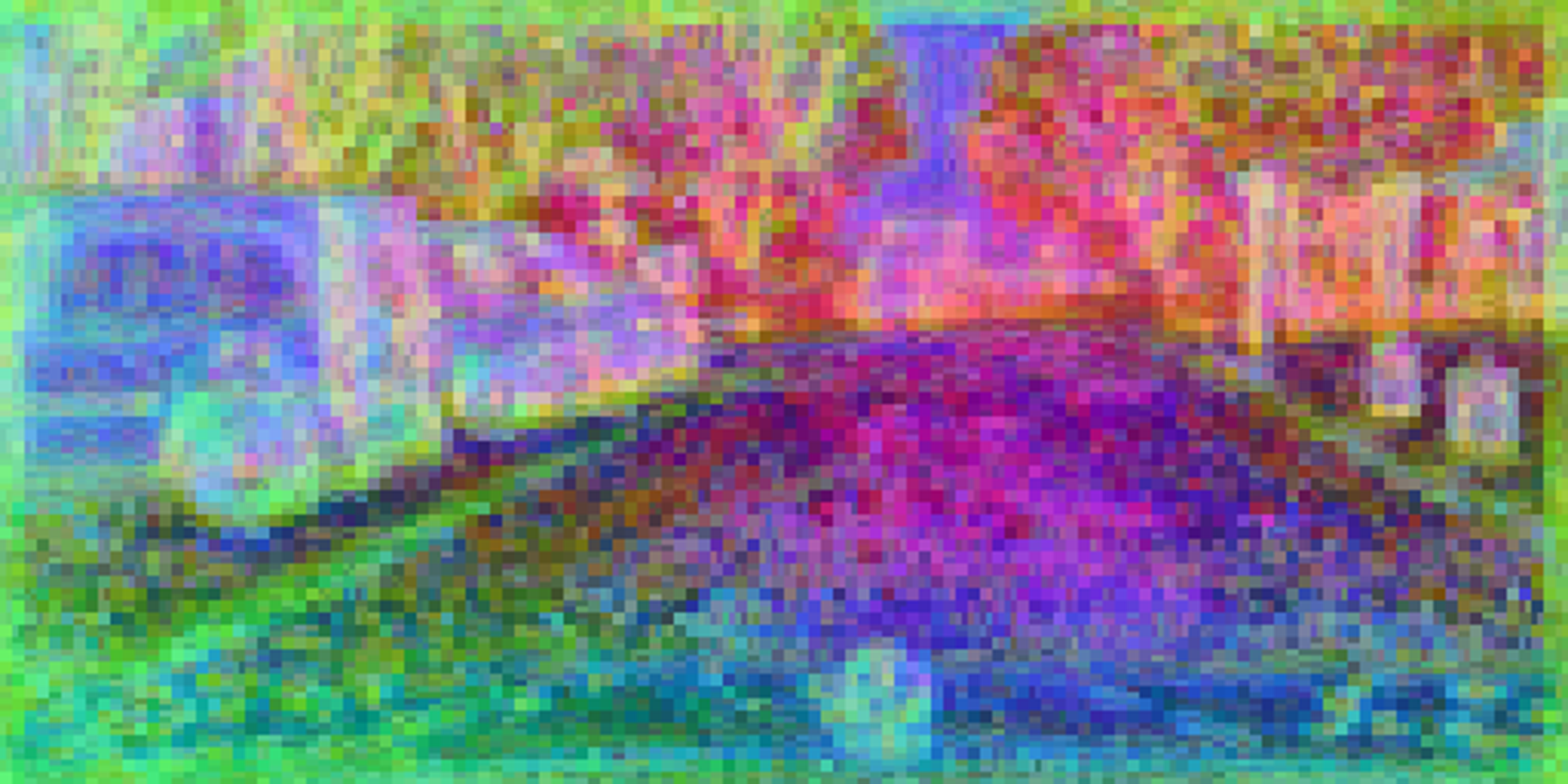}};
  \node[left=\DescriptorCompTitleHSep of Descriptor1, anchor=south, rotate=90] {\small{Descriptor}};

  
  \node[left=\DescriptorCompImageHSepLarge of Descriptor1] (Input3) {\includegraphics[width=\DescriptorCompImageWidth\linewidth, trim={0cm 0cm 18.4cm 0cm}, clip]{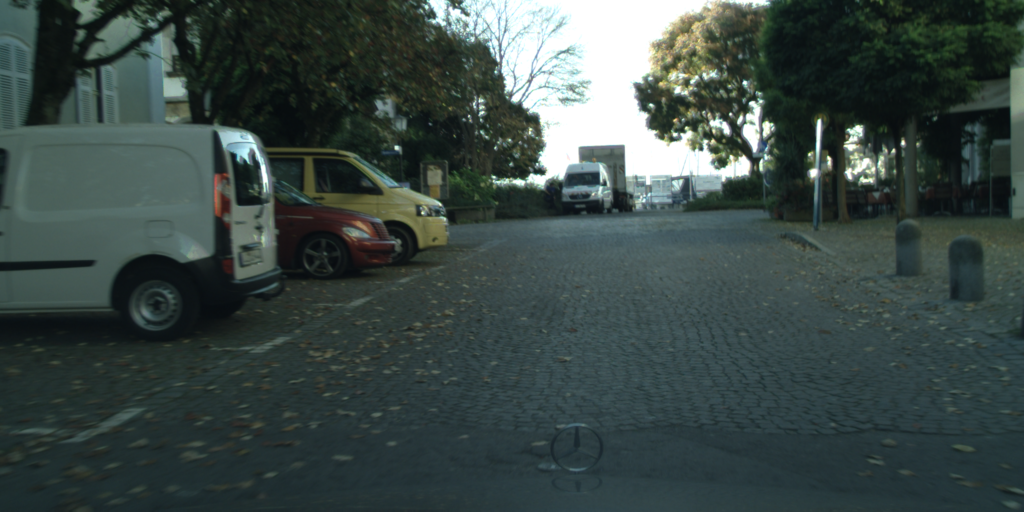}};
  \node[left=\DescriptorCompTitleHSep of Input3, anchor=south, rotate=90] {\small{Cityscapes}};

  \node[below=\DescriptorCompImageVSep of Descriptor1] (sample1) {\includegraphics[width=\DescriptorCompImageWidth\linewidth, trim={0cm 0cm 18.4cm 0cm}, clip]{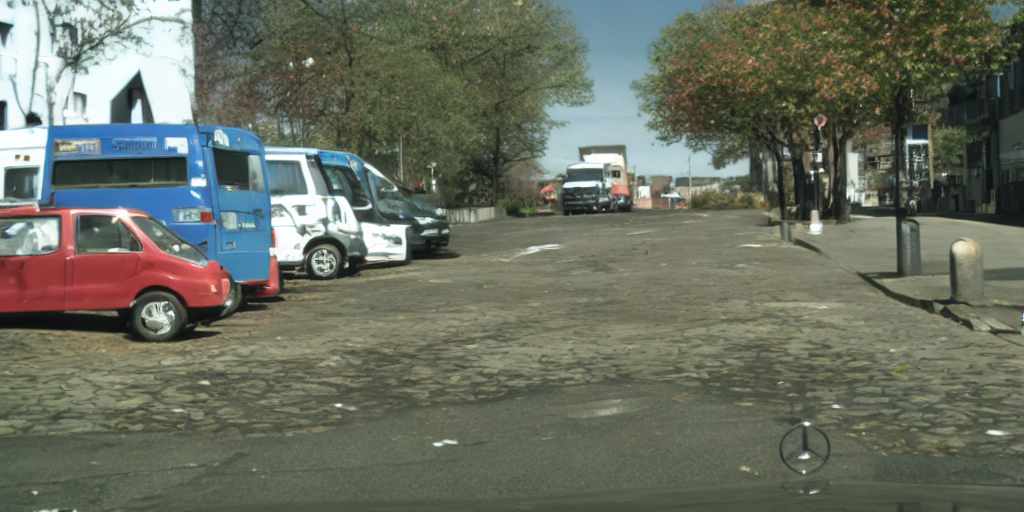}};
  \node[right=\DescriptorCompImageHSep of sample1, anchor=west] (sample2) {\includegraphics[width=\DescriptorCompImageWidth\linewidth, trim={0cm 0cm 18.4cm 0cm}, clip]{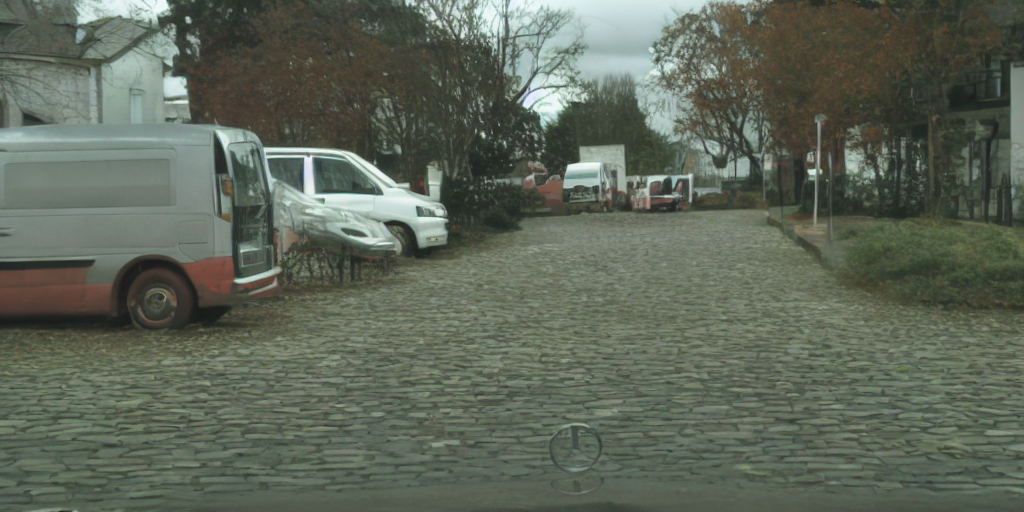}};
  \node[right=\DescriptorCompImageHSep of sample2, anchor=west] (sample3) {\includegraphics[width=\DescriptorCompImageWidth\linewidth, trim={0cm 0cm 18.4cm 0cm}, clip]{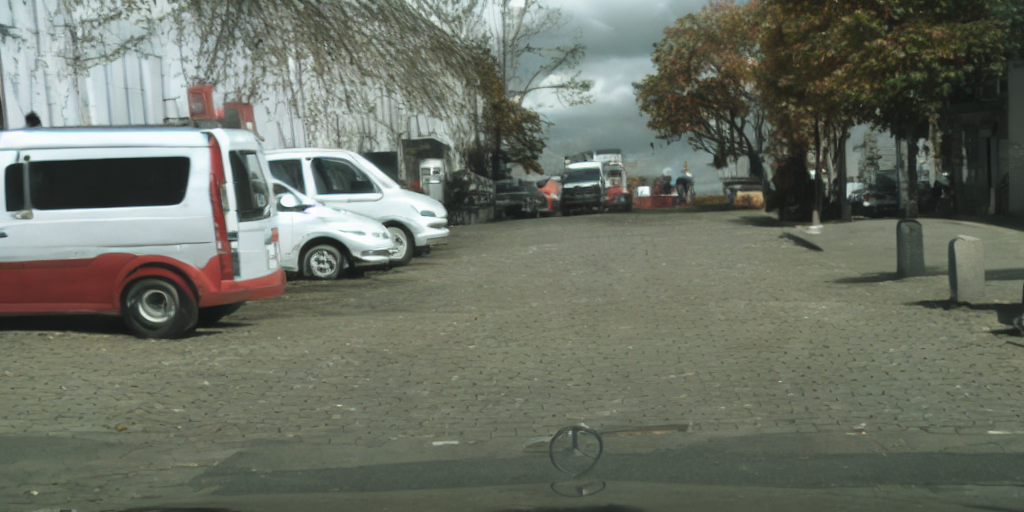}};
  \node[right=\DescriptorCompImageHSep of sample3, anchor=west] (sample4) {\includegraphics[width=\DescriptorCompImageWidth\linewidth, trim={0cm 0cm 18.4cm 0cm}, clip]{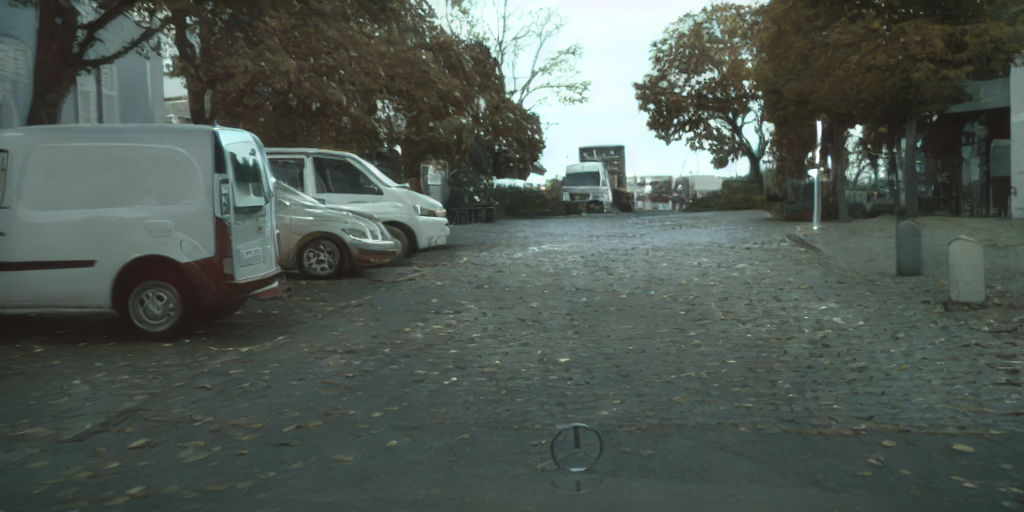}};
  \node[right=\DescriptorCompImageHSep of sample4, anchor=west] (sample5) {\includegraphics[width=\DescriptorCompImageWidth\linewidth, trim={0cm 0cm 18.4cm 0cm}, clip]{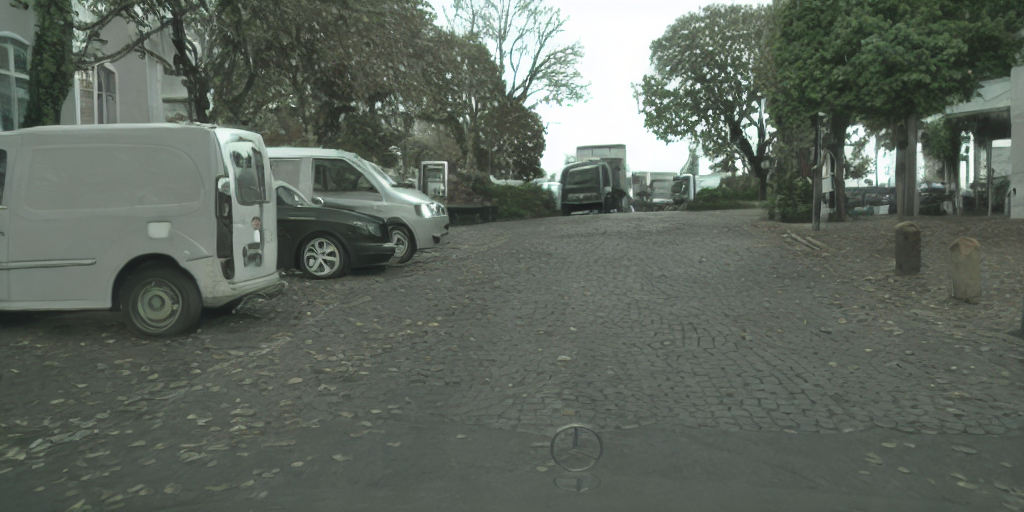}};
  \node[left=\DescriptorCompTitleHSep of sample1, anchor=south, rotate=90] {\small{Samples}};

  \node (rect) [rectangle,draw,very thick,red,minimum width=0.4cm,minimum height=0.6cm] at (1.7, -2.0) (){};
  \node (rect) [rectangle,draw,very thick,red,minimum width=1.4cm,minimum height=1.0cm] at (4.0, -2.0) (){};
  \node (rect) [rectangle,draw,very thick,red,minimum width=1.4cm,minimum height=1.0cm] at (5.9, -2.0) (){};
  \node (rect) [rectangle,draw,very thick,red,minimum width=1.4cm,minimum height=1.0cm] at (7.7, -2.0) (){};

  \node (rect) [rectangle,draw,very thick,red,minimum width=.8cm,minimum height=1.2cm] at (1.8, -5.4) (){};
  \node (rect) [rectangle,draw,very thick,red,minimum width=1.7cm,minimum height=0.7cm] at (4.1, -6.0) (){};
  \node (rect) [rectangle,draw,very thick,red,minimum width=1.7cm,minimum height=0.7cm] at (5.9, -6.0) (){};
  \node (rect) [rectangle,draw,very thick,red,minimum width=1.7cm,minimum height=0.7cm] at (7.7, -6.0) (){};

  \node (rect) [rectangle,draw,very thick,red,minimum width=1.5cm,minimum height=1.0cm] at (2.15, -9.0) (){};
  \node (rect) [rectangle,draw,very thick,red,minimum width=0.6cm,minimum height=0.6cm] at (4.4, -9.05) (){};
  \node (rect) [rectangle,draw,very thick,red,minimum width=0.7cm,minimum height=0.4cm] at (5.4, -8.5) (){};

  \end{tikzpicture}
  \vspace{-0.5em}
\caption{
Comparison of images generated with different conditioning types on ADE20k, COCO-Stuff and Cityscapes. Neural layouts provide rich description of the desired images, while other inputs contain limited information and are semantically ambiguous.
}
\label{fig:vis-compare}
\vspace{-2em}
\end{figure*}

\myparagraph{Effect of Conditioning Types.}
We evaluate the effects that different conditioning have on image synthesis using the challenging COCO-Stuff\newcite{caesar2018coco} and ADE20k\newcite{zhou2017ade20k} datasets.
As \cref{table:comparisonOtherData} shows, using neural layout  results in images that best preserve the semantic content, outperforming others in terms of both mIoU and SI Depth while achieving better image quality.
Surprisingly, Sem. Seg. achieves only similar or worse results in terms of mIoU compared to all other conditioning. 
We believe this is due to the large number of difficult semantic classes in COCO-Stuff and ADE20k with sometimes semantically ambiguous labels, and due to the long tail distribution on rare classes.
We also see that HED performs the best among the existing conditions in terms of FID, mIoU, and SI Depth, as it well encodes the object boundaries with additional image details being captured by soft edges.
However, the semantic class of the object within the boundary can be ambiguous, resulting in a lower mIoU  (also see \cref{fig:vis-compare}).

Note that we observe again the trade-off between information content constraining the image and the diversity and editability of the results. 
Canny edge and MiDaS depth have low semantic content and sem. seg. does not constrain appearance or geometry, consequently, they often achieve the best LPIPS and TIFA at the cost of worse alignment or image quality. 
We also observe that despite the lower TIFA, \ourdm~still responds well to a variety of out-of-distribution prompt edits (see \cref{fig:text-prompting}). 
As we will see in \cref{sec:appplication}, \ourdm~provides enough image variation from sample diversity and text editability in practice for the downstream network to achieve the best performance.

\newcommand{\PromptImageWidth}{0.1635} 
\newcommand{\PromptImageVSep}{0.9cm}
\newcommand{\PromptCapImageVSep}{0.25cm}
\newcommand{\PromptTitleVSep}{0.25cm}

\begin{figure*}[t]
    \centering
  \begin{tikzpicture}[every node/.style={inner sep=0,outer sep=0}]%
  
  \node (sample2_1) at (0,0) {\includegraphics[width=\PromptImageWidth\linewidth,trim={2cm 3cm 2cm 1cm},clip]{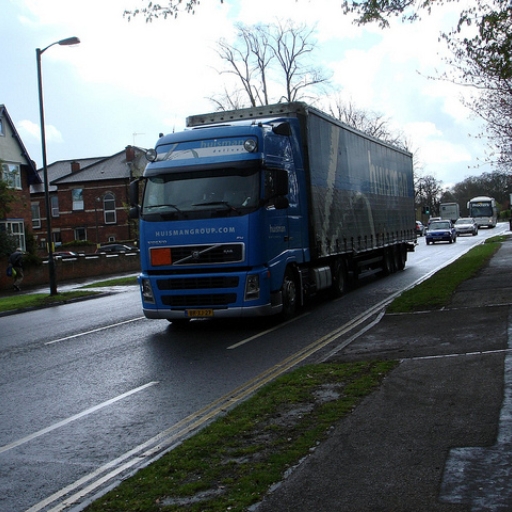}};
\node[below=\PromptCapImageVSep of sample2_1, anchor=south] {\tiny{\phantom{hg}\textbf{Reference}\phantom{hg}}};

  \node[right=\PCACompImageHSep of sample2_1, anchor=west] (sample2_1p5) {\includegraphics[width=\PromptImageWidth\linewidth,trim={2cm 3cm 2cm 1cm},clip]{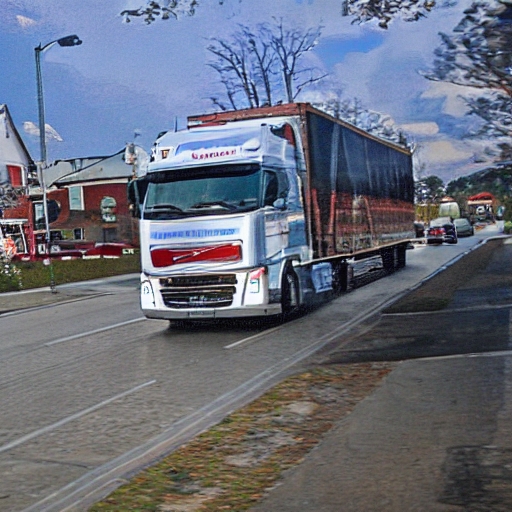}};
  \node[below=\PromptCapImageVSep of sample2_1p5, anchor=south] {\tiny{\phantom{hg}Sample\phantom{hg}}};

  \node[right=\PCACompImageHSep of sample2_1p5, anchor=west] (sample2_2) {\includegraphics[width=\PromptImageWidth\linewidth,trim={2cm 3cm 2cm 1cm},clip]{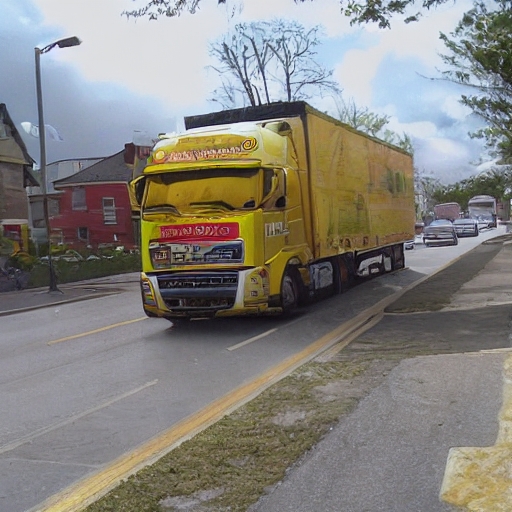}};
\node[below=\PromptCapImageVSep of sample2_2, anchor=south] {\tiny{\phantom{hg}$\rightarrow$ banana truck\phantom{hg}}};
  
  \node[right=\PCACompImageHSep of sample2_2, anchor=west] (sample2_4) {\includegraphics[width=\PromptImageWidth\linewidth,trim={2cm 3cm 2cm 1cm},clip]{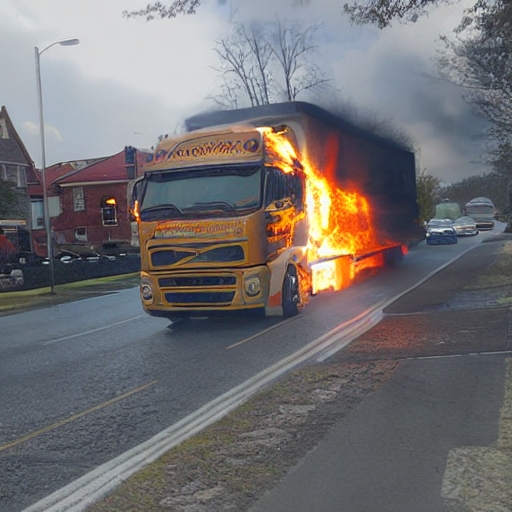}};
\node[below=\PromptCapImageVSep of sample2_4, anchor=south] {\tiny{\phantom{hg}$\rightarrow$ burning truck\phantom{hg}}};
  
  \node[right=\PCACompImageHSep of sample2_4, anchor=west] (sample2_5) {\includegraphics[width=\PromptImageWidth\linewidth,trim={2cm 3cm 2cm 1cm},clip]{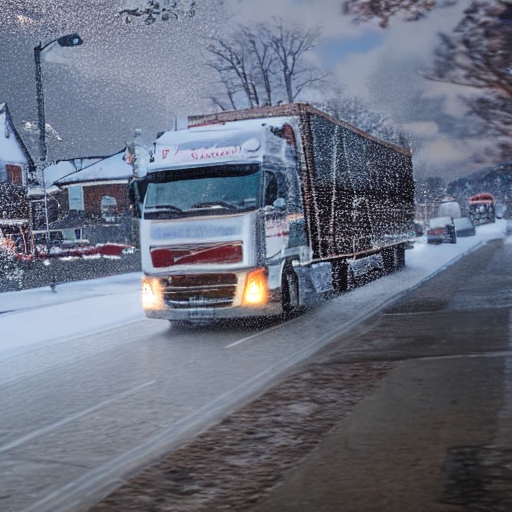}};
  \node[below=\PromptCapImageVSep of sample2_5, anchor=south] {\tiny{\phantom{hg}$+$ winter\phantom{hg}}};
  
  \node[right=\PCACompImageHSep of sample2_5, anchor=west] (sample2_6) {\includegraphics[width=\PromptImageWidth\linewidth,trim={2cm 3cm 2cm 1cm},clip]{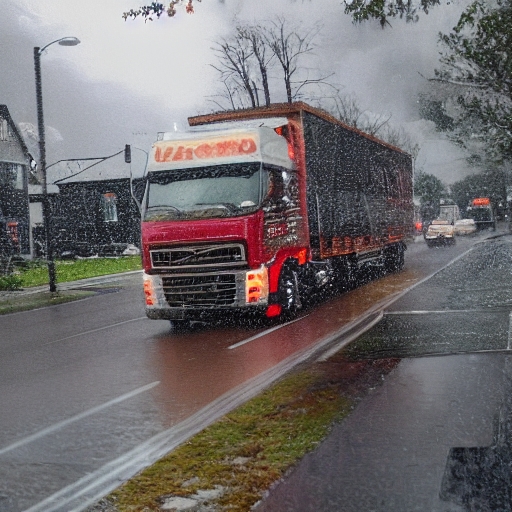}};
  \node[below=\PromptCapImageVSep of sample2_6, anchor=south] {\tiny{\phantom{hg}$+$ in rainy weather\phantom{hg}}};

   \node[above=\PromptTitleVSep of sample2_1, anchor=west] {\hspace{-0.073\linewidth}\small{Original caption: A \underline{semi truck} is driving down a street}.};

  \node[below=\PromptImageVSep of sample2_1] (sample4_1) {\includegraphics[width=\PromptImageWidth\linewidth]{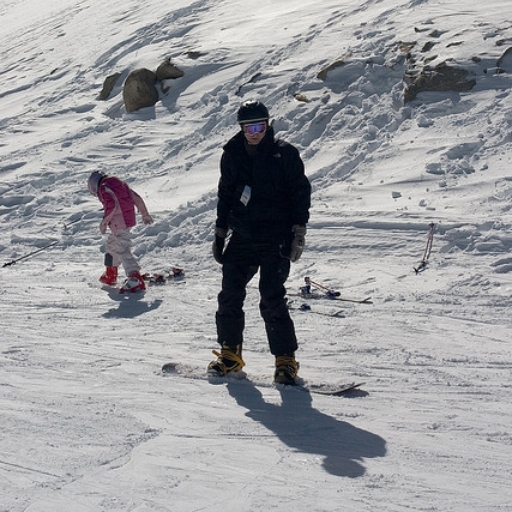}};
\node[below=\PromptCapImageVSep of sample4_1, anchor=south] {\tiny{\phantom{hg}\textbf{Reference}\phantom{hg}}};
  
  \node[right=\PCACompImageHSep of sample4_1, anchor=west] (sample4_1p5) {\includegraphics[width=\PromptImageWidth\linewidth]{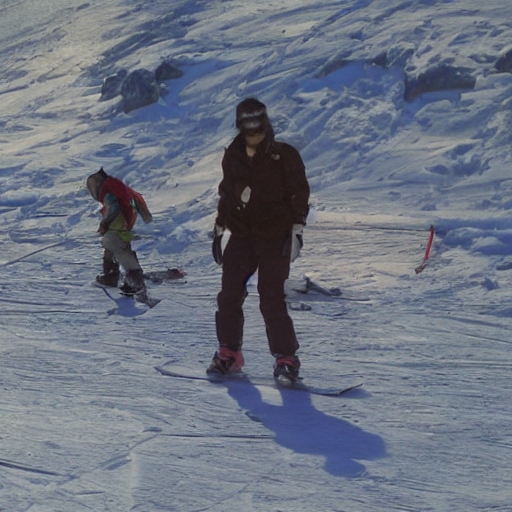}};
  \node[below=\PromptCapImageVSep of sample4_1p5, anchor=south] {\tiny{\phantom{hg}Sample\phantom{hg}}};

  \node[right=\PCACompImageHSep of sample4_1p5, anchor=west] (sample4_2) {\includegraphics[width=\PromptImageWidth\linewidth]{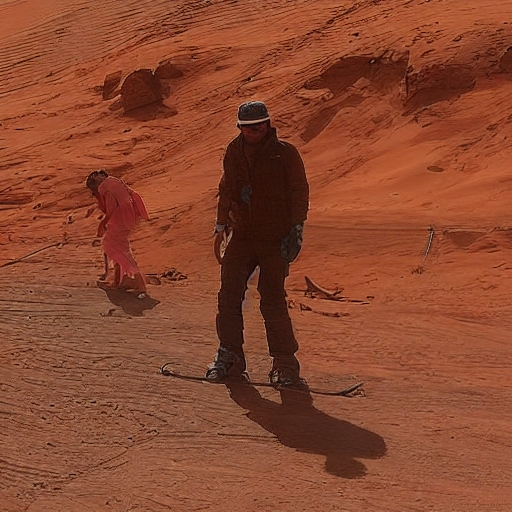}};
  \node[below=\PromptCapImageVSep of sample4_2, anchor=south] {\tiny{\phantom{hg}$\rightarrow$ in the desert\phantom{hg}}};

  \node[right=\PCACompImageHSep of sample4_2, anchor=west] (sample4_2_2) {\includegraphics[width=\PromptImageWidth\linewidth]{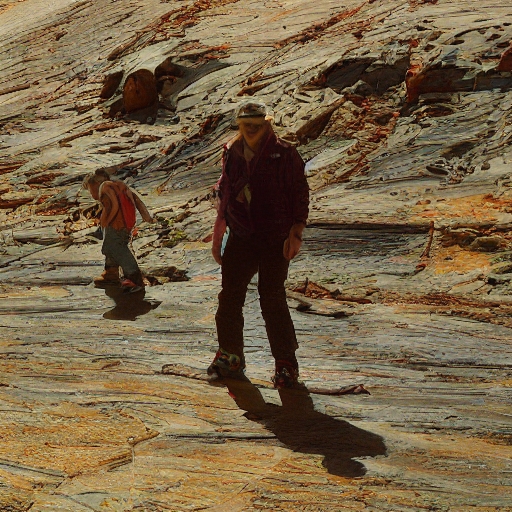}};
  \node[below=\PromptCapImageVSep of sample4_2_2, anchor=south] {\tiny{\phantom{hg}$\rightarrow$ hiking\phantom{hg}}};

  \node[right=\PCACompImageHSep of sample4_2_2, anchor=west] (sample4_3) {\includegraphics[width=\PromptImageWidth\linewidth]{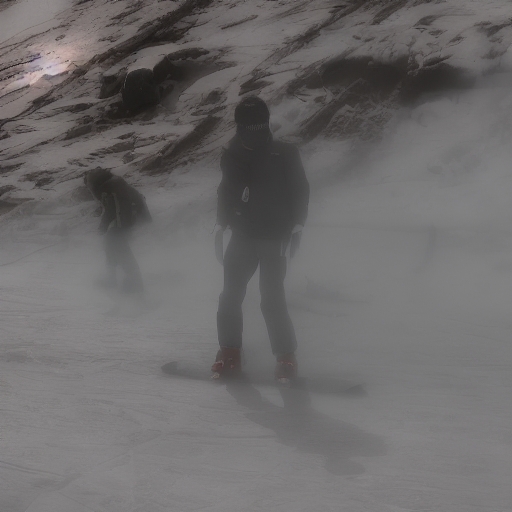}};
  \node[below=\PromptCapImageVSep of sample4_3, anchor=south] {\tiny{\phantom{hg}$+$ in foggy weather\phantom{hg}}};

  \node[right=\PCACompImageHSep of sample4_3, anchor=west] (sample4_4) {\includegraphics[width=\PromptImageWidth\linewidth]{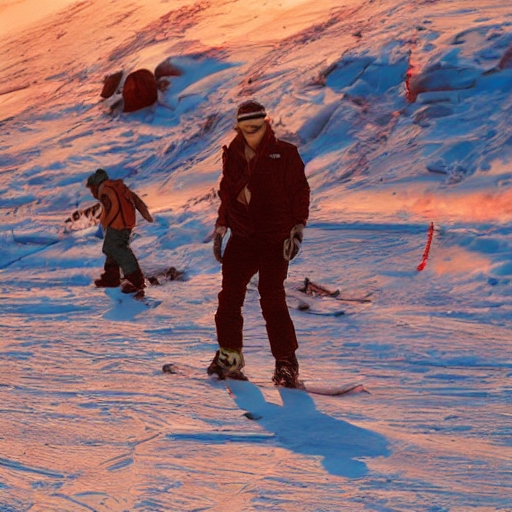}};
  \node[below=\PromptCapImageVSep of sample4_4, anchor=south] {\tiny{\phantom{hg}$+$ at sunrise\phantom{hg}}};

   \node[above=\PromptTitleVSep of sample4_1, anchor=west] {\hspace{-0.073\linewidth}\small{Original caption: There is a man and a young girl \underline{snowboarding}}.};

  \end{tikzpicture}
  \vspace{-1.5em}
\caption{
Image manipulation through prompt editing on the COCO-Stuff validation set.
We show an unedited sample and additional results from replacing the underlined words in the original caption ($\rightarrow$) or appending additional words at its end ($+$).
}
\vspace{-2.5em}
\label{fig:text-prompting}
\end{figure*}
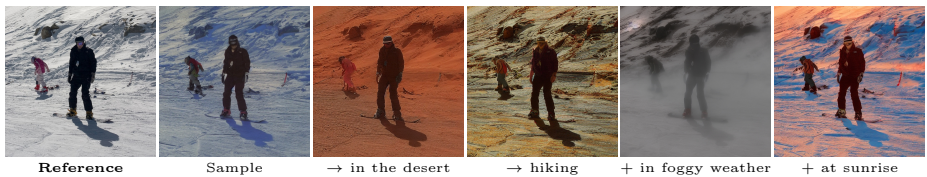

\begin{wraptable}{R}{0.5\textwidth}
\vspace{-4.5em}
\begin{center}
\resizebox{0.5\textwidth}{!}
{
\begin{tabular}{l|c|ccc}
Conditioning & Data & mIoU $\uparrow$ & SI Depth $\downarrow$ & FID $\downarrow$  \\
\hline\hline

\multirow{2}*{Edges (Canny)} & L & 35.7 & 15.8  & 37.6 \\
 & L + U & 38.8 & \color{red}16.0 & \color{red}43.2 \\
 \grayline

\multirow{2}*{Edges (HED)} & L & 51.3 & \underline{11.7} & 31.0 \\
 & L + U  & 53.9 & \textbf{11.4} & \textbf{29.1} \\
  \grayline
\multirow{2}*{Depth (MiDaS)} & L & 44.2 & 16.4 & 37.8 \\
 & L + U   & 46.7 & \color{red}16.5 & 35.9 \\
  \grayline
\multirow{2}*{Sem. Seg.} & L & \underline{57.9} & 16.9 & 44.8 \\
 & L + U* & \color{red}53.3 & 16.2 & \color{red}\emph{46.5}  \\
  \grayline
\multirow{2}*{\textbf{Neural Layout}} & L & 57.6 & 12.1 & 31.2 \\
 & L + U  & \textbf{58.2} & 11.8 & \underline{30.8} 
\end{tabular}   
}
\end{center} 
\vspace{-1.5em}
\caption{
Impact of using \emph{unlabelled} data (U) to augment labelled data (L) to for training the image generator. 
%
%
%
}
\label{table:ablation-data}

\end{wraptable}

\myparagraph{Effect of Data Scaling.}
Our method can utilize additional unlabeled data to enhance synthesis quality.   
This is exemplified with Cityscapes~\cite{cordts2016cityscapes}, which contains only $2975$ training images with pixel-level labels (L). 
However, there are $\sim$20k images without labels (U) from Cityscapes that can be leveraged for training.

The positive impact of this extra data is evident in~\cref{table:ablation-data}.
When trained with only a very limited number of labeled images, 
Sem. Seg. achieves the best alignment in terms of mIoU (57.9), while neural layout achieves comparable mIoU performance (57.6) and much better results in terms of scene geometry, estimated with SI Depth (12.1 vs. 16.9 of Sem. Seg.). 
With the inclusion of more data (L + U), all methods except for Sem. Seg. improve in terms of mIoU.
Notably, our neural layout simultaneously improves in mIoU, SI Depth, and FID, outperforming Sem. Seg. that relies only on manual labels.
To allow Sem. Seg. to also leverage the extra data, we employ the same DRN segmenter\newcite{Yu2017DRN} for pseudo labelling as for mIoU evaluation. However, we observed a significant drop in its mIoU performance, likely attributed to the noise in these pseudo labels.

\vspace{-0.5em}
\subsection{Downstream Applications}
\label{sec:appplication} \vspace{-0.5em}

In this section, we show the applicability of data from~\ourdm~to different downstream applications. 
Since image diversity is crucial for model training, $N=10$ PCA components are kept in the neural layout.

\renewcommand{\PromptImageWidth}{0.1635} 
\renewcommand{\PromptImageVSep}{0.05cm}
\renewcommand{\PromptCapImageVSep}{0.4cm}
\renewcommand{\PromptTitleVSep}{0.25cm}

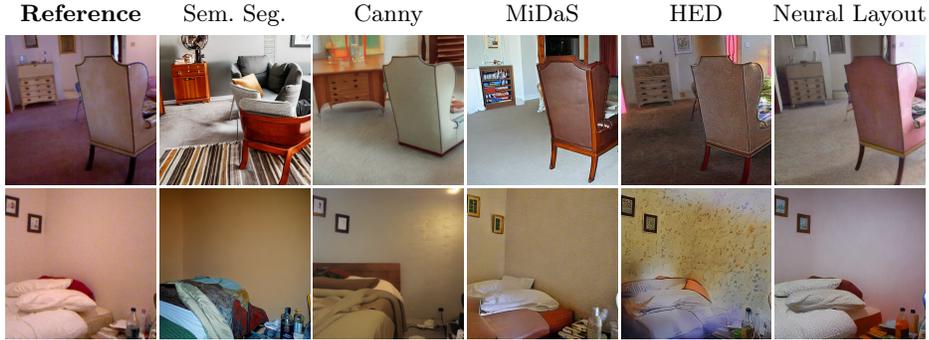
\begin{figure*}[t]
    \centering
  \begin{tikzpicture}[every node/.style={inner sep=0,outer sep=0}]%

  \node (sample2_1) at (0,0) {\includegraphics[width=\PromptImageWidth\linewidth,trim={0cm 0cm 0cm 0cm},clip]{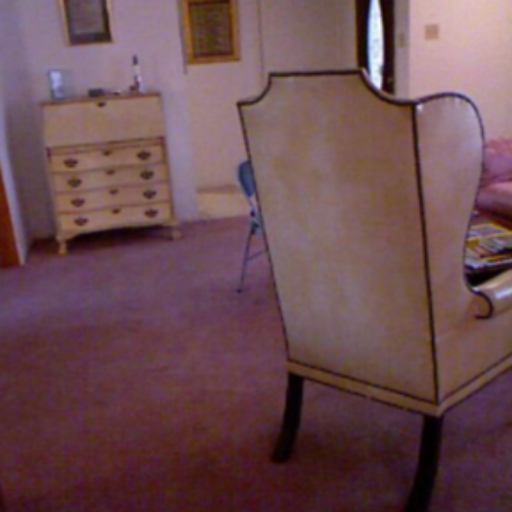}};
\node[above=\PromptCapImageVSep of sample2_1, anchor=north] {\small{\phantom{hg}\textbf{Reference}\phantom{hg}}};

  \node[right=\PCACompImageHSep of sample2_1, anchor=west] (sample2_1p5) {\includegraphics[width=\PromptImageWidth\linewidth,trim={0cm 0cm 0cm 0cm},clip]{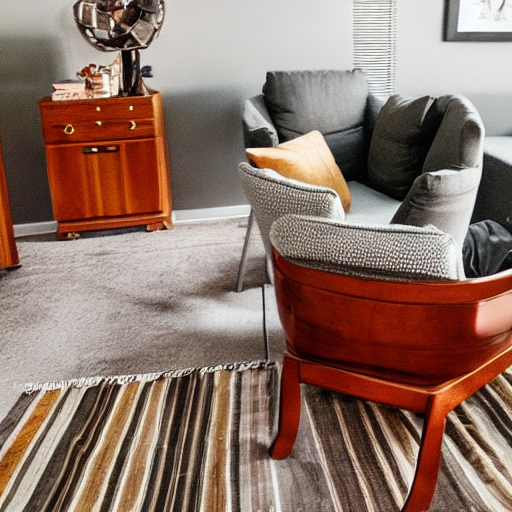}};
  \node[above=\PromptCapImageVSep of sample2_1p5, anchor=north] {\small{\phantom{hg} Sem. Seg. \phantom{hg}}};
  
  \node[right=\PCACompImageHSep of sample2_1p5, anchor=west] (sample2_2) {\includegraphics[width=\PromptImageWidth\linewidth,trim={0cm 0cm 0cm 0cm},clip]{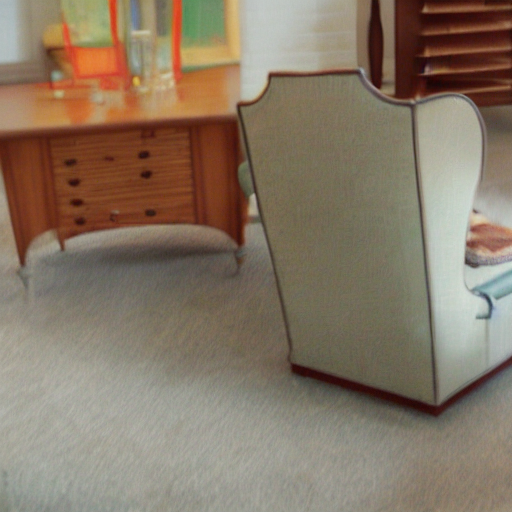}};
\node[above=\PromptCapImageVSep of sample2_2, anchor=north] {\small{\phantom{hg}Canny\phantom{hg}}};
  
  \node[right=\PCACompImageHSep of sample2_2, anchor=west] (sample2_4) {\includegraphics[width=\PromptImageWidth\linewidth,trim={0cm 0cm 0cm 0cm},clip]{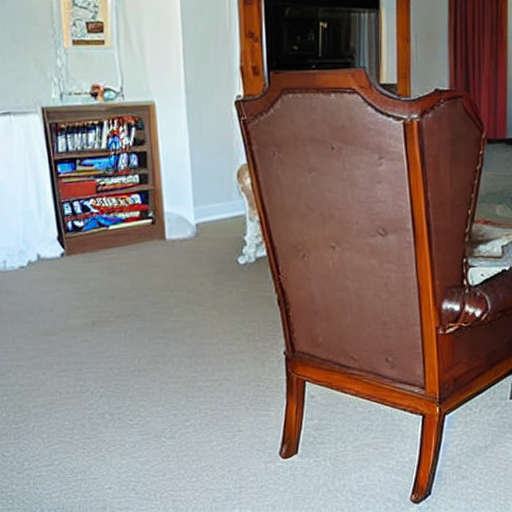}};
\node[above=\PromptCapImageVSep of sample2_4, anchor=north] {\small{\phantom{hg}MiDaS\phantom{hg}}};
  
  \node[right=\PCACompImageHSep of sample2_4, anchor=west] (sample2_5) {\includegraphics[width=\PromptImageWidth\linewidth,trim={0cm 0cm 0cm 0cm},clip]{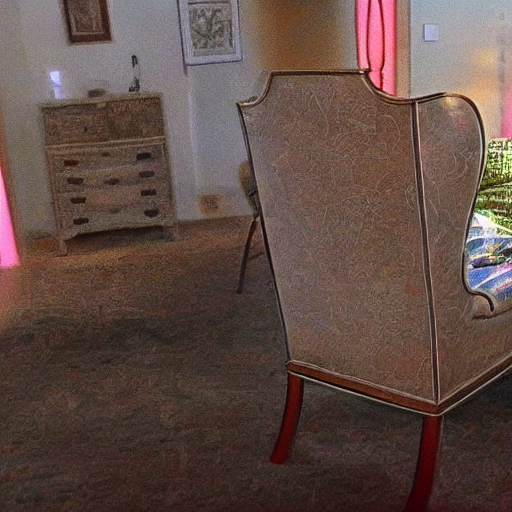}};
  \node[above=\PromptCapImageVSep of sample2_5, anchor=north] {\small{\phantom{hg}HED\phantom{hg}}};
  
  \node[right=\PCACompImageHSep of sample2_5, anchor=west] (sample2_6) {\includegraphics[width=\PromptImageWidth\linewidth,trim={0cm 0cm 0cm 0cm},clip]{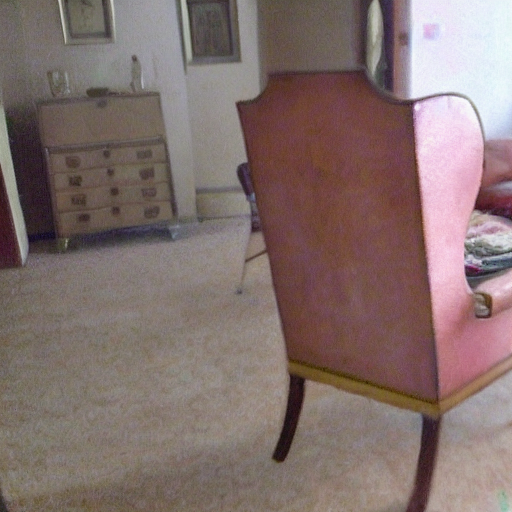}};
  \node[above=\PromptCapImageVSep of sample2_6, anchor=north] {\small{\phantom{hg} Neural Layout \phantom{hg}}};

  \node[below=\PromptImageVSep of sample2_1] (sample3_1)  {\includegraphics[width=\PromptImageWidth\linewidth,trim={0cm 0cm 0cm 0cm},clip]{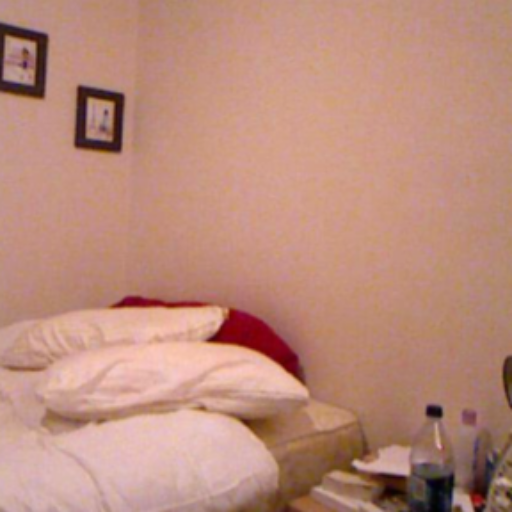}};
  \node[right=\PCACompImageHSep of sample3_1, anchor=west] (sample3_1p5) {\includegraphics[width=\PromptImageWidth\linewidth,trim={0cm 0cm 0cm 0cm},clip]{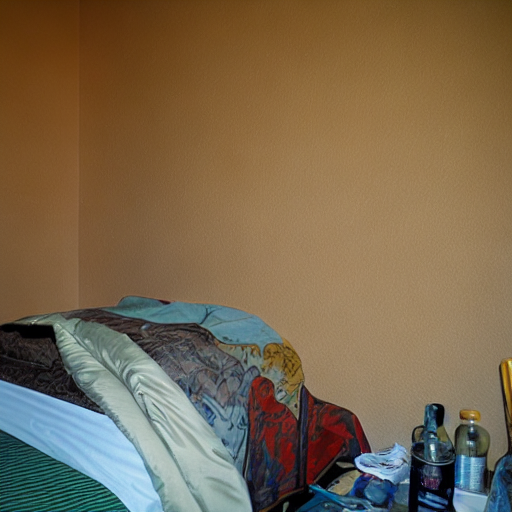}};

  \node[right=\PCACompImageHSep of sample3_1p5, anchor=west] (sample3_2) {\includegraphics[width=\PromptImageWidth\linewidth,trim={0cm 0cm 0cm 0cm},clip]{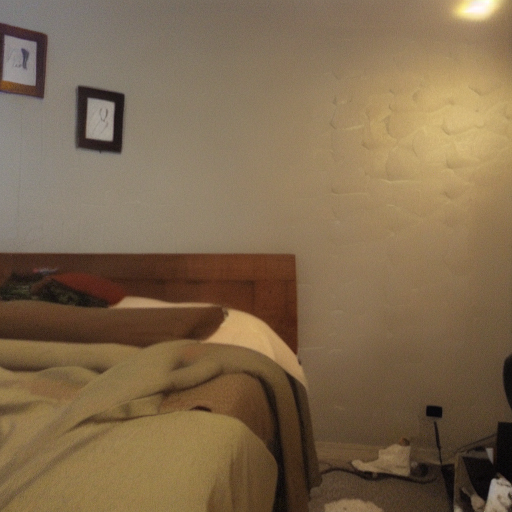}};

  \node[right=\PCACompImageHSep of sample3_2, anchor=west] (sample3_4) {\includegraphics[width=\PromptImageWidth\linewidth,trim={0cm 0cm 0cm 0cm},clip]{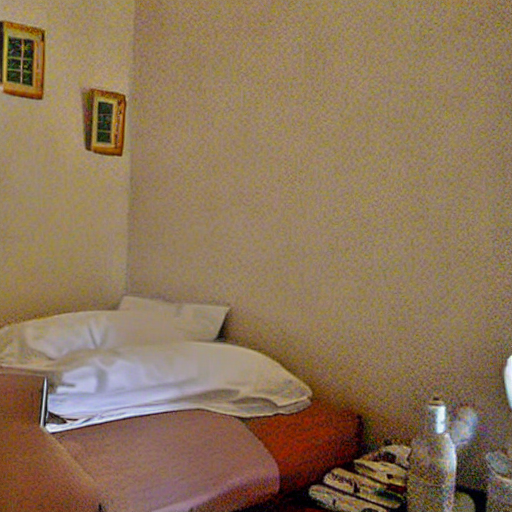}};

  \node[right=\PCACompImageHSep of sample3_4, anchor=west] (sample3_5) {\includegraphics[width=\PromptImageWidth\linewidth,trim={0cm 0cm 0cm 0cm},clip]{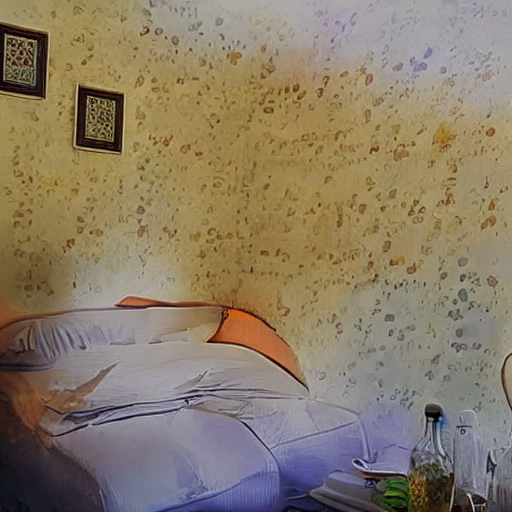}};

  \node[right=\PCACompImageHSep of sample3_5, anchor=west] (sample3_6) {\includegraphics[width=\PromptImageWidth\linewidth,trim={0cm 0cm 0cm 0cm},clip]{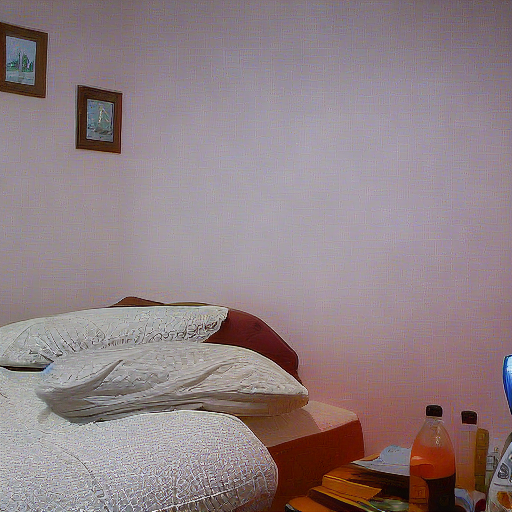}};

  \end{tikzpicture}
  \vspace{-2em}
\caption{
Example of cross-domain (COCO to NYUv2) synthesis. Images created using neural layout best align with the reference image. Other images are mismatched in terms of instance separation, room/object geometry, or object semantics.}
\label{fig:cross-domain}
\vspace{-1.0em}
\end{figure*}
\begin{table}[t]

\begin{center}
\resizebox{0.95\textwidth}{!}
{
\begin{tabular}{l|cc|cc|ccccc}
 & \multicolumn{2}{c|}{sem. seg} & \multicolumn{2}{c|}{depth} & \multicolumn{5}{c}{surface normal}\\
Conditioning &  mIoU $\uparrow$ & pixAcc $\uparrow$ & AbsErr $\downarrow$ & RelErr $\downarrow$ & mean $\downarrow$ & median $\downarrow$ & <11 $\uparrow$ & <22 $\uparrow$ & <33 $\uparrow$\\
\hline\hline
Baseline  & 44.6 & 69.9 & \textbf{1.17} & \textbf{0.56} & 25.8 & 18.7 & 31.7 & 57.3 & 68.6\\ 
\grayline
Edges (Canny) & 45.2  & 70.2 & 1.26 & 0.58 & 26.0 & 18.8 & 31.2 & 57.0 & 68.2\\ 
Edges (HED) & 45.7  & 70.7 & 1.27 & 0.59 & 25.3 & 18.0 & 32.6 & 58.7 & 69.7\\ 
Depth (MiDaS) & 45.8 & 71.2 & 1.24 & 0.58 & 25.2 & 18.0 & 32.4 & 58.7 & 69.7\\ 
SemSeg (pred) & 47.0 & 71.4 & 1.27 & 0.59 & 25.4 & 18.3 & 31.8 & 58.1 & 69.3\\ 
Neural Layout & \textbf{47.3} & \textbf{72.0} & 1.25 & 0.58 & \textbf{24.9} & \textbf{17.7} & \textbf{32.7} & \textbf{59.4} & \textbf{70.4} \\ 
\end{tabular}   
}
\end{center}
\vspace{-0.5em}
\caption{
Synthetic images improve the multitask performance on the NYUv2, despite that the generator is trained on COCO-Stuff. Additional synthetic training data improves upon the baseline in all cases, but most so for neural layouts.
}
\label{table:crossdomain-synthesis}
\vspace{-3.0em}
\end{table}
\myparagraph{Cross Domain Multi-Task Training.} 
Being label-free, we can leverage unsupervised fine-tuning on a large dataset in order to augment a labeled training set with diverse variants.
We demonstrate this by using~\ourdm~finetuned on COCO-Stuff~\cite{caesar2018coco} to create synthetic variants of training data from NYUv2~\cite{silberman2012indoor}. 
NYUv2~\cite{silberman2012indoor} is a small dataset of indoor scenes commonly used for multi-task learning.
LibMTL~\cite{lin2023libmtl}, with its standard setting, was used to train a multi-task network based on ResNet-50~\cite{he2016deep} to predict the semantic segmentation, image depth, and surface normal. 
To obtain the baseline results, we used only the real training data to train the multi-task network.
For all other methods, we use the stated conditioning extracted from reference images in the real training data to create a synthetic dataset of equal size.
In the case of SemSeg conditioning, we treat the segmentation network as a cross domain feature extractor and simply use the predicted COCO-Stuff classes.
The synthetic and real datasets are combined to train the multi-task network.

We report the standard metrics for NYUv2 multi-task learning in~\cref{table:crossdomain-synthesis}. 
Neural layout is able to outperform all other methods, including the real data baseline, for both semantic segmentation and surface normal estimation simultaneously as it preserves both semantics and geometry (see \cref{fig:cross-domain}).
In contrast, even though SemSeg conditioning performs well in the semantic segmentation, the lack of geometric information makes it worse than depth conditioning for estimating surface norms.
Finally, despite the ill-posed nature of the monocular metric depth estimation due to depth-scale ambiguity and the out-of-domain image synthesis, our method performs on-par to explicit depth-based conditioning.

\newcommand{\DomainAdaptImageWidth}{0.240}
\newcommand{\DomainAdaptImageHSep}{0.05cm}
\newcommand{\DomainAdaptImageHSepLarge}{0.5cm}
\newcommand{\DomainAdaptImageHSepH}{0.025cm}
\newcommand{\DomainAdaptImageVSep}{0.05cm}
\newcommand{\DomainAdaptImageVSepLarge}{0.60cm}
\newcommand{\DomainAdaptImageBlockHSep}{0.2cm}
\newcommand{\DomainAdaptImageBlockVSep}{0.2cm}
\newcommand{\DomainAdaptTitleHSep}{0.15cm}
\newcommand{\DomainAdaptTitleVSep}{0.35cm}

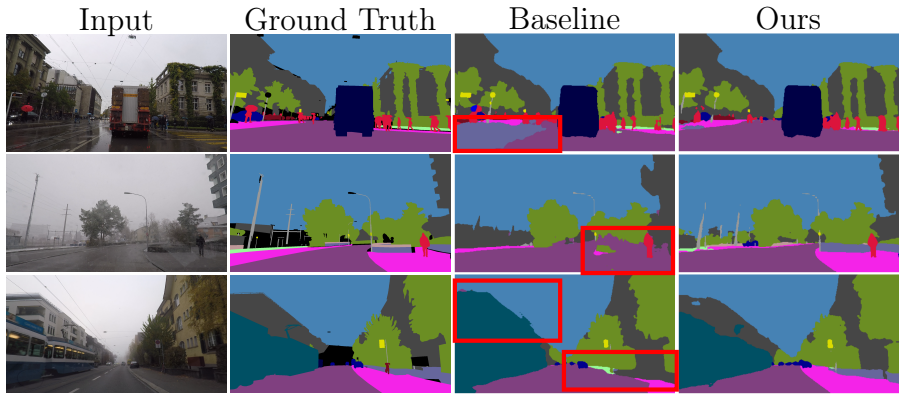
\begin{figure*}[t]
  \centering
  \begin{tikzpicture}[every node/.style={inner sep=0,outer sep=0}]%


  \node (Input1) at (0,0) {\includegraphics[width=\DomainAdaptImageWidth\linewidth, trim={0cm 2.25cm 4.5cm 2.25cm}, clip]{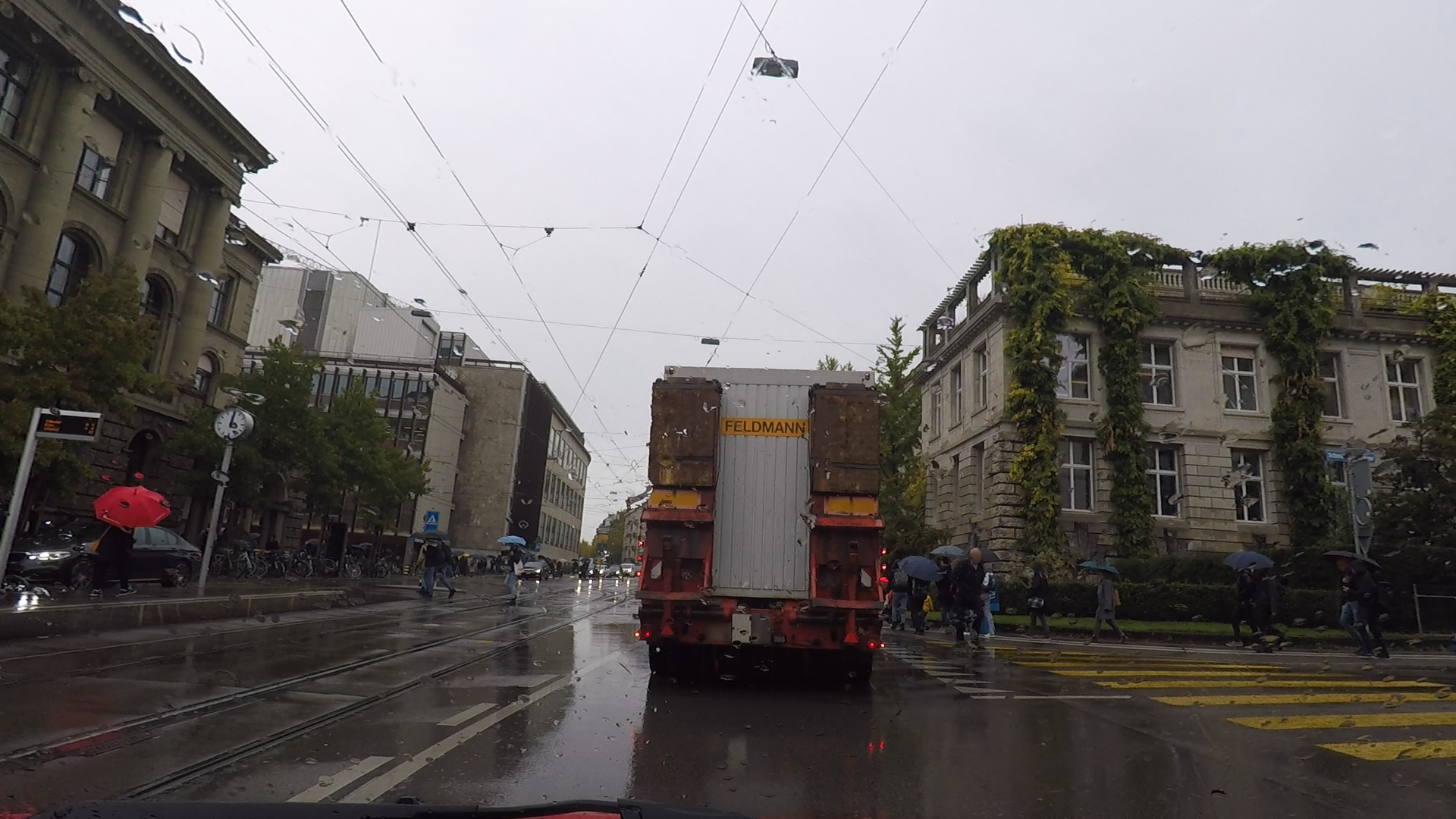}};
  \node[right=\DomainAdaptImageHSep of Input1, anchor=west] (DescriptorSupp1) {\includegraphics[ width=\DomainAdaptImageWidth\linewidth, trim={0cm 2.25cm 4.5cm 2.25cm}, clip]{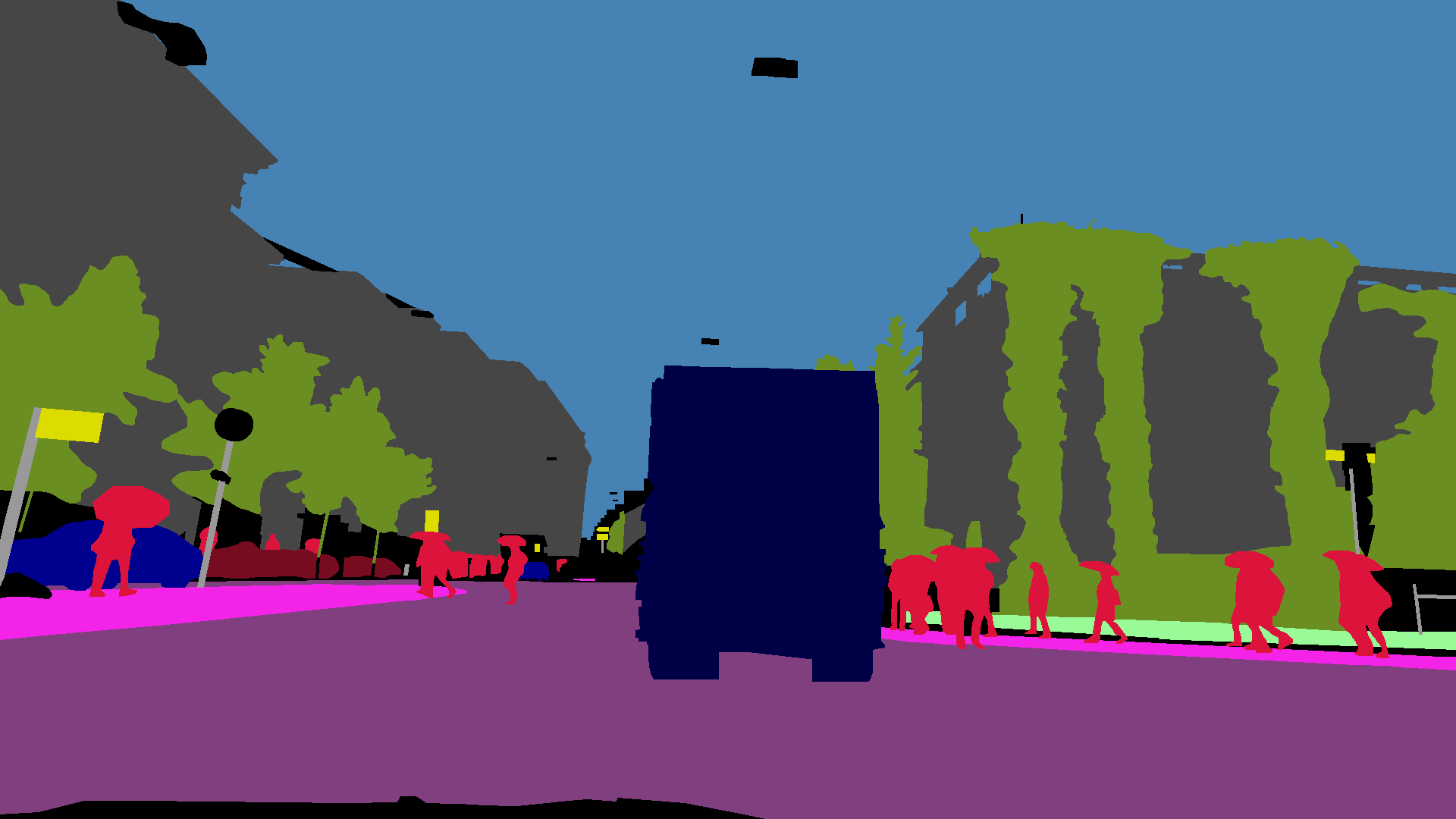}};
  \node[right=\DomainAdaptImageHSep of DescriptorSupp1, anchor=west] (DescriptorSupp2) {\includegraphics[width=\DomainAdaptImageWidth\linewidth, trim={0cm 0cm 1cm 0cm}, clip]{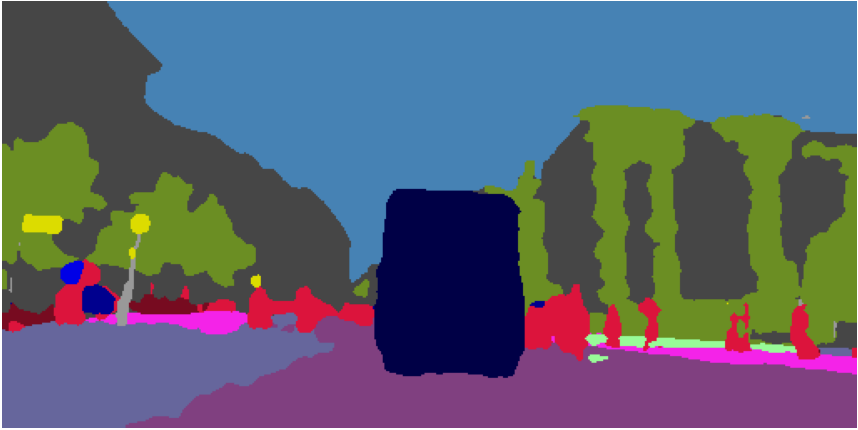}};
  \node[right=\DomainAdaptImageHSep of DescriptorSupp2, anchor=west] (DescriptorSupp3) {\includegraphics[width=\DomainAdaptImageWidth\linewidth, trim={0cm 0cm 1cm 0cm}, clip]{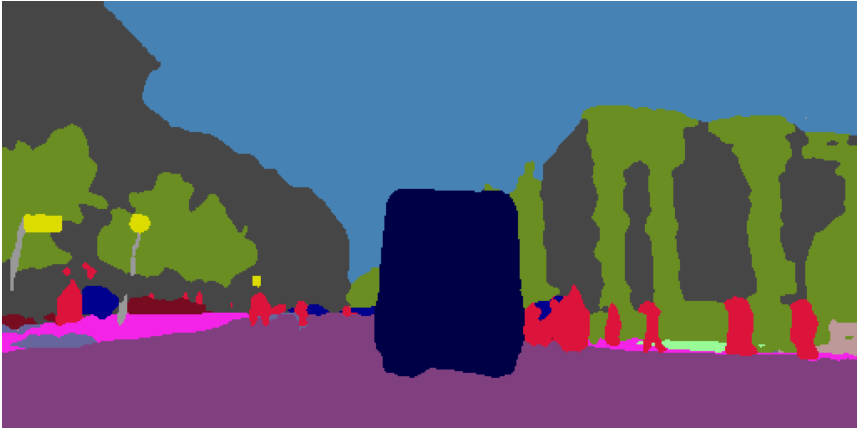}};

   \node[above=\DomainAdaptTitleVSep of Input1, anchor=north] {\large{Input}};
  \node[above=\DomainAdaptTitleVSep of DescriptorSupp1, anchor=north] {\large{Ground Truth}};
  \node[above=\DomainAdaptTitleVSep of DescriptorSupp2, anchor=north] 
  {\large{Baseline}};
  \node[above=\DomainAdaptTitleVSep of DescriptorSupp3, anchor=north] {\large{Ours}};

   \node[below=\DomainAdaptImageVSep of Input1] (Input2) {\includegraphics[width=\DomainAdaptImageWidth\linewidth, trim={0cm 2.25cm 4.5cm 2.25cm}, clip]{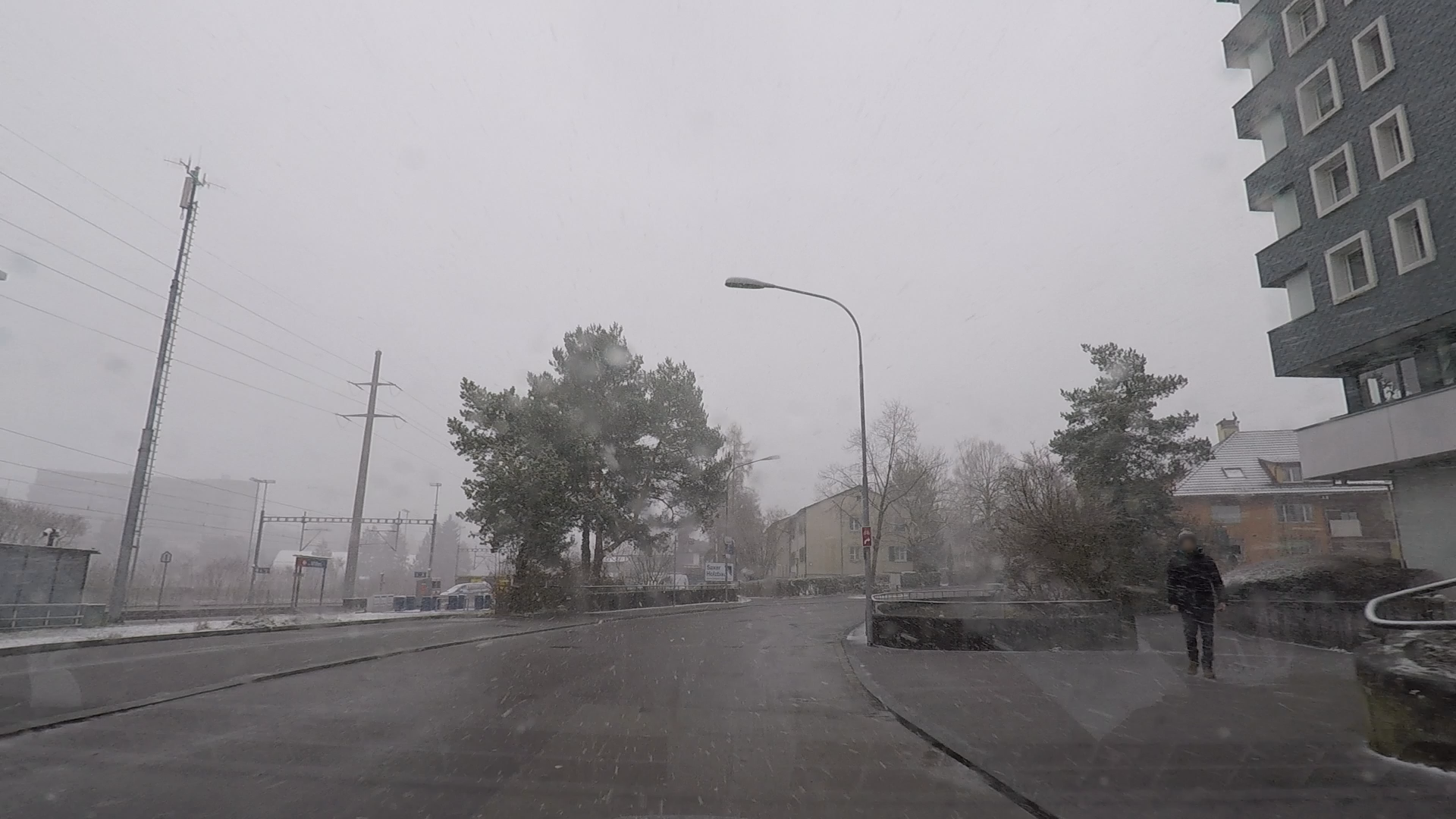}};
  \node[right=\DomainAdaptImageHSep of Input2, anchor=west] (DescriptorSupp1) {\includegraphics[ width=\DomainAdaptImageWidth\linewidth, trim={0cm 2.25cm 4.5cm 2.25cm}, clip]{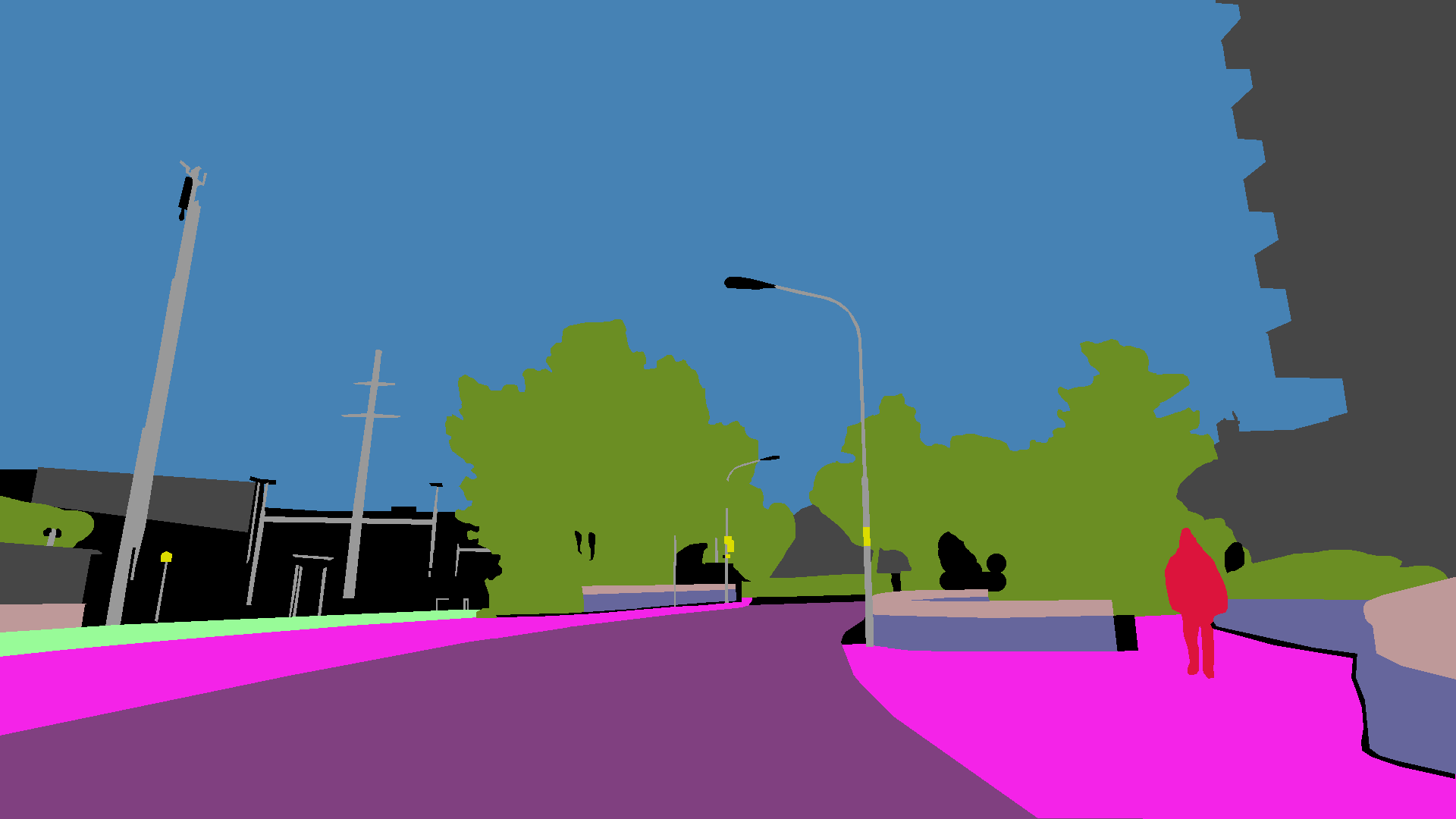}};
  \node[right=\DomainAdaptImageHSep of DescriptorSupp1, anchor=west] (DescriptorSupp2) {\includegraphics[width=\DomainAdaptImageWidth\linewidth, trim={0cm 0cm 1cm 0cm}, clip]{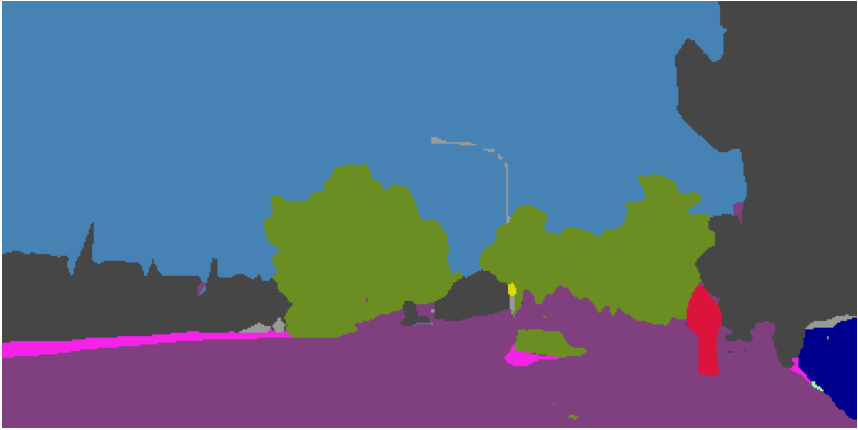}};
  \node[right=\DomainAdaptImageHSep of DescriptorSupp2, anchor=west] (DescriptorSupp3) {\includegraphics[width=\DomainAdaptImageWidth\linewidth, trim={0cm 0cm 1cm 0cm}, clip]{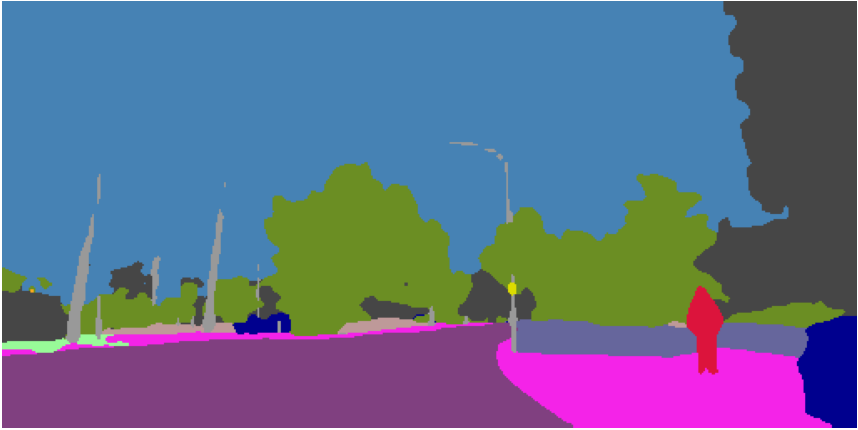}};
  
   \node[below=\DomainAdaptImageVSep of Input2] (Input3) {\includegraphics[width=\DomainAdaptImageWidth\linewidth, trim={0cm 2.25cm 4.5cm 2.25cm}, clip]{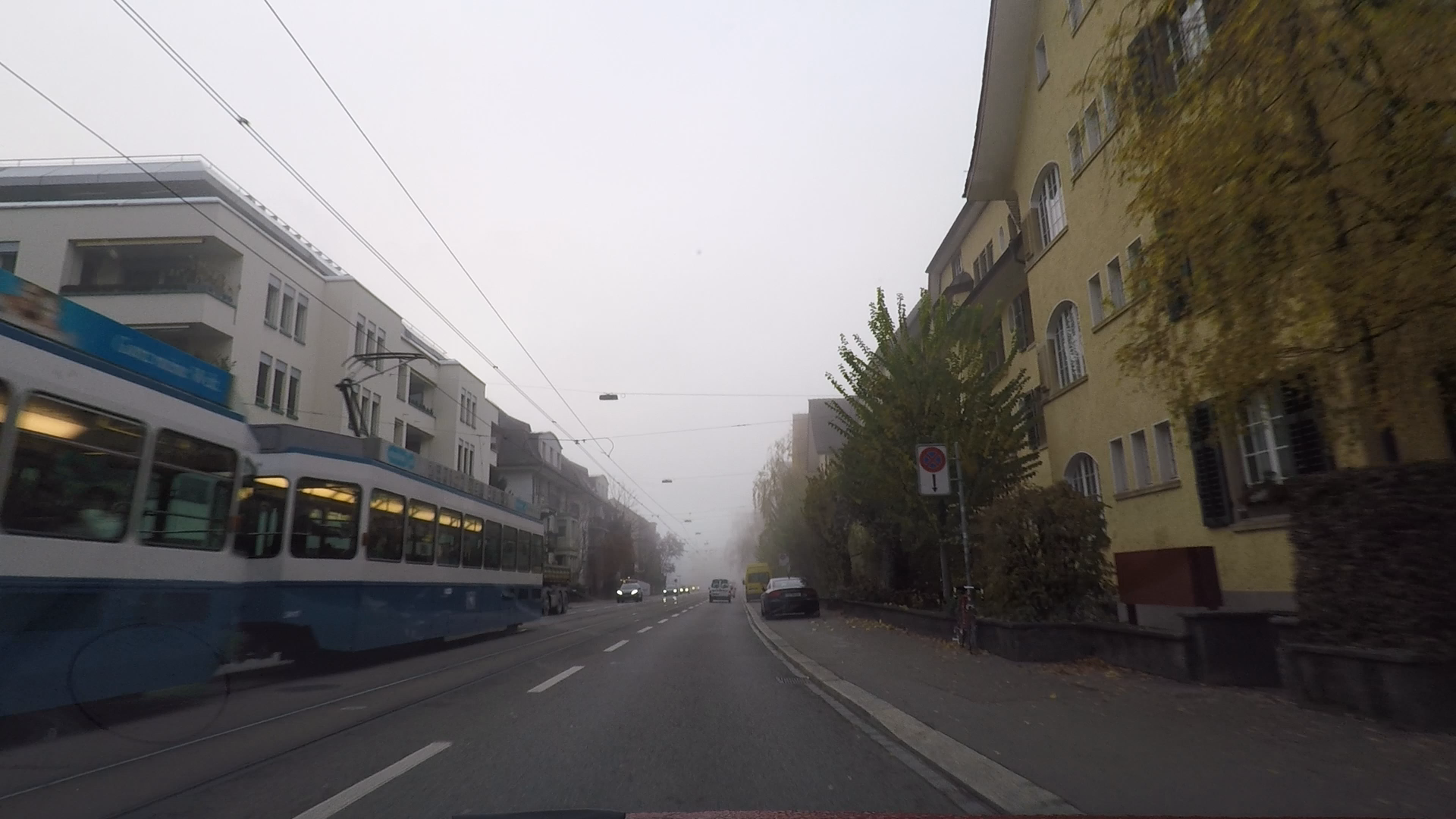}};
  \node[right=\DomainAdaptImageHSep of Input3, anchor=west] (DescriptorSupp1) {\includegraphics[ width=\DomainAdaptImageWidth\linewidth, trim={0cm 2.25cm 4.5cm 2.25cm}, clip]{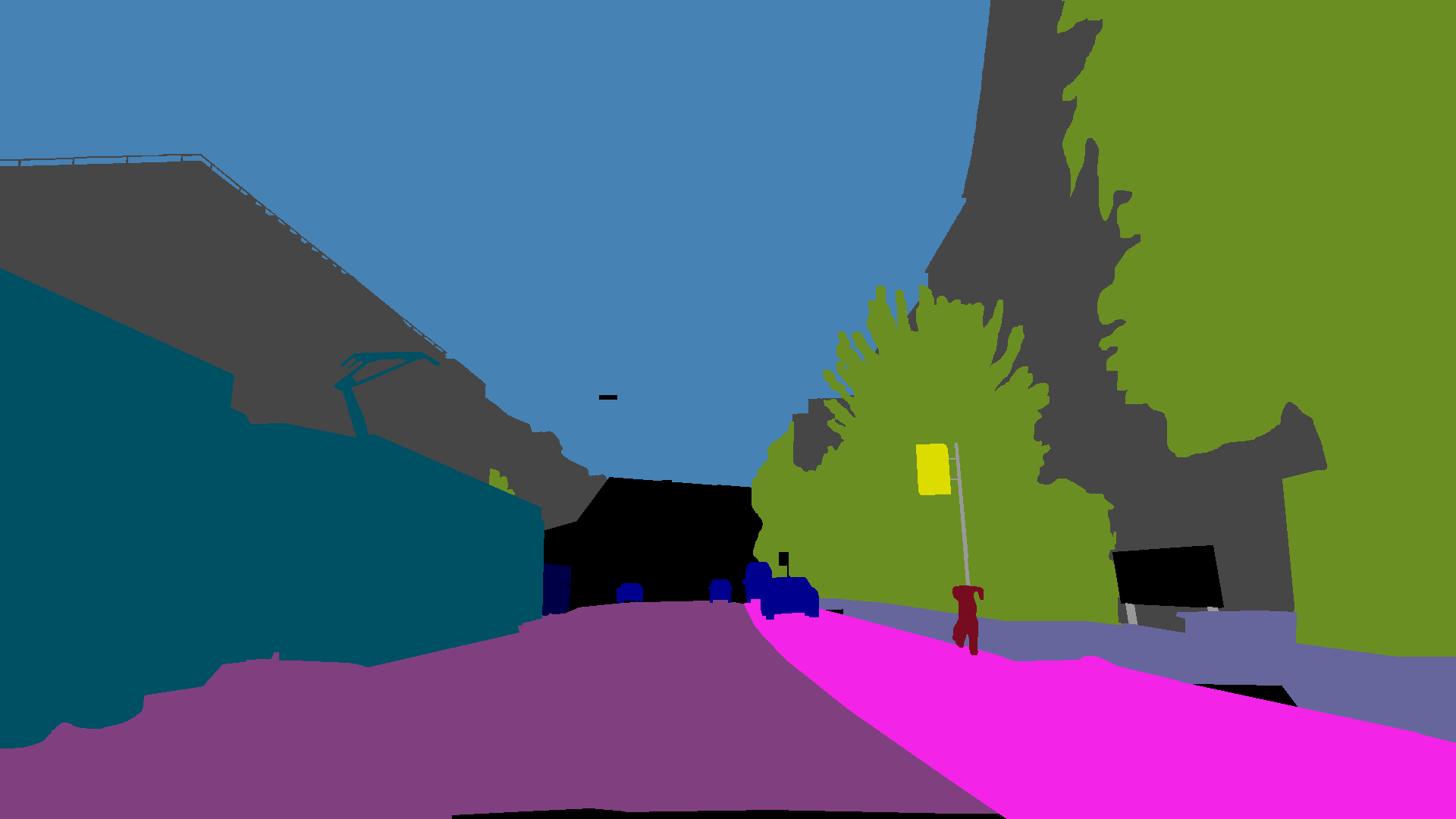}};
  \node[right=\DomainAdaptImageHSep of DescriptorSupp1, anchor=west] (DescriptorSupp2) {\includegraphics[width=\DomainAdaptImageWidth\linewidth, trim={0cm 0cm 1cm 0cm}, clip]{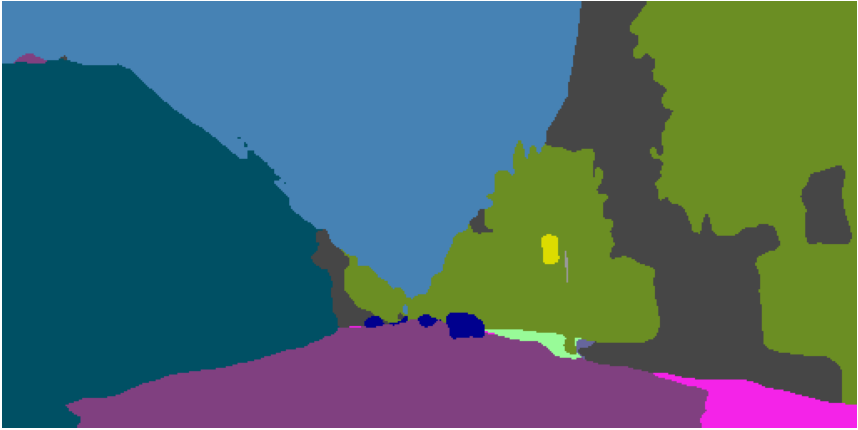}};
  \node[right=\DomainAdaptImageHSep of DescriptorSupp2, anchor=west] (DescriptorSupp3) {\includegraphics[width=\DomainAdaptImageWidth\linewidth, trim={0cm 0cm 1cm 0cm}, clip]{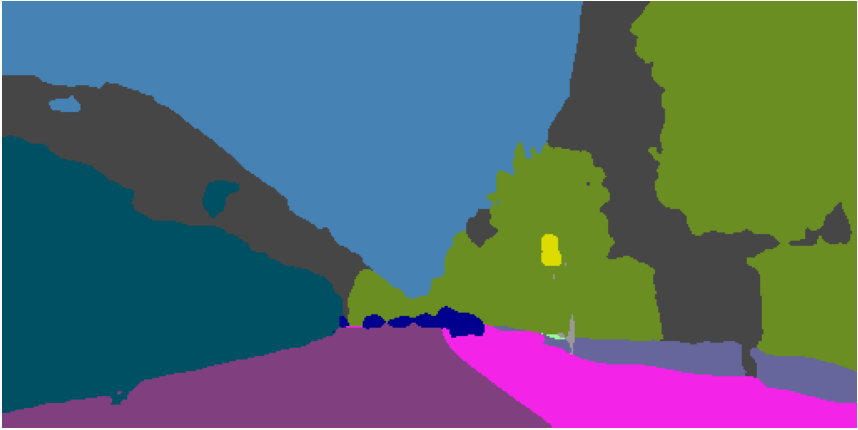}};

  \node (rect) [rectangle,draw,ultra thick,red,minimum width=1.4cm,minimum height=0.45cm] at (5.2, -0.55) (){};
  \node (rect) [rectangle,draw,ultra thick,red,minimum width=1.2cm,minimum height=0.6cm] at (6.8, -2.1) (){};
  \node (rect) [rectangle,draw,ultra thick,red,minimum width=1.4cm,minimum height=0.8cm] at (5.2, -2.9) (){};
  \node (rect) [rectangle,draw,ultra thick,red,minimum width=1.5cm,minimum height=0.5cm] at (6.7, -3.7) (){};
  
  \end{tikzpicture}
  \vspace{-0.5em}
\caption{
Our additional synthesized data improves upon baseline of using only real Cityscapes data for domain generalization to the ADAC dataset.
}
\label{fig:DG-supp}
\vspace{-2em}
\end{figure*}


\begin{wraptable}{R}{0.5\textwidth}
\vspace{-3em}
\begin{center}
\resizebox{0.5\textwidth}{!}
{ \begin{tabular}{l|c|ccccc}
        Method & CS & Rain & Fog & Snow & Night & Avg. \\ \hline\hline
        Baseline (CS)  & 67.9 & 50.2 & 60.5 & 48.9 & 28.6 & 47.0\\ 
        \grayline
        PnP-Diffusion\newcite{tumanyan2023plug} & 67.8 & 50.6 & 63.5 & 50.4 & 30.3 & 48.7 \\
        ControlNet [SemSeg] & 66.8 & 50.4 & 64.7 & 51.7 & 34.6 & 50.3 \\
        FreestyleNet*~\newcite{xue2023freestyle}& \textbf{69.7} & 52.7 & \textbf{69.0} & 54.3 & 32.9 & 52.2 \\
       \textbf{{\ourdm} (Ours)} & 68.5 & \textbf{53.4} & 67.4 & \textbf{55.6} & \textbf{35.1} & \textbf{52.9} \\ 
        \grayline
        Oracle (CS + ACDC) & 68.2 & 63.7 & 74.1 & 68.0 & 48.8 & 63.6 
        \vspace{-1em}
    \end{tabular}
}
\end{center}
\vspace{-1em}
\caption{
Comparison of augmentation techniques for domain generalization via prompt edit. 
Quality is measured using mIoU of a SegFormer~\cite{xie2021segformer}.
}\label{table:CS-DG-compare}
\vspace{-2em}
\end{wraptable}

\myparagraph{Domain Generalization.} 
By choosing to use ControlNet as our backbone generator, we can use the text prompt to control the domain (mainly appearance) of the synthesized images while using neural layout to control the semantic and geometric content.
We use this to perform domain generalization experiments from the daytime only Cityscapes to ACDC\newcite{sakaridis2021acdc}, containing adverse weather and lighting conditions (for more details see suppl. material).
As shown in~\cref{table:CS-DG-compare} and in~\cref{fig:DG-supp}, we verified that images generated by {\ourdm} can significantly improve the model's generalization ability upon the baseline, which is trained only on  Cityscapes.
We compare against other diffusion-based methods PnP-Diffusion\newcite{tumanyan2023plug}, FreestyleNet\newcite{xue2023freestyle}, as well as ControlNet\newcite{zhang2023controlnet} with predicted Sem. Seg. conditioning. 
The prompt editing of PnP-Diffusion cannot generalize well to the image domain of Cityscapes, leading to little benefits for domain generalization. 
FreestyleNet require manual annotation to train the image generator, in contrast to our label-free~{\ourdm}. 
Yet, our method still outperforms FreestyleNet overall.

\begin{figure}[t]
    \begin{centering}
    \setlength{\tabcolsep}{0.0em}
    \renewcommand{\arraystretch}{0}
    \par\end{centering}
    \begin{centering}
    \vspace{-0.5em}
    \hfill{}%
	\begin{tabular}{@{}c@{}c@{}c@{}c}
        \centering
		\small Original & \small ``Anime'' &  \small ``Sketch'' & \small ``3D Render''
  \tabularnewline
		\includegraphics[width=.20\linewidth, trim={0cm 0cm 0cm 0cm},clip]{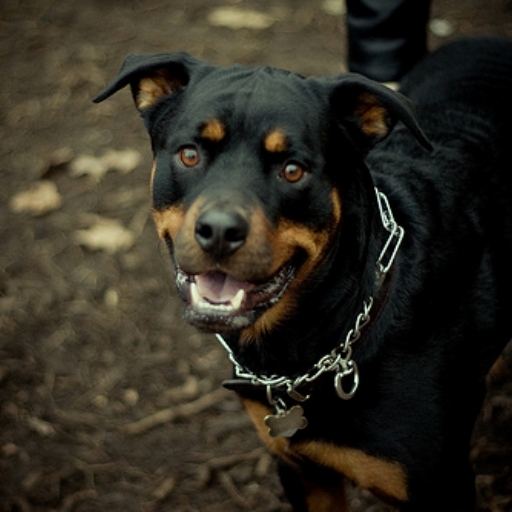} & {}
    \includegraphics[width=.20\linewidth, trim={0cm 0cm 0cm 0cm},clip]{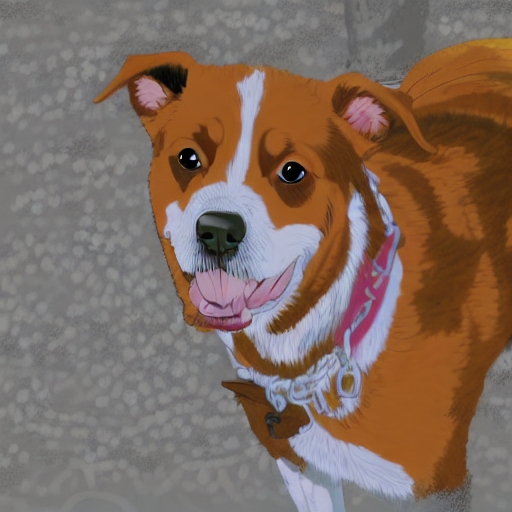}  & {}
    \includegraphics[width=.20\linewidth, trim={0cm 0cm 0cm 0cm},clip]{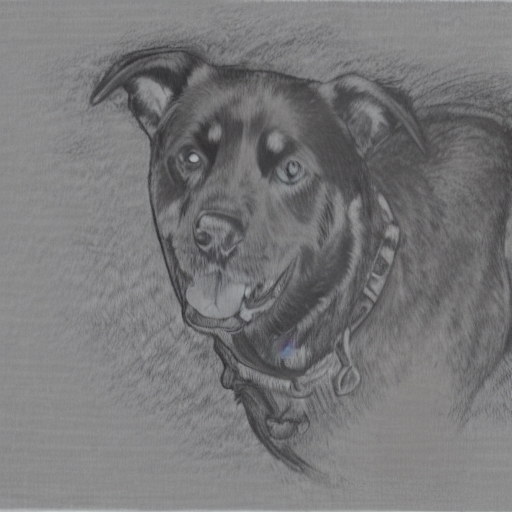} & {}
    \includegraphics[width=.20\linewidth, trim={0cm 0cm 0cm 0cm},clip]{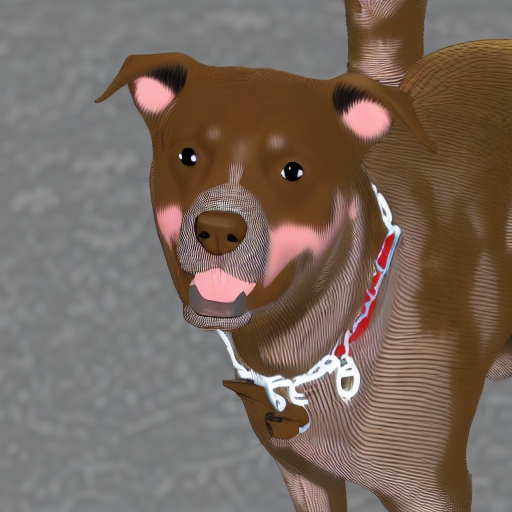}
		\tabularnewline
		\end{tabular}
\hfill{}
\par\end{centering}
    \vspace{-1.em}
    \caption{LUMEN can create artistic stylized versions of an image using prompt editing.
    }
    
\label{fig:rebuttal_style}
\end{figure}

\begin{figure}[t]
    \begin{centering}
    \setlength{\tabcolsep}{0.0em}
    \renewcommand{\arraystretch}{0}
    \par\end{centering}
    \begin{centering}
    \vspace{-0.5em}
    \hfill{}%
	\begin{tabular}{@{}c@{}c@{}c@{}c}
        \centering
		\small Image 1 &  \small Image 2 & \small Neural Layout &  \small Edited
  \tabularnewline
		\includegraphics[width=.20\linewidth, trim={0cm 0cm 0cm 5.8cm},clip]{rebuttal_figures/composite/bathroom.jpg}  &
		{} \includegraphics[width=.20\linewidth, trim={0cm 0cm 0cm 5.8cm},clip]{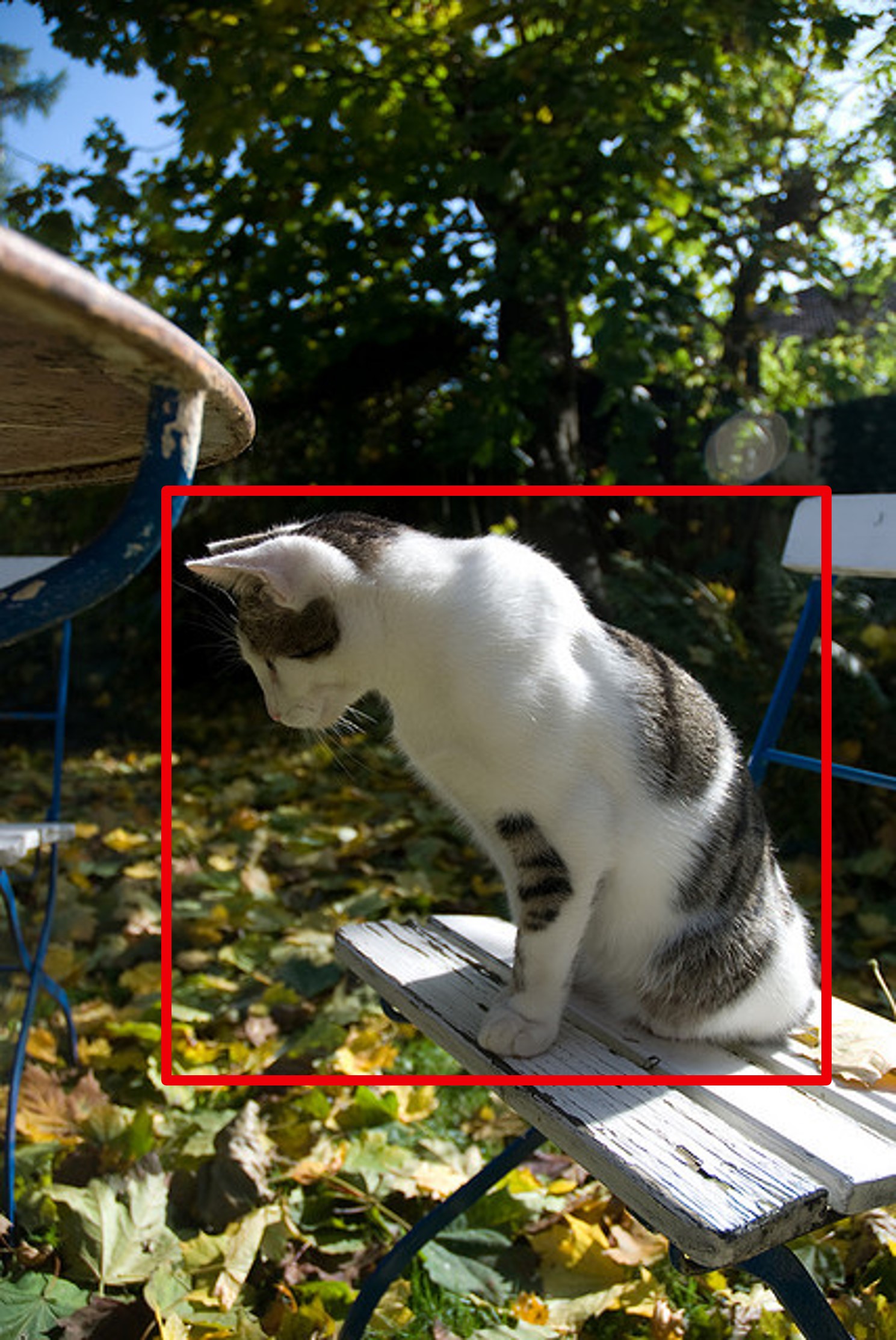} &
        {}
		\includegraphics[width=.20\linewidth, trim={0cm 0cm 0cm 0cm},clip]{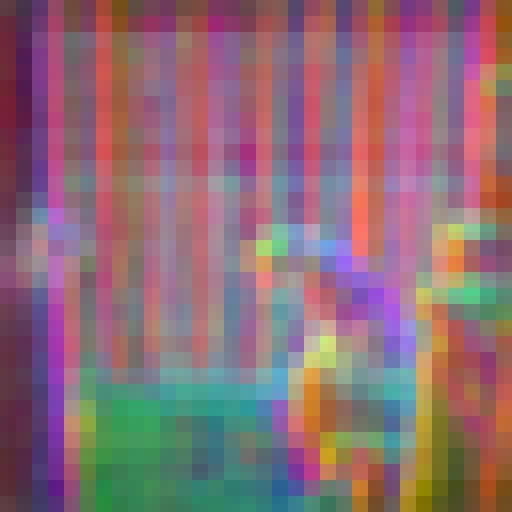}  & {}
		\includegraphics[width=.20\linewidth, trim={0cm 0cm 0cm 0cm},clip]{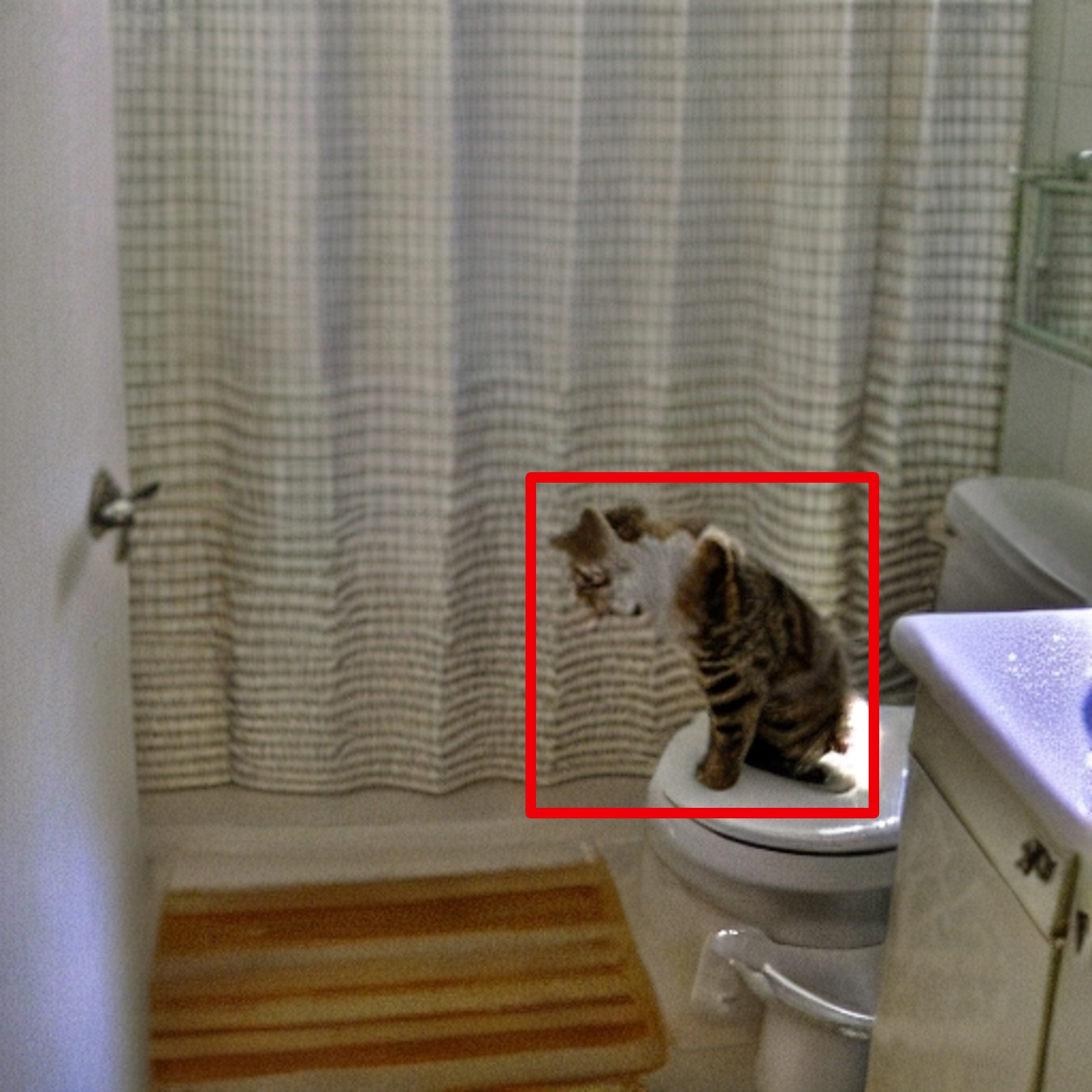} 
		\tabularnewline
		\end{tabular}
\hfill{}
\par\end{centering}
\vspace{-1.0em}
    \caption{
    Neural layouts can be combined to define desired layout in the final image.
    }
    \label{fig:rebuttal_composite}
\vspace{-2.0em}
\end{figure}

\myparagraph{Content Creation.} Although we motivate our method by considering applications where~\ourdm~is used to create training data, it is suitable for content creation as well. 
We show in \cref{fig:rebuttal_style} that~\ourdm~is not limited to synthesize realistic images, but is also versatile enough to generate images with a variety of different artistic styles. 
This suggests that the semantically separated neural layout is able to discard stylistic content of the reference. 
We further show in \cref{fig:rebuttal_composite} that, with user interaction, object layouts and scene composition can be altered by combining different images. 
In general, ~\ourdm~enables a new workflow where a creator can specify both the desired spatial and semantic layout of the output while relying on the diffusion model to fill in the appearance.

\vspace{-0.5em}
\section{Conclusion}
\vspace{-0.4em}
We introduced the concept of neural semantic image synthesis and established~\ourdm~ as a strong label-free baseline that can simultaneous specify semantic and spatial concepts of the outputs. 
Neural semantic image synthesis has additional applications that we hope will be explored in future works. 
For example, targeted removal of irrelevant information could be possible by replacing PCA with a learned projector specialized to a downstream task.
Similarly, one could use neural semantic image synthesis to create images in a target domain from images in a source domain if a projection direction can be learned to eliminate the domain differences.
This allows the reuse of labels from one image domain to another in order to save costly annotation efforts.

%
%
\bibliographystyle{splncs04}
\bibliography{reference}

\begin{thebibliography}{10}
\providecommand{\url}[1]{\texttt{#1}}
\providecommand{\urlprefix}{URL }
\providecommand{\doi}[1]{https://doi.org/#1}

\bibitem{midjourney}
Midjourney. \url{https://www.midjourney.com/} (2023)

\bibitem{amir2021deepvit}
Amir, S., Gandelsman, Y., Bagon, S., Dekel, T.: Deep vit features as dense visual descriptors. arXiv preprint arXiv:2112.05814  (2021)

\bibitem{balaji2022ediffi}
Balaji, Y., Nah, S., Huang, X., Vahdat, A., Song, J., Kreis, K., Aittala, M., Aila, T., Laine, S., Catanzaro, B., et~al.: ediffi: Text-to-image diffusion models with an ensemble of expert denoisers. arXiv preprint arXiv:2211.01324  (2022)

\bibitem{caesar2018coco}
Caesar, H., Uijlings, J., Ferrari, V.: Coco-stuff: Thing and stuff classes in context. In: CVPR (2018)

\bibitem{canny1986computational}
Canny, J.: A computational approach to edge detection. TPAMI  (1986)

\bibitem{caron2021dino}
Caron, M., Touvron, H., Misra, I., J{\'e}gou, H., Mairal, J., Bojanowski, P., Joulin, A.: Emerging properties in self-supervised vision transformers. In: ICCV (2021)

\bibitem{chen2023anydoor}
Chen, X., Huang, L., Liu, Y., Shen, Y., Zhao, D., Zhao, H.: Anydoor: Zero-shot object-level image customization. arXiv preprint arXiv:2307.09481  (2023)

\bibitem{Cheng2022mask2former}
Cheng, B., Misra, I., Schwing, A.G., Kirillov, A., Girdhar, R.: Masked-attention mask transformer for universal image segmentation. In: CVPR (2022)

\bibitem{cordts2016cityscapes}
Cordts, M., Omran, M., Ramos, S., Rehfeld, T., Enzweiler, M., Benenson, R., Franke, U., Roth, S., Schiele, B.: The cityscapes dataset for semantic urban scene understanding. In: CVPR (2016)

\bibitem{eigen2014depth}
Eigen, D., Puhrsch, C., Fergus, R.: Depth map prediction from a single image using a multi-scale deep network. NeurIPS  (2014)

\bibitem{he2016deep}
He, K., Zhang, X., Ren, S., Sun, J.: Deep residual learning for image recognition. In: ICCV (2016)

\bibitem{hedlin2023unsupervised}
Hedlin, E., Sharma, G., Mahajan, S., Isack, H., Kar, A., Tagliasacchi, A., Yi, K.M.: Unsupervised semantic correspondence using stable diffusion. arXiv preprint arXiv:2305.15581  (2023)

\bibitem{heusel2017gansFIDal}
Heusel, M., Ramsauer, H., Unterthiner, T., Nessler, B., Hochreiter, S.: Gans trained by a two time-scale update rule converge to a local nash equilibrium. In: NeurIPS (2017)

\bibitem{hu2023tifa}
Hu, Y., Liu, B., Kasai, J., Wang, Y., Ostendorf, M., Krishna, R., Smith, N.A.: T{I}{F}{A}: Accurate and interpretable text-to-image faithfulness evaluation with question answering. arXiv preprint arXiv:2303.11897  (2023)

\bibitem{isola2017image}
Isola, P., Zhu, J.Y., Zhou, T., Efros, A.A.: Image-to-image translation with conditional adversarial networks. In: CVPR (2017)

\bibitem{li2023maskdino}
Li, F., Zhang, H., Xu, H., Liu, S., Zhang, L., Ni, L.M., Shum, H.Y.: Mask dino: Towards a unified transformer-based framework for object detection and segmentation. In: CVPR (2023)

\bibitem{li2022blip}
Li, J., Li, D., Xiong, C., Hoi, S.: Blip: Bootstrapping language-image pre-training for unified vision-language understanding and generation. In: ICML (2022)

\bibitem{gligen-cvpr}
Li, Y., Liu, H., Wu, Q., Mu, F., Yang, J., Gao, J., Li, C., Lee, Y.J.: Gligen: Open-set grounded text-to-image generation. In: CVPR (2023)

\bibitem{li2024adversarial}
Li, Y., Keuper, M., Zhang, D., Khoreva, A.: Adversarial supervision makes layout-to-image diffusion models thrive. In: ICLR (2024)

\bibitem{lin2023libmtl}
Lin, B., Zhang, Y.: {LibMTL}: A {P}ython library for multi-task learning. Journal of Machine Learning Research  \textbf{24}(209), ~1--7 (2023)

\bibitem{mou2023t2i-adapter}
Mou, C., Wang, X., Xie, L., Zhang, J., Qi, Z., Shan, Y., Qie, X.: T2i-adapter: Learning adapters to dig out more controllable ability for text-to-image diffusion models. arXiv preprint arXiv:2302.08453  (2023)

\bibitem{oquab2023dinov2}
Oquab, M., Darcet, T., Moutakanni, T., Vo, H., Szafraniec, M., Khalidov, V., Fernandez, P., Haziza, D., Massa, F., El-Nouby, A., et~al.: Dinov2: Learning robust visual features without supervision. arXiv preprint arXiv:2304.07193  (2023)

\bibitem{park2019spade}
Park, T., Liu, M.Y., Wang, T.C., Zhu, J.Y.: Semantic image synthesis with spatially-adaptive normalization. In: CVPR (2019)

\bibitem{GlueGen23ICCV}
Qin, C., Yu, N., Xing, C., Zhang, S., Chen, Z., Ermon, S., Fu, Y., Xiong, C., Xu, R.: Gluegen: Plug and play multi-modal encoders for x-to-image generation. In: ICCV (2023)

\bibitem{clip}
Radford, A., Kim, J.W., Hallacy, C., Ramesh, A., Goh, G., Agarwal, S., Sastry, G., Askell, A., Mishkin, P., Clark, J., Krueger, G., Sutskever, I.: Learning transferable visual models from natural language supervision. In: ICML (2021)

\bibitem{ramesh2022dalle2}
Ramesh, A., Dhariwal, P., Nichol, A., Chu, C., Chen, M.: Hierarchical text-conditional image generation with clip latents. arXiv preprint arXiv:2204.06125  (2022)

\bibitem{Ranftl2022midas}
Ranftl, R., Lasinger, K., Hafner, D., Schindler, K., Koltun, V.: Towards robust monocular depth estimation: Mixing datasets for zero-shot cross-dataset transfer. TPAMI  (2022)

\bibitem{rombach2022SD}
Rombach, R., Blattmann, A., Lorenz, D., Esser, P., Ommer, B.: High-resolution image synthesis with latent diffusion models. In: CVPR (2022)

\bibitem{palette}
Saharia, C., Lee, C.A., Fleet, D.J., Chang, H., Ho, J., Norouzi, M., Salimans, T., Chan, W.: Palette: Image-to-image diffusion models. In: SIGGRAPH (2022)

\bibitem{sakaridis2021acdc}
Sakaridis, C., Dai, D., Van~Gool, L.: Acdc: {T}he adverse conditions dataset with correspondences for semantic driving scene understanding. In: ICCV (2021)

\bibitem{silberman2012indoor}
Silberman, N., Hoiem, D., Kohli, P., Fergus, R.: Indoor segmentation and support inference from rgbd images. In: ECCV. pp. 746--760 (2012)

\bibitem{sushko2022oasis}
Sushko, V., Sch{\"o}nfeld, E., Zhang, D., Gall, J., Schiele, B., Khoreva, A.: Oasis: only adversarial supervision for semantic image synthesis. IJCV  (2022)

\bibitem{tan2021clade}
Tan, Z., Chen, D., Chu, Q., Chai, M., Liao, J., He, M., Yuan, L., Hua, G., Yu, N.: Efficient semantic image synthesis via class-adaptive normalization. TPAMI  (2021)

\bibitem{tumanyan2023plug}
Tumanyan, N., Geyer, M., Bagon, S., Dekel, T.: Plug-and-play diffusion features for text-driven image-to-image translation. In: CVPR (2023)

\bibitem{vaswani2017attention}
Vaswani, A., Shazeer, N., Parmar, N., Uszkoreit, J., Jones, L., Gomez, A.N., Kaiser, {\L}., Polosukhin, I.: Attention is all you need. In: NeurIPS (2017)

\bibitem{wang2022piti}
Wang, T., Zhang, T., Zhang, B., Ouyang, H., Chen, D., Chen, Q., Wen, F.: Pretraining is all you need for image-to-image translation. arXiv preprint arXiv:2205.12952  (2022)

\bibitem{wang2018pix2pixhd}
Wang, T.C., Liu, M.Y., Zhu, J.Y., Tao, A., Kautz, J., Catanzaro, B.: High-resolution image synthesis and semantic manipulation with conditional gans. In: CVPR (2018)

\bibitem{wang2021scgan}
Wang, Y., Qi, L., Chen, Y.C., Zhang, X., Jia, J.: Image synthesis via semantic composition. In: ICCV (2021)

\bibitem{xie2021segformer}
Xie, E., Wang, W., Yu, Z., Anandkumar, A., Alvarez, J.M., Luo, P.: Segformer: {S}imple and efficient design for semantic segmentation with transformers. In: NeurIPS (2021)

\bibitem{xie15hed}
"Xie, S., Tu, Z.: Holistically-nested edge detection. In: ICCV (2015)

\bibitem{xue2023freestyle}
Xue, H., Huang, Z., Sun, Q., Song, L., Zhang, W.: Freestyle layout-to-image synthesis. In: CVPR (2023)

\bibitem{ye2023ip}
Ye, H., Zhang, J., Liu, S., Han, X., Yang, W.: Ip-adapter: Text compatible image prompt adapter for text-to-image diffusion models. arXiv preprint arXiv:2308.06721  (2023)

\bibitem{Yu2017DRN}
Yu, F., Koltun, V., Funkhouser, T.: Dilated residual networks. In: CVPR (2017)

\bibitem{zhang2023tale}
Zhang, J., Herrmann, C., Hur, J., Cabrera, L.P., Jampani, V., Sun, D., Yang, M.H.: A tale of two features: Stable diffusion complements dino for zero-shot semantic correspondence. In: NeurIPS (2023)

\bibitem{zhang2023controlnet}
Zhang, L., Agrawala, M.: Adding conditional control to text-to-image diffusion models. In: ICCV (2023)

\bibitem{zhang2018perceptual}
Zhang, R., Isola, P., Efros, A.A., Shechtman, E., Wang, O.: The unreasonable effectiveness of deep features as a perceptual metric. In: CVPR (2018)

\bibitem{zhao2023unicontrolnet}
Zhao, S., Chen, D., Chen, Y.C., Bao, J., Hao, S., Yuan, L., Wong, K.Y.K.: Uni-controlnet: All-in-one control to text-to-image diffusion models. In: NeurIPS (2023)

\bibitem{zhao2023vpd}
Zhao, W., Rao, Y., Liu, Z., Liu, B., Zhou, J., Lu, J.: Unleashing text-to-image diffusion models for visual perception. In: ICCV (2023)

\bibitem{zhou2017ade20k}
Zhou, B., Zhao, H., Puig, X., Fidler, S., Barriuso, A., Torralba, A.: Scene parsing through ade20k dataset. In: CVPR (2017)

\bibitem{zhou2022extract}
Zhou, C., Loy, C.C., Dai, B.: Extract free dense labels from clip. In: ECCV (2022)

\bibitem{zhu2017toward}
Zhu, J.Y., Zhang, R., Pathak, D., Darrell, T., Efros, A.A., Wang, O., Shechtman, E.: Toward multimodal image-to-image translation. In: NeurIPS (2017)

\end{thebibliography}
\end{document}